\documentclass[10pt,twocolumn,twoside]{IEEEtran}
\usepackage{amsmath,amssymb,cite}
\usepackage{hyperref} 
\usepackage{acronym}  
\usepackage{psfrag}
\usepackage{hhline}
\usepackage{upgreek}
\usepackage{booktabs}
\usepackage{algorithm}
\usepackage{relsize}
\usepackage{algorithmic}
\usepackage{setspace}
\usepackage{subfigure}
\usepackage{mathtools}
\usepackage{fancyhdr}
\usepackage{stfloats}
\usepackage{acronym} 
\usepackage{graphicx,color,float}
\usepackage{multirow}
\usepackage{mathrsfs}

\newtheorem{definition}{Definition}
\newtheorem{theorem}{Theorem}
\newtheorem{corollary}{Corollary}
\newtheorem{lemma}{Lemma}
\newtheorem{proposition}{Proposition}
\newtheorem{remark}{Remark}

\newcommand\V[1]  { \mathbf{#1} }
\newcommand\B[1]  { \boldsymbol{#1} }
\newcommand\up[1] {\mathrm{#1}}
\newcommand\set[1] {\mathcal{#1}}
\newcommand\sset[1] {\mathsmaller{\mathcal{#1}}}
\newcommand{\pun}{\hspace{0.2mm}\B{\cdot}\hspace{0.2mm}}
\newcommand{\paperTitle}{Generalized Maximum Entropy\\ for Supervised Classification}
\newcommand{\paperTitleMarkboth}{Generalized Maximum Entropy for Supervised Classification}

\acrodef{SVM}{support vector machine}
\acrodef{NN}{nearest neighbor}
\acrodef{NB}{naive Bayes}
\acrodef{LR}{logistic regression}
\acrodef{MEM}{maximum entropy machine}
\acrodef{CRF}{conditional random field}
\acrodef{LP}{linear program}
\acrodef{GP}{geometric program}
\acrodef{ERM}{empirical risk minimization}
\acrodef{RKHS}{reproducing kernel Hilbert space}
\acrodef{MRC}{minimax risk classifier}
\acrodef{RRM}{robust risk minimization}

\begin{document}

    \title{\paperTitle}

\author{
	\vspace{0.2cm}
	\IEEEauthorblockN{
		Santiago~Mazuelas,~\IEEEmembership{Senior Member,~IEEE,}  
		Yuan~Shen,~\IEEEmembership{Senior Member,~IEEE,} and Aritz~P\'{e}rez} 	
    \thanks{This research was supported in part by the Ramon y Cajal Grant RYC-2016-19383, Project PID2019-105058GA-I00, and Grant SEV-2017-0718 funded by MCIN / AEI / 10.13039/501100011033 and by FSE ``invest in your future,''  the National Natural Science Foundation of China under Grant 61871256, and the Basque Government through the ELKARTEK and BERC 2018-2021 programmes.}   
    \thanks{
        S.~Mazuelas is with the BCAM-Basque Center for Applied Mathematics, and IKERBASQUE-Basque Foundation for Science, Bilbao 48009, Spain (e-mail: smazuelas@bcamath.org).}
           \thanks{
        A.~P\'{e}rez is with the BCAM-Basque Center for Applied Mathematics, Bilbao 48009, Spain (e-mail: aperez@bcamath.org).}
	\thanks{Y.~Shen is with the Department of Electronic Engineering, Tsinghua University, and Beijing National Research Center for Information Science and Technology, Beijing 100084, China (e-mail: shenyuan\_ee@tsinghua.edu.cn).}
}
        \maketitle 
\markboth{IEEE Transactions on Information Theory}{Mazuelas, Shen, and P\'{e}rez: \paperTitleMarkboth}



\begin{abstract}
The maximum entropy principle advocates to $\mbox{evaluate}$ events' probabilities using a distribution that $\mbox{maximizes}$ entropy among those that satisfy certain expectations' constraints.  Such principle can be generalized for arbitrary decision problems where it corresponds to minimax approaches.  This paper establishes a framework for supervised classification based on the generalized maximum entropy principle that leads to minimax risk classifiers (MRCs). We develop learning techniques that determine MRCs for general entropy functions and provide performance guarantees by means of convex optimization. In addition, we describe the relationship of the presented techniques with existing classification methods, and quantify MRCs performance in comparison with the proposed  bounds and conventional methods. 
 \end{abstract}
\section{Introduction}\label{sec:intro}
\IEEEPARstart{T}{he maximum entropy} principle advocates to evaluate events' probabilities using a distribution that $\mbox{maximizes}$ entropy among those that satisfy certain expectations' constraints (uncertainty set) \cite{Jay:57,Jay:03,ShoJoh:80}. In addition, the maximum entropy provides a performance bound as long as the true underlying distribution is included in the uncertainty set.
Such principle was pioneered by Jaynes \cite{Jay:57} and dates back to Shannon's seminal work on coding \cite{Sha:48}: \emph{``...we consider the source with the maximum entropy subject to the statistical conditions we wish to retain. The entropy of this source determines the channel capacity.''} Maximum entropy methods have been generalized to use broader expectations' constraints \cite{DudPhiSha:07,DudPhi:04,SutSutMoh:19} and entropy functions \cite{Csi:91,Csi:96,AltSmo:06,GruDaw:04}. In addition, the minimax approach for decision making has been shown to correspond to the maximum entropy principle  \cite{GruDaw:04,Gru:00,Smi:74}. Techniques based on maximum entropy have been developed in multiple fields \cite{BucMac:91} including spectrum estimation \cite{CovTho:12}, coding \cite{HarTop:01}, neurocience \cite{TanJac:08}, species distribution modeling \cite{PhiAnd:06}, natural language processing, \cite{BerPie:96} and supervised classification \cite{MehRos:18,FarTse:16}. In addition, maximum entropy approaches with a reference distribution lead to minimum relative entropy (Kullback-Leibler divergence) techniques that have been used in multiple fields including Bayesian networks \cite{Min:01}, sequential inference \cite{MazSheWin:J13}, and inverse problems \cite{WooUlr:98}.  

\begin{figure}
\psfrag{l}[l][t][0.8]{\hspace{-3mm}$\begin{array}{cc}\mbox{Expected loss}\\ \ell(\up{h},\up{p})\end{array}$}
\psfrag{H}[l][t][0.8]{\hspace{-8mm}$\begin{array}{cc}\mbox{Maximum entropy}\\ H_\ell(\set{U}^{\V{a},\V{b}})\end{array}$}
\psfrag{Rh}[l][t][0.8]{\hspace{-6mm}$\begin{array}{cc}\mbox{MRC risk}\\R_\ell(\up{h}^{\V{a},\V{b}})\end{array}$}
\psfrag{R}[l][t][0.8]{\hspace{-3mm}$\begin{array}{cc}\mbox{Bayes risk}\\H_\ell(\up{p}^*)\end{array}$}
\psfrag{h*}[l][t][0.7]{\hspace{-0mm}Bayes classifier}
\psfrag{p*}[l][t][0.8]{\hspace{-0mm}$\up{p}^*$}
\psfrag{pa}[l][t][0.8]{\hspace{-0mm}$\up{p}^{\V{a},\V{b}}$}
\psfrag{ha}[l][t][0.8]{\hspace{-7mm}MRC $\up{h}^{\V{a},\V{b}}$}
\psfrag{U}[l][t][0.8]{\hspace{-3mm}$\set{U}^{\V{a},\V{b}}$}
\psfrag{Hp}[l][t][0.8]{\hspace{-18mm}$\begin{array}{cc}\mbox{Entropy}\\H_\ell(\up{p})\end{array}$}
\psfrag{M}[l][t][0.8]{\hspace{-7mm}$\ell(\up{h},\up{p}^{\V{a},\V{b}})$}
\includegraphics[width=0.485\textwidth]{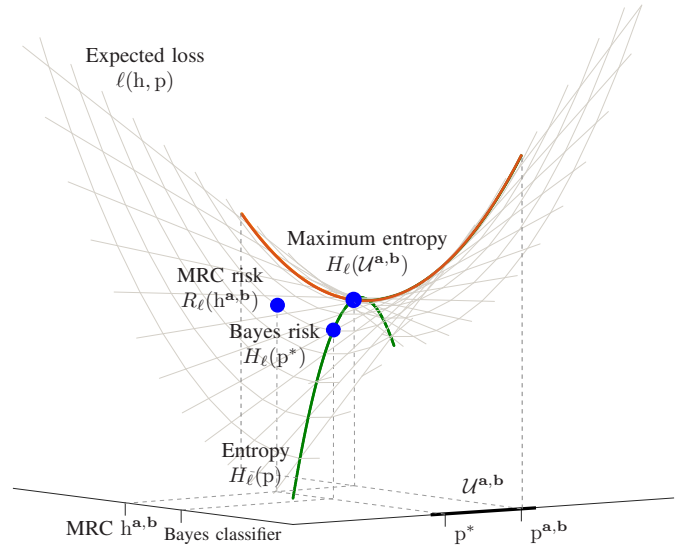}
	\caption{The maximum entropy principle for classification leads to MRCs since the maximum entropy coincides with  the minimum worst-case expected loss.\label{fig_entropy}}
\vspace{-0.3cm}
\end{figure}
Supervised classification techniques use training samples to obtain rules that serve to assign labels to instances. 
So-called pure-discriminative approaches \cite{Jeb:12,Bis:06}, such as \acp{SVM} and \acp{NN}, use training samples to learn a correspondence from instances to labels; while so-called conditional approaches \cite{Jeb:12,Bis:06}, such as \ac{LR} and \acp{CRF}, use training samples to learn a conditional distribution of labels given instances. In the prediction stage, discriminative approaches directly use the learnt correspondence to classify instances; while conditional approaches use the learnt conditional distribution to classify instances based on the assessed probabilities, e.g., as the most probable label.

Several current techniques for supervised classification have been derived as maximum entropy methods considering specific entropy functions and uncertainty sets of distributions with empirical marginals. In particular, \ac{LR} and \acp{MEM} have been shown to correspond with Shannon's conditional entropy and 0-1 entropy, respectively \cite{CorKuz:15,MehRos:18,LebLaf:01,BerPie:96,FarTse:16}. In addition, as shown in this paper, several other current classification techniques \cite{AsiXinBeh:15,FatAnq:16,GloTis:04} can be also seen as maximum entropy methods corresponding with \mbox{0-1} entropy and Shannon's relative entropy (mutual information). Current techniques maximize entropy over uncertainty sets of distributions that satisfy certain expectations' constraints but also have fixed marginals that coincide with the empirical (see e.g., \cite{FarTse:16} or Section 13.4 in \cite{MehRos:18}). Such uncertainty sets can result in a simplified learning process at the expense of inferior performance guarantees since they cannot include the true underlying distribution for a finite number of samples.

This paper shows that the principle of maximum entropy for classification yields a manifold of classification techniques by varying the entropy function and the uncertainty set considered. Such classification techniques are \acp{MRC} since the maximum entropy coincides with the minimum worst-case expected loss. In addition, \acp{MRC} minimize expected loss with respect to maximum entropy distributions, and the maximum entropy provides a performance bound for \acp{MRC}' risks in cases where the uncertainty set includes the true underlying distribution (see Fig.~\ref{fig_entropy}).

In this paper we develop the maximum entropy principle for supervised classification with general entropies and uncertainty sets. In particular, the main contributions of the paper are as follows.
\vspace{-0.009cm}
\begin{itemize}
\item We establish a framework for supervised classification based on the generalized maximum entropy principle. Such framework leads to \acp{MRC} that can utilize general entropies and provide performance guarantees.
\item We develop learning techniques that determine \acp{MRC} by means of convex optimization problems given by expectations' estimates. In addition, we show the correspondence between the accuracy of expectations' estimates and L1 regularization.  
\item We develop techniques that obtain tight performance guarantees for \acp{MRC} at learning. In addition, we present \acp{MRC}' generalization bounds in terms of the smallest maximum entropy and the training size. 
\item We describe how multiple existing techniques can be considered as special cases of the proposed framework using specific entropy functions and marginals' constraints.
\item We numerically quantify the performance of \acp{MRC} with respect to the proposed performance bounds and conventional techniques using benchmark datasets.
\end{itemize}
\emph{Notation:} calligraphic upper case letters denote sets, e.g., $\set{Z}$; $|\set{Z}|$ denotes de cardinality of set $\set{Z}$; $\Delta(\set{Z})$ denotes the set of probability distributions with support $\set{Z}$; vectors and matrices are denoted by bold lower and upper case letters, respectively, e.g., $\V{v}$ and $\V{M}$; for a vector $\V{v}$, $v^{(i)}$ denotes its $i$-th component, and $\V{v}^{\text{T}}$ and $\V{v}_+$ denote its transpose and positive part, respectively;
probability distributions and classification rules are denoted by upright fonts, e.g., $\up{p}$ and $\up{h}$; $\mathbb{E}_{z\sim \up{p}}$ or simply $\mathbb{E}_{\up{p}}$ denotes the expectation w.r.t. probability distribution $\up{p}$ of random variable $z$; $\preceq$ and $\succeq$ denote vector (component-wise) inequalities; $\V{1}$ denotes a vector with all components equal to $1$;  $\mathbb{I}\{\pun\}$ denotes the indicator function; finally, for a function of two variables $f:\set{X}\times\set{Y}\to\set{Z}$, $f(x,\pun)$ denotes the function $f(x,\pun):\set{Y}\to\set{Z}$ obtained by fixing $x\in\set{X}$. 
\section{Theoretical framework}
In this section, we first briefly describe general decision problems and generalized entropies, then we present the framework of generalized maximum entropy for supervised classification. For the readers' convenience, we provide in Table~\ref{notions} a list with the main notions used in the paper and their corresponding notations. 
\subsection{The maximum entropy principle for general decision problems}\label{sec-dec-prob}
Given a set of feasible probability distributions (uncertainty set) described by constraints on the values of certain expectations, the principle of maximum entropy advocates to evaluate events' probabilities by using a distribution with maximum entropy  among those in the uncertainty set \cite{Jay:57,ShoJoh:80}.  The entropy considered by classical approaches is Shannon entropy, while such principle can be extended for general decision problems and their corresponding generalized entropies \cite{GruDaw:04} as described below. In the next subsection, we show how such general theory leads to the presented framework for the specific decision problem corresponding to supervised classification.


A decision problem comprises the selection of an action (decision) whose consequences depend on the outcome of a state variable (see e.g., \cite{GruDaw:04,Ber:85}). For instance, a prefix code can be chosen to encode words or a classification rule can be chosen to assign labels to instances. Mathematically, a general decision problem consists of three objects $(\set{S},\set{A},\ell)$, where $\set{S}$ is the set of possible states, e.g., set of words or set of instance-label pairs; $\set{A}$ is the set of possible actions, e.g., prefix codes or classification rules; and $\ell:\set{A}\times\set{S}\to(-\infty,\infty]$ is a function such that $\ell(a,s)$ quantifies the loss achieved by action $a\in\set{A}$ when state $s\in\set{S}$ takes place, e.g., bit-length of an encoded word or error of a classified instance. 

In the following, we overload the notation for loss functions and denote expected loss of action $a\in\set{A}$ with respect to a probability distribution $\up{p}\in\Delta(\set{S})$ as $\ell(a,\up{p})\coloneqq\mathbb{E}_{s\sim \up{p}}\{\ell(a,s)\}$. The expected $\ell$-loss that determines the ultimate performance of an action is that computed with respect to the true underlying distribution of the state denoted by $\up{p}^*\in\Delta(\set{S})$. Such expected $\ell$-loss is known as $\ell$-risk denoted by $R_\ell(a)\coloneqq\ell(a,\up{p}^*)$, its smallest value is known as Bayes risk, and the action achieving such minimum is known as Bayes act.

Each decision problem has a corresponding $\ell$-entropy function $H_\ell$ that assigns each probability distribution $\up{p}\in\Delta(\set{S})$ with the smallest expected $\ell$-loss with respect to such distribution, that is
$$H_\ell(\up{p})\coloneqq\inf_{a\in\set{A}}\ell(a,\up{p})$$
which is a concave function for any decision problem since $\ell(a,\up{p})$ is linear on $\up{p}$. Such function is referred to as generalized entropy \cite{GruDaw:04} since, as shown below, it generalizes the entropy function introduced by Shannon to general decision problems.    

\emph{Example}: in the problem of coding (``code-length-game'' \cite{HarTop:01}), $\set{S}$ is the set of words, $\set{A}$ is the set of prefix codes, and $\ell(a,s)$ is the number of bits that code $a\in\set{A}$ uses to encode word  $s\in\set{S}$, i.e., $\ell(a,s)=[a(s)]$ for $[a(s)]$ the length of code $a(s)$. Since prefix codes can be represented by probability distributions over words, $\up{q}(s)=2^{-[a(s)]}\in\Delta(\set{S})$, the entropy function corresponding with the problem of coding is Shannon entropy, that is
$$H_\ell(\up{p})=\min_{\up{q}\in\Delta(\set{S})}\mathbb{E}_{s\sim \up{p}}\{-\log_2\up{q}(s)\}=-\mathbb{E}_{s\sim \up{p}}\{\log_2 \up{p}(s)\}.$$

The principle of maximum entropy corresponds to the minimax approach for decision making. Such correspondence follows because, under suitable regularity conditions \cite{GruDaw:04}, the maximum entropy over an uncertainty set of distributions $\set{U}\subset\Delta(\set{S})$ coincides with the minimum worst-case expected loss with respect to distributions in $\set{U}$. Specifically,  we have that
$$\sup_{\up{p}\in\set{U}}H_\ell(\up{p})=\underset{\up{p}\in\set{U}}{\sup}\,\underset{\rule[-.44\baselineskip]{0pt}{\baselineskip} a\in\set{A}}{\inf}\,\ell(a,\up{p})=\underset{\rule[-.44\baselineskip]{0pt}{\baselineskip}a\in\set{A}}{\inf}\,\underset{\up{p}\in\set{U}}{\sup}\,\ell(a,\up{p}).$$
Furthermore, if the uncertainty set $\set{U}$ includes the true underlying distribution of the state $\up{p}^*$, the maximum entropy provides an upper bound for the Bayes risk and for the risk of the minimax action.
\begin{table*}\caption{Main notations used in the paper}
\setstretch{1.1}
\small
\centering
\begin{tabular}{ll}\toprule
Notation &Meaning\\\hline\hline
$\Delta(\set{X}\times\set{Y})$& set of probability distributions over instances $\set{X}$ and labels $\set{Y}$\\
$\up{p}(x,y)$&probability assigned by distribution $\up{p}$ to instance $x$ and label $y$\\
$ \text{T}(\set{X},\set{Y})$& set of classification rules from instances $\set{X}$ to labels $\set{Y}$\\
$\up{h}(y|x)$&probability assigned by classification rule $\up{h}$ to label $y$ for instance $x$\\
$\ell(\up{h},(x,y))$, $\ell(\up{h},\up{p})$& loss and expected loss of classification rule $\up{h}$ at example $(x,y)$ and w.r.t. distribution $\up{p}$\\
$L(\up{q},y)$& score of probability distribution $\up{q}\in\Delta(\set{Y})$ at label $y$\\
$\up{p}^*$& true underlying distribution of instance-label pairs\\
$R_\ell(\up{h})$& $\ell$-risk of classification rule $\up{h}$ given by \eqref{risk}\\
$H_\ell(\up{p})$, $H_\ell(\set{U})$&$\ell$-entropy of $\up{p}$ given by \eqref{entropy} and maximum $\ell$-entropy of probability distributions in $\set{U}$ \\
$(x_1,y_1),(x_2,y_2),\ldots,(x_n,y_n)$& training samples\\
$\Phi:\set{X}\times\set{Y}\to\mathbb{R}^m$&feature mapping\\
$\V{a},\V{b}$& lower and upper endpoints of interval estimates for the expectations of $\Phi$\\
$\set{U}^{\V{a},\V{b}}$&uncertainty set of distributions given by expectations' constraints as in \eqref{uncertainty}\\
$\up{h}^{\V{a},\V{b}}$, $\up{p}^{\V{a},\V{b}}$&$\ell$-\ac{MRC} for $\set{U}^{\V{a},\V{b}}$ and maximum $\ell$-entropy distribution over $\set{U}^{\V{a},\V{b}}$\\
$\underline{R}^{\,\V{a},\V{b}}_{\;\!\ell}$&lower bound for the $\ell$-risk of $\up{h}^{\V{a},\V{b}}$ given by $\underline{\mathscr{P}_{\ell}}^{\V{a},\V{b}}$ in \eqref{opt-bound}\\
$\B{\tau}_\infty$, $\B{\tau}_n$, $\V{a}_n$, $\V{b}_n$ &expectation of feature mapping, and estimates from $n$ training samples given by \eqref{interval}\\
$\up{p}_n(x)$, $\up{p}_0(y)$& empirical instances' marginal of $n$ samples and reference labels' marginal \\
$\set{U}^{\V{a},\V{b}}_x$&uncertainty set of distributions in $\set{U}^{\V{a},\V{b}}$ with instances' marginal $\up{p}_n(x)$ as in \eqref{uncertainty-emp-x}\\
$\set{U}^{\V{a},\V{b}}_y$&uncertainty set of distributions in $\set{U}^{\V{a},\V{b}}$ with labels' marginal $\up{p}_0(y)$ as in \eqref{uncertainty-emp-y}\\
$\set{U}^{\V{a},\V{b}}_{x,y}$&uncertainty set of distributions in $\set{U}^{\V{a},\V{b}}_x\cap\set{U}^{\V{a},\V{b}}_y$ as in \eqref{uncertainty-emp-xy}\\
\bottomrule\label{notions}
\end{tabular}
\end{table*}
\subsection{Generalized entropies for supervised classification}
Learning techniques for supervised classification use training samples to obtain a classification rule that serves to assign labels to instances. The learning stage in supervised classification can be seen as a decision problem in which the set of states corresponds to instance-label pairs, the set of actions corresponds to classification rules, and the loss function quantifies the classification loss.

Let $\set{X}$ be the set of \emph{instances}, e.g., grayscale images, and $\set{Y}$ be the set of \emph{labels}, e.g., 0 to 9 digits; in the following, we take both $\set{X}$ and $\set{Y}$ to be finite and we represent $\set{Y}$ by $\{1,2,\ldots,|\set{Y}|\}$. Commonly, the cardinality of $\set{X}$ is much larger than that of $\set{Y}$, for example in hand-written digit classification with 28x28 8-bits pixel images, $|\set{X}|=256^{784}$ and $|\set{Y}|=10$. We denote by $\Delta(\set{X}\times\set{Y})$ the set of probability distributions on $\set{X}\times\set{Y}$; if $\up{p}\in \Delta(\set{X}\times\set{Y})$, we denote by $\up{p}(x,y)$ the probability assigned by $\up{p}$ to the instance-label pair $(x,y)$, and by $\up{p}(x)$ and $\up{p}(y)$ the corresponding marginal probability distributions for instances and labels, respectively.

\emph{Classification rules} are given by functions from instances to probability distributions on labels (Markov transitions). We denote the set of all classification rules by $\text{T}(\set{X},\set{Y})$, and for $\up{h}\in  \text{T}(\set{X},\set{Y})$ we denote by $\up{h}(y|x)$ the probability assigned by $\up{h}$ to label $y$ for instance $x$. In discriminative approaches, $\up{h}(y|x)$ represents the probability with which instance $x\in\set{X}$ is classified by label $y\in\set{Y}$ ($\up{h}(y|x)\in\{0,1\}$ if labels are deterministically assigned to instances). In conditional approaches, $\up{h}(y|x)$ represents the probability assessment for label $y\in\set{Y}$ given instance $x\in\set{X}$. Note that probability distributions $\up{p}\in \Delta(\set{X}\times\set{Y})$ and classification rules $\up{h}\in \text{T}(\set{X},\set{Y})$ can be represented by positive vectors in $\mathbb{R}^{|\set{X}||\set{Y}|}$ since $\set{X}$ and $\set{Y}$ are taken to be finite sets.

The \emph{classification loss} of rule $\up{h}$ at instance-label pair $(x,y)$ quantifies the loss of the rule evaluated at instance $x$, $\up{h}(\pun|x)\in\Delta(\set{Y})$, when the label is $y\in\set{Y}$. Therefore, there is a correspondence between classification losses $\ell$ and functions $L:\Delta(\set{Y})\times \set{Y}\to (-\infty,\infty]$ with 
$$\ell(\up{h},(x,y))=L(\up{h}(\pun|x),y).$$
In the statistical literature (see e.g., \cite{GneRaf:07}), such functions $L$ that assess distributions in $\Delta(\set{Y})$ at values in $\set{Y}$ are referred to as scoring functions and are usually employed to evaluate probability assessments.  We consider general classification losses $\ell$ corresponding to scoring functions $L$ that are lower semi-continuous and convex in their first argument. In the context of probability assessments, preferred scoring functions are those referred to as proper \cite{GneRaf:07}, while, in the context of discriminative approaches for classification, preferred scoring functions are those referred to as classification-calibrated or Fisher-consistent \cite{BarJorMca:06,Lin:04,ReiWill:10}.

The $\ell$\emph{-risk for classification} of rule $\up{h}$ is its expected classification loss with respect to the true underlying distribution of instance-label pairs $\up{p}^*\in\Delta(\set{X}\times\set{Y})$, i.e.,
\begin{align}\label{risk}R_\ell(\up{h})\coloneqq\ell(\up{h},\up{p}^*).\end{align}
The smallest possible risk is known as Bayes risk and the classification rule achieving such minimum is known as Bayes  classifier.
\begin{definition}\label{def-entropy}
The $\ell$\emph{-entropy for classification} is the entropy function associated with classification loss $\ell$, that is, for each distribution $\up{p}\in\Delta(\set{X},\set{Y})$ 
\begin{align}\label{entropy}H_\ell(\up{p})\coloneqq&\min_{\up{h}\in \text{T}(\set{X},\set{Y})}\ell(\up{h},\up{p})\end{align}
with $\text{T}(\set{X},\set{Y})$ the set of classification rules from instances $\set{X}$ to labels $\set{Y}$.
In addition, we denote the \emph{maximum $\ell$-entropy for classification} of distributions in a compact uncertainty set $\set{U}\subset\Delta(\set{X}\times\set{Y})$ by\footnote{The $\inf$ and $\sup$ in the general definition of entropy and maximum entropy become $\min$ and $\max$ here since the score functions considered are lower semi-continuous and convex, and $\text{T}(\set{X},\set{Y})$ is a compact set since $\set{X}$ are $\set{Y}$ are finite.}
\begin{align}H_\ell(\set{U})\coloneqq\max_{\up{p}\in\set{U}}H_\ell(\up{p}).\end{align}
The $\ell$-entropy for classification of the true underlying distribution coincides with the smallest $\ell$-risk, hence, the entropy functions above have been also referred to as Bayes risk functions \cite{ReiWill:10}. 
\end{definition}

\subsection{Examples}
		\begin{table*}
\caption{Examples of classification losses with their corresponding risks and entropies.}
\small
\centering
\setlength{\tabcolsep}{10pt}
\def\arraystretch{1.3}
\begin{tabular}{lllll}\toprule
&\hspace{0.cm}Loss $\ell(\up{h},(x,y))$&\hspace{0.07cm}Score $L(\up{q},y)$&\hspace{1.4cm} Risk $R_\ell(\up{h})$ &\hspace{0.95cm} Entropy $H_\ell(\up{p})$\\ \hhline{=====}\rule{0pt}{3ex}0-1-loss& $1-\up{h}(y|x)$& $1-\up{q}(y)$& $1-\underset{x\in\set{X},y\in\set{Y}}{\sum}\up{p}^*(x,y)\up{h}(y|x)$ &$1-\underset{x\in\set{X}}{\sum}\underset{y\in\set{Y}}{\max}\,\up{p}(x,y)$\\\midrule
log-loss& $-\log \up{h}(y|x)$&$-\log \up{q}(y)$&$-\underset{x\in\set{X},y\in\set{Y}}{\sum}\up{p}^*(x,y)\log \up{h}(y|x)$&$\underset{x\in\set{X},y\in\set{Y}}{\sum}\up{p}(x,y)\log \frac{\up{p}(x)}{\up{p}(x,y)}$\\\midrule
$\alpha$-loss& $\beta\big(1-\up{h}(y|x)^{\frac{1}{\beta}}\big)$&$\beta\big(1-\up{q}(y)^{\frac{1}{\beta}}\big)$&$\beta\Big(1-\underset{x\in\set{X},y\in\set{Y}}{\sum}\up{p}^*(x,y)\up{h}(y|x)^{\frac{1}{\beta}}\Big)$&$\beta\Big(1-\underset{x\in\set{X}}{\sum}\big(\underset{y\in\set{Y}}{\sum}\up{p}(x,y)^\alpha\big)^{1/\alpha}\Big)$\\\midrule
log rel-loss&$\log\frac{\up{p}_0(y)}{\up{h}(y|x)}$&$\log\frac{\up{p}_0(y)}{\up{q}(y)}$&$\underset{x\in\set{X},y\in\set{Y}}{\sum}\up{p}^*(x,y)\log\frac{\up{p}_0(y)}{\up{h}(y|x)}$&
$\underset{x\in\set{X},y\in\set{Y}}{\sum}\up{p}(x,y)\log\frac{\up{p}(x)\up{p}_0(y)}{\up{p}(x,y)}$\\
\bottomrule\label{tab-examples}\end{tabular}
\end{table*}




Table~\ref{tab-examples} shows the loss function, score, risk, and entropy corresponding with 0-1-loss, log-loss, $\alpha$-loss, and log relative-loss. The risk and entropy functions in the table can be directly derived from the loss expressions using \eqref{risk} and \eqref{entropy}.

The \mbox{0-1-loss} is an appropriate classification loss for the discriminative approach since it quantifies the classification error. The corresponding score function is classification-calibrated and 0-1-risk coincides with the probability of classification error. It has been recently shown that 0-1-loss can enable tractable learning using non-deterministic classification rules \cite{FatAnq:16,FarTse:16,MazZanPer:20}. However, alternative loss functions such as hinge-loss, exponential-loss, and log-loss are often used as surrogates of 0-1-loss \cite{BarJorMca:06,Zha:04} in cases when the latter becomes intractable. 

The log-loss is an appropriate classification loss for the conditional approach since it quantifies the negative log-likelihood. The corresponding score function is strictly proper and log-entropy coincides with Shannon's conditional entropy. The log-loss is especially suitable among loss functions that evaluate probability assessments \cite{GneBalRaf:07} because it can enable tractable techniques and it provides an upper bound on any proper loss \cite{PaiWor:18}.


The $\alpha$-losses defined for $\alpha\in(0,1)\cup(1,\infty)$ and shown in Table~\ref{tab-examples} taking $\beta=\alpha/(\alpha-1)$, generalize 0-1-loss and \mbox{log-loss} since they correspond to $\alpha$-losses when $\alpha$ tends to infinity and one, respectively \cite{SypDiaSanKai:19,LiaKosSanCal:18,SypDiaSanDas:20}.  Large (resp. small) values of $\alpha$ reduce (resp. increase) the effects of low probability assessments \cite{SypDiaSanDas:20}. The corresponding score functions have been introduced as a measure of information leakage in \cite{LiaKosSanCal:18}, and $\alpha$-entropy can be also written in terms of  the Arimoto conditional entropy of order $\alpha$ \cite{SypDiaSanDas:20}.

In case of having a reference rule $\up{h}_0$, a relative loss \cite{GruDaw:04} with respect to $\up{h}_0$ can be defined as 
\begin{align}\ell_{\up{h}_0}(\up{h},(x,y))=\ell(\up{h},(x,y))-\ell(\up{h}_0,(x,y)).\end{align}
The log-relative loss corresponds to log-loss with respect to a reference rule that is constant over instances, i.e., $\up{h}_0(\cdot|x)=\up{p}_0\in\Delta(\set{Y})$, $\forall x\in\set{X}$, (e.g., prior class-probabilities). In this case, the associated log-relative entropy for classification  coincides with the negative Shannon's mutual information if $\up{p}_0(y)=\up{p}(y)$.

\subsection{Generalized maximum entropy and minimax risk classifiers}
Supervised classification techniques aim to determine a classification rule $\up{h}\in \text{T}(\set{X},\set{Y})$ with reduced risk $R_\ell(\up{h})=\ell(\up{h},\up{p}^*)$ in cases where the true underlying distribution of instance-label pairs $\up{p}^*\in\Delta(\set{X}\times\set{Y})$ is unknown and only training samples $(x_1,y_1),(x_2,y_2),\ldots,(x_n,y_n)$ are available. Given an uncertainty set of distributions that satisfy certain expectations' constraints determined by the training samples, the principle of maximum entropy applied to supervised classification prescribes to evaluate expected classification losses using a distribution that maximizes entropy over the uncertainty set.


The following results show that the maximum entropy principle for supervised classification corresponds to \acp{MRC} that mininize the worst-case expected classification loss (see Fig.~\ref{fig_entropy} for a pictorial description of such results).
\begin{definition}\label{def-MRC}
Let $\set{U}\subset\Delta(\set{X}\times\set{Y})$ be a convex and compact uncertainty set of distributions. We say that a classification rule $\up{h}^{\sset{U}}\in \text{T}(\set{X},\set{Y})$ is an $\ell$-\ac{MRC} for $\set{U}$ if $\up{h}^{\sset{U}}$ minimizes the maximum expected $\ell$-loss with respect to distributions in $\set{U}$, i.e.,
\begin{align}\label{mrc-def}\up{h}^{\sset{U}}\in\arg\min_{\up{h}\in \text{T}(\set{X},\set{Y})}\max_{\up{p}\in\set{U}}\ell(\up{h},\up{p}).\end{align}
\end{definition}

\begin{theorem}\label{th0}
If $\set{U}\subset\Delta(\set{X}\times\set{Y})$ is a convex and compact uncertainty set, $\up{h}^{\sset{U}}\in \text{T}(\set{X},\set{Y})$ is an $\ell$-\ac{MRC} for $\set{U}$, and $\up{p}^{\sset{U}}\in\Delta(\set{X}\times\set{Y})$ is a probability distribution with maximum $\ell$-entropy over $\set{U}$, then, 
\begin{align}H_\ell(\set{U})=\ell(\up{h}^{\sset{U}},\up{p}^{\sset{U}})=\min_{\up{h}\in \text{T}(\set{X},\set{Y})}\max_{\up{p}\in\set{U}}\ell(\up{h},\up{p})\end{align}
that is, $\up{h}^{\sset{U}}$ minimizes the expected $\ell$-loss with respect to $\up{p}^{\sset{U}}$, $\up{p}^{\sset{U}}$ maximizes the expected $\ell$-loss of $\up{h}^{\sset{U}}$ over distributions in $\set{U}$, and such expected $\ell$-loss coincides with the maximum \mbox{$\ell$-entropy} over $\set{U}$. 
\end{theorem}
\begin{IEEEproof}
Firstly, 
$$\max_{\up{p}\in\set{U}}\min_{\up{h}\in \text{T}(\set{X},\set{Y})}\ell(\up{h},\up{p})=\max_{\up{p}\in\set{U}}H_\ell(\up{p})=H_\ell(\up{p}^{\sset{U}})$$
by definition of $\ell$-entropy and $\up{p}^{\sset{U}}$. Then,
\begin{align*}\max_{\up{p}\in\set{U}}\min_{\up{h}\in \text{T}(\set{X},\set{Y})}\ell(\up{h},\up{p})&=\min_{\up{h}\in \text{T}(\set{X},\set{Y})}\ell(\up{h},\up{p}^{\sset{U}})\\&\leq \ell(\up{h}^{\sset{U}},\up{p}^{\sset{U}}).\end{align*}
In addition 
\begin{align*}\min_{\up{h}\in \text{T}(\set{X},\set{Y})}\max_{\up{p}\in\set{U}}\ell(\up{h},\up{p})&=\max_{\up{p}\in\set{U}}\ell(\up{h}^{\sset{U}},\up{p})\\&\geq\ell(\up{h}^{\sset{U}},\up{p}^{\sset{U}}).\end{align*}
Then, the result follows because
$$\max_{\up{p}\in\set{U}}\min_{\up{h}\in \text{T}(\set{X},\set{Y})}\ell(\up{h},\up{p})=\min_{\up{h}\in \text{T}(\set{X},\set{Y})}\max_{\up{p}\in\set{U}}\ell(\up{h},\up{p})$$
that is obtained using classical minimax theorems (see e.g., \cite{Sim:95}) since $\ell(\up{h},\up{p})$ is lower semi-continuous and convex in $\up{h}$ because $L$ is lower semi-continuous and convex in its first argument, $\ell(\up{h},\up{p})$ is concave in $\up{p}$ because it is linear in $\up{p}$, and both $\set{U}$ and $\text{T}(\set{X},\set{Y})$ are convex and compact.
\end{IEEEproof}

An important direct consequence of the above theorem is as follows.
\begin{corollary}\label{cor0}
If the uncertainty set $\set{U}$ includes the true underlying distribution  $\up{p}^*$, then both the Bayes $\ell$-risk and the $\ell$-risk of an $\ell$-\ac{MRC} for $\set{U}$ are upper bounded by the maximum $\ell$-entropy over $\set{U}$, that is 
\begin{align}\min_{\up{h}\in \text{T}(\set{X},\set{Y})}R_\ell(\up{h})&\leq H_\ell(\set{U})\nonumber\\
R_\ell(\up{h}^{\sset{U}})&\leq H_\ell(\set{U}).\end{align}
In addition, if $\up{p}^*$ maximizes the $\ell$-entropy over $\set{U}$, then \mbox{$\ell$-\acp{MRC}} for $\set{U}$ are Bayes classifiers and the maximum \mbox{$\ell$-entropy} over $\set{U}$ is the Bayes risk, i.e.,
\begin{align}\min_{h\in \text{T}(\set{X},\set{Y})}R_\ell(h)= R_\ell(\up{h}^{\sset{U}})=H_\ell(\set{U}).\end{align}
\end{corollary}

As described above, the uncertainty sets of distributions considered by maximum entropy methods are defined by expectations' constraints.
Such uncertainty sets can be determined from expectations' estimates of vector-valued functions  $\Phi:\set{X}\times\set{Y}\to\mathbb{R}^m$ referred to as \emph{feature mappings}. In what follows we consider uncertainty sets of distributions $\set{U}^{\V{a},\V{b}}$ given by constraints $\V{a}\preceq\mathbb{E}_\up{p}\{\Phi(x,y)\}\preceq\V{b}$ where $\V{a}, \V{b}\in\mathbb{R}^m$ are lower and upper endpoints of expectations' interval estimates that are obtained from training samples. 

Feature mappings are commonly used in machine learning (see e.g., \cite{MehRos:18}) to represent examples as real vectors using functionals over instances $\psi:\set{X}\to\mathbb{R}$ referred to as scalar features. The most common manner to define feature mappings over $\set{X}$ and $\set{Y}$ is to use multiple scalar features over $\set{X}$ together with one-hot encodings of the elements of $\set{Y}$ as follows \cite{TsoJoaHofAlt:05,CraDekKesShaSin:06,MehRos:18,FarTse:16}
\begin{align}\label{feature_map}\Phi(x,y)=\V{e}_y\otimes \Psi(x)=\left[\begin{array}{c}\mathbb{I}\{y=1\}\Psi(x)\\\mathbb{I}\{y=2\}\Psi(x)\\\vdots\\\mathbb{I}\{y=|\set{Y}|\}\Psi(x)\end{array}\right]\end{align}
where $\Psi(x)=[\psi_1(x),\psi_2(x),\ldots,\psi_t(x)]^{\text{T}}$ is given by $t$ scalar features $\psi_1,\psi_2,\ldots,\psi_t$; $\V{e}_y$ is the $y$-th vector in the standard basis of $\mathbb{R}^{|\set{Y}|}$; and $\otimes$ denotes the Kronecker product. Such feature mappings represent examples $(x,y)$ by $m$-dimensional real vectors given by $t$ scalar features from instances and the one-hot encoding of labels so that $m=|\set{Y}|t$. Usual scalar features are those given by polynomials of instances' components or by functions in a \ac{RKHS}. 

\section{Generalized maximum entropy for supervised classification with performance guarantees}\label{sec-3} 
This section shows how $\ell$-\acp{MRC} following the generalized maximum entropy principle can be determined by means of convex optimization. In addition, we develop techniques that obtain tight performance guarantees at learning.
\subsection{Uncertainty sets given by expectations' constraints}
We consider uncertainty sets of distributions $\set{U}^{\V{a},\V{b}}$ given by expectations' constraints as
\begin{align}\label{uncertainty}\set{U}^{\V{a},\V{b}}=\{\up{p}\in\Delta(\set{X}\times\set{Y}):\  \V{a}\preceq\mathbb{E}_\up{p}\{\Phi(x,y)\}\preceq\V{b}\}\end{align}  
where $\V{a}$ and $\V{b}$ are lower and upper endpoints of interval expectations' estimates of a feature mapping $\Phi:\set{X}\times\set{Y}\to~\mathbb{R}^m$.\footnote{When we use expectations' point estimates, i.e., $\V{a}=\V{b}$, we denote the corresponding uncertainty sets and \acp{MRC} as $\set{U}^{\V{a}}$ and $\up{h}^{\V{a}}$.} Such uncertainty sets are compact and convex subsets of $\Delta(\set{X}\times\set{Y})$ because $\set{X}$ and $\set{Y}$ are finite, and the expectation operator is linear.



Uncertainty sets given by expectations' constraints can be chosen to include the true underlying distribution with high probability. 
Specifically, if $\V{a}$ and $\V{b}$ define expectations' confidence intervals at level $1-\delta$, the uncertainty set $\set{U}^{\V{a},\V{b}}$ includes the true underlying distribution $\up{p}^*$ with probability at least $1-\delta$. Such expectations' interval estimates can be obtained from empirical expectations of training samples $(x_1,y_1),(x_2,y_2),\ldots,(x_n,y_n)$ 
as 
\begin{align}\label{interval} &\V{a}_n=\B{\tau}_n-\frac{\B{\lambda}}{\sqrt{n}},\  \V{b}_n=\B{\tau}_n+\frac{\B{\lambda}}{\sqrt{n}}\nonumber\\
&\mbox{for }\B{\tau}_n=\sum_{i=1}^n\frac{\Phi(x_{i},y_{i})}{n}\end{align}
where $\B{\lambda}\succeq \V{0}$ determines the size of the interval estimates for different confidence levels, and $\B{\lambda}=\V{0}$ corresponds with the case of using expectations' point estimates $\V{a}_n=\V{b}_n=\B{\tau}_n$. 
\subsection{\acp{MRC} for general classification losses}
The following result shows how \acp{MRC} can be determined by a linear-affine combination of the feature mapping. The coefficients of such combination can be obtained at learning by solving the maximum entropy dual problem
\begin{align}\label{opt-prob}\begin{split}
\mathscr{P}_{\ell}^{\V{a},\V{b}}:\quad \underset{\B{\mu},\B{\eta},\nu}{\min} &\quad \frac{1}{2}(\V{b}-\V{a})^{\text{T}}\B{\eta}-\frac{1}{2}(\V{b}+\V{a})^{\text{T}}\B{\mu}-\nu\\
\mbox{s.t.}&\quad \B{\Phi}(x,\pun)\B{\mu}+\nu\V{1}\in\set{L},\quad  \forall x\in\set{X}
\\& \quad
\B{\eta}+\B{\mu}\succeq\V{0},\;\B{\eta}-\B{\mu}\succeq\V{0}\end{split}\end{align}
where $\B{\Phi}(x,\pun)$ denotes the $|\set{Y}|\times m$ matrix with rows given by $\Phi^{\text{T}}(x,y)$ for $y\in\set{Y}$, and $\set{L}\subset\mathbb{R}^{|\set{Y}|}$ is the convex set 
\begin{align}\label{gamma}\set{L}=\{&\V{c}\in\mathbb{R}^{|\set{Y}|}:\    \exists\, \up{q}\in\Delta(\set{Y})\mbox{ with } \V{c}+L(\up{q},\pun)\preceq \V{0}\}\end{align}
given by the score function $L$ corresponding to the classification loss $\ell$.
\begin{theorem}\label{th1}
Let $\V{a},\V{b}\in\mathbb{R}^m$ satisfy $\set{U}^{\V{a},\V{b}}\neq~\emptyset$, and $\B{\mu}^*,\B{\eta}^*,\nu^*$ be a solution of $\mathscr{P}_{\ell}^{\V{a},\V{b}}$. 
If for any $x\in\set{X}$, $y\in\set{Y}$, a classification rule $\up{h}^{\V{a},\V{b}}\in \text{T}(\set{X},\set{Y})$ satisfies
\begin{align}\label{opt-sol}\ell(\up{h}^{\V{a},\V{b}},(x,y))+\Phi(x,y)^{\text{T}}\B{\mu}^*+\nu^*\leq0 \end{align}
then $\up{h}^{\V{a},\V{b}}$ is an $\ell$-\ac{MRC} for $\set{U}^{\V{a},\V{b}}$. 
In addition, the maximum $\ell$-entropy over $\set{U}^{\V{a},\V{b}}$ is 
\begin{align}\label{max-ent}H_\ell(\set{U}^{\V{a},\V{b}})=\frac{1}{2}(\V{b}-\V{a})^{\text{T}}\B{\eta}^*-\frac{1}{2}(\V{b}+\V{a})^{\text{T}}\B{\mu}^*-\nu^*.\end{align}
\end{theorem}
\begin{IEEEproof}
	See Appendix~\ref{apd:proof_th_1}.
	\end{IEEEproof}
\begin{remark}
Problem $\mathscr{P}_{\ell}^{\V{a},\V{b}}$ is a convex optimization problem because $\set{L}$ is a convex set since $L$ is a convex function in its first argument. Moreover, a classification rule $\up{h}^{\V{a},\V{b}}$ satisfying \eqref{opt-sol} always exists because
for each $x\in\set{X}$, there exists $\up{h}^{\V{a},\V{b}}(\pun|x)\in\Delta(\set{Y})$ that verifies \eqref{opt-sol} since the solution $\B{\mu}^*,\nu^*$ satisfies the constraints of $\mathscr{P}_{\ell}^{\V{a},\V{b}}$.
 
Even though $\ell$-\acp{MRC} minimize the maximum expected loss over general classification rules, as shown in \eqref{opt-sol}, \mbox{$\ell$-\acp{MRC}} are characterized by a simple parametric expression. Specifically, $\ell$-\acp{MRC} are determined by the classification loss and the linear-affine combination of the feature mapping \mbox{$\Phi(x,y)^{\text{T}}\B{\mu}^*+\nu^*$} with parameters  $\B{\mu}^*,\nu^*$ obtained by solving $\mathscr{P}_{\ell}^{\V{a},\V{b}}$ at learning.
\end{remark}

If $\V{a}=\V{b}$, the maximum entropy dual problem $\mathscr{P}_{\ell}^{\V{a},\V{b}}$ becomes 
\begin{align}\label{opt-eq}
\begin{split}
\mathscr{P}_{\ell}^{\V{a}}:\quad \underset{\B{\mu},\nu}{\min} &\quad  -\V{a}^{\text{T}}\B{\mu}-\nu\\
\mbox{s.t.} & 
\quad \B{\Phi}(x,\pun)\B{\mu}+\nu\V{1}\in\set{L},\quad  \forall x\in\set{X}.
\end{split}
\end{align}
The following result shows that the usage of $\V{a}\neq\V{b}$ is equivalent to imposing a regularization that penalizes parameters corresponding to poorly estimated expectations.
\begin{corollary}\label{cor-reg}
Let $\V{a}_n,\V{b}_n\in\mathbb{R}^m$ be expectations' interval estimates obtained from training samples $(x_1,y_1),(x_2,y_2),\ldots,(x_n,y_n)$ as given by \eqref{interval}. Then, the maximum entropy dual problem $\mathscr{P}_{\ell}^{\V{a},\V{b}}$ becomes
\begin{align}\label{opt-reg}
\begin{split}\underset{\B{\mu},\nu}{\min}&\quad-\frac{1}{n}\overset{n}{\underset{i=1}{\sum}}\Phi(x_i,y_i)^{\text{T}}\B{\mu}-\nu+\frac{1}{\sqrt{n}}\overset{m}{\underset{i=1}{\sum}}\lambda^{(i)}|\mu^{(i)}|\\[0.1cm]
\mbox{s.t.}&\quad \B{\Phi}(x,\pun)\B{\mu}+\nu\V{1}\in\set{L},\quad  \forall x\in\set{X}\end{split}\end{align}
where
$$\frac{1}{\sqrt{n}}\sum_{i=1}^m\lambda^{(i)}|\mu^{(i)}|=\frac{\lambda}{\sqrt{n}}\|\B{\mu}\|_1$$
if $\lambda^{(i)}=\lambda$ for $i=1,2,\ldots,m$.
\end{corollary}
\begin{IEEEproof}
See Appendix~\ref{apd:proof_cor_2}.
\end{IEEEproof}
\begin{remark}	
Similarly to regularized maximum entropy approach for density estimation \cite{DudPhiSha:07,DudPhi:04}, the above result shows that the usage of uncertainty sets $\set{U}^{\V{a},\V{b}}$ with $\V{a}\neq\V{b}$ corresponds to L1-norm regularization, which results on sparse parametrization of \acp{MRC}. L1-norm regularization is broadly used in machine learning (see e.g., \cite{MehRos:18,DemVitRos:09}) and the regularization parameter used to weight the L1-norm is commonly obtained by cross-validation methods. The expressions in Corollary~\ref{cor-reg} can directly provide appropriate regularization parameters from the length of the expectations' interval estimates. 

The above Corollary also shows that feature mapping components with poorly estimated expectations, i.e., \mbox{$i\in\{1,2,\ldots,m\}$} with large $\lambda^{(i)}$, have a reduced or null influence on the \ac{MRC} since the coefficients corresponding to such components are highly penalized at learning. In addition, the fact that \acp{MRC} are described by a sparse parametrization can be exploited in practice by considering feature mappings with numerous components at learning and then using only the components with non-zero parameters at prediction.
\end{remark}
\subsection{Performance guarantees}\label{sec-gua}
\vspace{-0.cm}
We next develop techniques that obtain tight bounds for the $\ell$-risk of an \ac{MRC}. In particular, a lower bound for the $\ell$-risk of $\up{h}^{\V{a},\V{b}}$ can be obtained by solving the \ac{LP}
\begin{align}\label{opt-bound}
\begin{split}\underline{\mathscr{P}}_{\;\!\ell}^{\V{a},\V{b}}:\quad \underset{\B{\mu},\B{\eta},\nu}{\max} 
&\quad \frac{1}{2}(\V{b}+\V{a})^{\text{T}}\B{\mu}-\frac{1}{2}(\V{b}-\V{a})^{\text{T}}\B{\eta}+\nu\\
\mbox{s.t.}&\quad \Phi(x,y)^{\text{T}}\B{\mu}+\nu\leq \ell(\up{h}^{\V{a},\V{b}},(x,y)),\\&\qquad \qquad \qquad \qquad \qquad
\forall x\in\set{X},y\in\set{Y}\\ &\quad 
\B{\eta}+\B{\mu}\succeq\V{0},\;\B{\eta}-\B{\mu}\succeq\V{0}.\end{split}\end{align}
\begin{theorem}\label{th2}
Let  $\V{a},\V{b}
\in\mathbb{R}^m$ satisfy $\set{U}^{\V{a},\V{b}}\neq~\emptyset$, $\up{h}^{\V{a},\V{b}}$ be an $\ell$-\ac{MRC} for $\set{U}^{\V{a},\V{b}}
$, and $\underline{R}^{\,\V{a},\V{b}}_{\;\!\ell}$ be the maximum value of $\underline{\mathscr{P}}_{\;\!\ell}^{\V{a},\V{b}}$. If the uncertainty set $\set{U}^{\V{a},\V{b}}$ includes the true underlying distribution, i.e., $\up{p}^*\in\set{U}^{\V{a},\V{b}}$, then 
$$\underline{R}^{\,\V{a},\V{b}}_{\;\!\ell}\leq R_\ell(\up{h}^{\V{a},\V{b}})\leq H_\ell(\set{U}^{\V{a},\V{b}}).$$
In addition, $R_\ell(\up{h}^{\V{a},\V{b}})=H_\ell(\set{U}^{\V{a},\V{b}})$ if $\up{p}^*$ maximizes the \mbox{$\ell$-entropy} over $\set{U}^{\V{a},\V{b}}$, and $R_\ell(\up{h}^{\V{a},\V{b}})=\underline{R}^{\,\V{a},\V{b}}_{\;\!\ell}$ if $\up{p}^*$ minimizes the expected $\ell$-loss of $\up{h}^{\V{a},\V{b}}$ over $\set{U}^{\V{a},\V{b}}$.
\end{theorem}
\begin{IEEEproof}
	See Appendix~\ref{apd:proof_th_2}.
	\end{IEEEproof}

The above result enables to obtain tight performance bounds for $\ell$-\acp{MRC} from training samples. The upper bound given by the maximum entropy $H_\ell(\set{U}^{\V{a},\V{b}})$ is directly obtained at \acp{MRC}' learning as given by \eqref{max-ent}. On the other hand, the lower bound $\underline{R}^{\,\V{a},\V{b}}_{\;\!\ell}$ requires to solve an additional \ac{LP} given by $\underline{\mathscr{P}}_{\;\!\ell}^{\V{a},\V{b}}$. The next theorem provides finite-sample generalization bounds for $\ell$-\acp{MRC} in terms of the training size and the smallest maximum entropy.

The smallest maximum $\ell$-entropy over uncertainty sets $\set{U}^{\V{a},\V{b}}$ that include the true underlying distribution is the non-random constant $H_\ell(\set{U}^{\B{\tau}_\infty})$ with $\B{\tau}_\infty=\mathbb{E}_{\up{p}^*}\{\Phi\}$. Such result is a direct consequence of the fact that  $\set{U}^{\B{\tau}_\infty}\subseteq\set{U}^{\V{a},\V{b}}$ for any uncertainty set $\set{U}^{\V{a},\V{b}}$ that includes $\up{p}^*$. The smallest maximum entropy $H_\ell(\set{U}^{\B{\tau}_\infty})$ corresponds with the $\ell$-\ac{MRC} $\up{h}^{\B{\tau}_\infty}$ that would require an infinite number of training samples to exactly determine the features' actual expectation $\B{\tau}_\infty$.
\begin{theorem}\label{th3}
Let $\delta\in(0,1)$, $\B{\tau}_\infty=\mathbb{E}_{\up{p}^*}\{\Phi\}$, and $\B{\tau}_n$, $\V{a}_n$, and $\V{b}_n$ be point and interval estimates of $\B{\tau}_\infty$ obtained from training samples as given by \eqref{interval} with
\begin{align*}\B{\lambda}=&\,\,\V{d}\sqrt{\frac{\log m+\log\frac{2}{\delta}}{2}}\\d^{(i)}=&\,\,\max_{x\in\set{X},y\in\set{Y}} \Phi(x,y)^{(i)}-\min_{x\in\set{X},y\in\set{Y}}\Phi(x,y)^{(i)}\\&\,\,\mbox{for }i=1,2,\ldots,m.\end{align*}
If $\up{h}^{\V{a}_n,\V{b}_n}$ and $\up{h}^{\B{\tau}_n}$ are $\ell$-\acp{MRC} for $\set{U}^{\V{a}_n,\V{b}_n}$ and $\set{U}^{\B{\tau}_n}$, respectively. Then, with probability at least $1-\delta$ 
\begin{align}
R_\ell(\up{h}^{\V{a}_n,\V{b}_n})&\leq H_\ell(\set{U}^{\V{a}_n,\V{b}_n})\label{bound0}\\
R_\ell(\up{h}^{\V{a}_n,\V{b}_n})&\leq H_\ell(\set{U}^{\B{\tau}_\infty})+\frac{2\|\B{\lambda}\|_\infty\|\B{\mu}^*\|_1}{\sqrt{n}}\label{bound1}\\
R_\ell(\up{h}^{\B{\tau}_n})&\leq H_\ell(\set{U}^{\B{\tau}_n})+\frac{\|\B{\lambda}\|_\infty\|\B{\mu}^*\|_1}{\sqrt{n}}\label{bound2}\\
R_\ell(\up{h}^{\B{\tau}_n})&\leq H_\ell(\set{U}^{\B{\tau}_\infty})+\frac{\|\B{\lambda}\|_\infty\|\B{\mu}^*_n-\B{\mu}^*\|_1}{\sqrt{n}}\label{bound4}\end{align}
where $\B{\mu}^*$ and $\B{\mu}_n^*$ are solutions of $\mathscr{P}_{\ell}^{\V{a}}$ in \eqref{opt-eq} for $\V{a}=\B{\tau}_\infty$ and $\V{a}=\B{\tau}_n$, respectively.
\end{theorem}
\begin{IEEEproof}
	See Appendix~\ref{apd:proof_th_3}.
	\end{IEEEproof}	
	
Inequality \eqref{bound1} and inequality \eqref{bound4} bound the risk of $\ell$-\acp{MRC} w.r.t. the smallest maximum $\ell$-entropy $H_\ell(\set{U}^{\B{\tau}_\infty})$; and inequality \eqref{bound0} and inequality \eqref{bound2} bound the difference between the risk of $\ell$-\acp{MRC} and the corresponding maximum entropy $H_\ell(\set{U}^{\V{a}_n,\V{b}_n})$ and $H_\ell(\set{U}^{\B{\tau}_n})$, respectively. Bounds \eqref{bound1}, \eqref{bound2}, and \eqref{bound4} show differences that decrease as $O(1/\sqrt{n})$ with proportionality constants that depend on the confidence $\delta$, and other parameters describing the complexity of feature mapping $\Phi$ such as its dimensionality $m$, and the difference between its maximum and minimum values $\V{d}$.
\section{\acp{MRC} for $\alpha$-entropies\\ with performance guarantees}\label{sec-0-1-log}
This section specializes the results above for $\alpha$-\acp{MRC}, and details the implementation of 0-1-\acp{MRC} and log-\acp{MRC}.
\begin{corollary}\label{cor-alpha}
For $\alpha$-loss $\ell_{\alpha}$ with $\alpha\in(0,1)\cup(1,\infty)$, the maximum entropy dual problem $\mathscr{P}_{\ell}^{\V{a},\V{b}}$ in \eqref{opt-prob} is equivalent to
\begin{align}\label{opt-prob-alpha}\begin{split}\mathscr{P}_{\alpha}^{\V{a},\V{b}}:\quad \underset{\B{\mu},\B{\eta},\nu}{\min}&\quad\frac{1}{2}(\V{b}-\V{a})^{\text{T}}\B{\eta}-\frac{1}{2}(\V{b}+\V{a})^{\text{T}}\B{\mu}-\nu\\
\mbox{s.t.}&\quad \underset{y\in\set{Y}}{\sum}\big(\frac{\Phi(x,y)^{\text{T}}\B{\mu}+\nu}{\beta}+1\big)_+^{\beta}\leq 1,\quad  \forall x\in\set{X} 
\\&\quad
\B{\eta}+\B{\mu}\succeq\V{0},\;\B{\eta}-\B{\mu}\succeq\V{0}\end{split}\end{align}
with $\beta=\alpha/(\alpha-1)$.\footnote{In expressions $(\pun)_+^\beta$, the positive part is computed before the power.}
In addition, for $\alpha$-loss the condition \eqref{opt-sol} characterizing \acp{MRC} becomes
\begin{align}\label{opt-sol-alpha}\up{h}^{\V{a},\V{b}}(y|x)\geq \left(\frac{\Phi(x,y)^{\text{T}}\B{\mu}^*+\nu^*}{\beta}+1\right)_+^{\beta}\end{align}
for $\B{\mu}^*,\B{\eta}^*,\nu^*$ a solution of $\mathscr{P}_{\alpha}^{\V{a},\V{b}}$.
\end{corollary}
\begin{IEEEproof}
The equivalence between optimization problem $\mathscr{P}_{\ell}^{\V{a},\V{b}}$ and $\mathscr{P}_{\alpha}^{\V{a},\V{b}}$ is obtained because for $\alpha$-loss we have that $\B{\Phi}(x,\pun)\B{\mu}+~\nu\V{1}\in\set{L}$
if it exists $\up{q}\in\Delta(\set{Y})$ such that $\forall y\in\set{Y}$
\begin{align*}&\Phi(x,y)^{\text{T}}\B{\mu}+\nu+\frac{\alpha}{\alpha-1}(1-\up{q}(y)^{\frac{\alpha-1}{\alpha}})\leq 0\\
\Leftrightarrow\quad  &\Phi(x,y)^{\text{T}}\B{\mu}+\nu+\frac{\alpha}{\alpha-1}\leq\frac{\alpha}{\alpha-1} \up{q}(y)^{\frac{\alpha-1}{\alpha}}\end{align*}
that happens if and only if 
$$\sum_{y\in\set{Y}}\Big(\frac{\alpha-1}{\alpha}(\Phi(x,y)^{\text{T}}\B{\mu}+\nu)+1\Big)_+^{\frac{\alpha}{\alpha-1}}\leq 1.$$
Analogous derivations show that condition \eqref{opt-sol} in Theorem~\ref{th1} leads condition \eqref{opt-sol-alpha} for $\alpha$-loss.
\end{IEEEproof}

We next specify Theorem~\ref{th1} for 0-1-entropy and log-entropy and show how it leads to the practical implementation of the learning and prediction stages for 0-1-\acp{MRC} and log-\acp{MRC}.
\begin{corollary}\label{cor-0-1}
For 0-1-loss $\ell_{0\text{-}1}$, the maximum entropy dual problem $\mathscr{P}_{\ell}^{\V{a},\V{b}}$ in \eqref{opt-prob} is equivalent to
\begin{align}\label{opt-prob-0-1}
\begin{split}\mathscr{P}_{0\text{-}1}^{\V{a},\V{b}}:\quad\underset{\B{\mu},\B{\eta},\nu}{\min}&\quad \frac{1}{2}(\V{b}-\V{a})^{\text{T}}\B{\eta}-\frac{1}{2}(\V{b}+\V{a})^{\text{T}}\B{\mu}-\nu\\
\mbox{s.t.}&\quad\underset{y\in\set{Y}}{\sum}(\Phi(x,y)^{\text{T}}\B{\mu}+\nu+1)_+\leq 1,\  \forall x\in\set{X}
\\&\quad
\B{\eta}+\B{\mu}\succeq\V{0},\B{\eta}-\B{\mu}\succeq\V{0}.\end{split}\end{align}
In addition, for 0-1-loss the condition \eqref{opt-sol} characterizing \acp{MRC} becomes
\begin{align}\label{opt-sol-0-1}\up{h}^{\V{a},\V{b}}(y|x)\geq \Phi(x,y)^{\text{T}}\B{\mu}^*+\nu^*+1\end{align}
for $\B{\mu}^*,\B{\eta}^*,\nu^*$ a solution of $\mathscr{P}_{0\text{-}1}^{\V{a},\V{b}}$.
\end{corollary}
\begin{IEEEproof}
Analogous to that of Corollary~\ref{cor-alpha}.
\end{IEEEproof}
\begin{corollary}\label{cor-log}
For log-loss $\ell_{\log}$, the maximum entropy dual problem $\mathscr{P}_{\ell}^{\V{a},\V{b}}$ in \eqref{opt-prob} is equivalent to
\begin{align}\label{opt-prob-log}\begin{split}\mathscr{P}_{\log}^{\V{a},\V{b}}:\quad\underset{\B{\mu},\B{\eta},\nu}{\min}&\quad\frac{1}{2}(\V{b}-\V{a})^{\text{T}}\B{\eta}-\frac{1}{2}(\V{b}+\V{a})^{\text{T}}\B{\mu}-\nu\\
\mbox{s.t.}&\quad\underset{y\in\set{Y}}{\sum}\exp\{\Phi(x,y)^{\text{T}}\B{\mu}+\nu\}\leq 1,\  \forall x\in\set{X}\\&\quad
\B{\eta}+\B{\mu}\succeq\V{0},\;\B{\eta}-\B{\mu}\succeq\V{0}.\end{split}\end{align}
In addition, for log-loss the condition \eqref{opt-sol} characterizing \acp{MRC} becomes
\begin{align}\label{opt-sol-log}\up{h}^{\V{a},\V{b}}(y|x)\geq \exp\{\Phi(x,y)^{\text{T}}\B{\mu}^*+\nu^*\}\end{align}
for $\B{\mu}^*,\B{\eta}^*,\nu^*$ a solution of $\mathscr{P}_{\log}^{\V{a},\V{b}}$.
\end{corollary}
\begin{IEEEproof}
Analogous to that of Corollary~\ref{cor-alpha}.
\end{IEEEproof}
\vspace{-0.cm}
\subsection{Implementation of 0-1-\acp{MRC} and log-\acp{MRC}}\label{sec-implementation}
\vspace{-0.cm}
Training samples are used at learning to obtain expectations' estimates $\V{a}_n$, $\V{b}_n$ and parameters $\B{\mu}^*$, $\nu^*$ that determine $\ell$-\acp{MRC} $\up{h}^{\V{a}_n,\V{b}_n}$. For 0-1-loss, parameters $\B{\mu}^*$, $\nu^*$ are used at prediction to obtain \begin{align}\label{h-0-1}\up{h}^{\V{a}_n,\V{b}_n}(y|x)=\left\{\begin{array}{cc}(\Phi(x,y)^{\text{T}}\B{\mu}^*+\nu^*+1)_+/c_x&\mbox{if }c_x\neq 0\\[3pt]
1/|\set{Y}|&\mbox{if }c_x= 0\end{array}\right.\end{align}
that satisfies \eqref{opt-sol-0-1} by taking $$c_x=\sum_{i\in\set{Y}}(\Phi(x,i)^{\text{T}}\B{\mu}^*+\nu^*+1)_+.$$
For log-loss, parameters $\B{\mu}^*$, $\nu^*$ are used at prediction to obtain
\begin{align}\label{h-log}\up{h}^{\V{a}_n,\V{b}_n}(y|x)&=\frac{\exp\{\Phi(x,y)^{\text{T}}\B{\mu}^*+\nu^*\}}{\sum_{i\in\set{Y}}\exp\{\Phi(x,i)^{\text{T}}\B{\mu}^*+\nu^*\}}\nonumber\\
&=\Big(\sum_{i\in\set{Y}}\exp\{(\Phi(x,i)-\Phi(x,y))^{\text{T}}\B{\mu}^*\}\Big)^{-1}\end{align}
that satisfies \eqref{opt-sol-log}. 

The constraints in the optimization problems solved at learning are given by the range of values of the feature mapping. We denote the range of values of $\Phi(x,y)\in\mathbb{R}^m$ for $y\in\set{Y}$ and varying $x\in\set{X}$ using $r$ vectors $\V{f}_{1}(y),\V{f}_{2}(y),\ldots,\V{f}_{r}(y)\in\mathbb{R}^m$ such that 
\begin{align}\label{M-matrices}\{\V{f}_{j}(y):\  j=1,2,\ldots,r\}=\{\Phi(x,y):\ x\in\set{X}\}\end{align}
for $y\in\set{Y}$.

The next result shows that the optimization problems solved at learning for 0-1-\acp{MRC} and log-\acp{MRC} are equivalent to \ac{LP} and \ac{GP} problems, respectively.
\begin{proposition}\label{prop-equivalence}
Let $\V{a}_n$ and $\V{b}_n$ be interval expectations' estimates of a feature mapping $\Phi$ obtained from $n$ training samples.
The maximum 0-1-entropy dual problem $\mathscr{P}_{0\text{-}1}^{\V{a},\V{b}}$ in \eqref{opt-prob-0-1} is equivalent to the \ac{LP}
\begin{align}\label{opt-prob-0-1-LP}\begin{split}\underset{\B{\mu},\B{\eta},\nu}{\min}&\quad\frac{1}{2}(\V{b}_n-\V{a}_n)^{\text{T}}\B{\eta}-\frac{1}{2}(\V{b}_n+\V{a}_n)^{\text{T}}\B{\mu}-\nu\\
\mbox{s.t.}&\quad\underset{y\in\set{C}}\sum\big(\V{f}_j(y)^{\text{T}}\B{\mu}+\nu+1\big)\leq 1\\
		&\qquad\  \  \,\forall j\in\{1,2,\ldots,r\}, \set{C}\subseteq\set{Y},\set{C}\neq\emptyset\\&
\B{\eta}+\B{\mu}\succeq\V{0},\;\B{\eta}-\B{\mu}\succeq\V{0}.\end{split}\end{align}
In addition, the maximum log-entropy dual problem $\mathscr{P}_{\log}^{\V{a},\V{b}}$ in \eqref{opt-prob-log} is equivalent to the \ac{GP}
\begin{align}\begin{split}\label{opt-prob-log-GP}\underset{\B{\mu},\B{\eta},\nu}{\min}&\quad \frac{1}{2}(\V{b}_n-\V{a}_n)^{\text{T}}\B{\eta}-\frac{1}{2}(\V{b}_n+\V{a}_n)^{\text{T}}\B{\mu}-\nu\\
\mbox{s.t.}&\quad \log\underset{y\in\set{Y}}{\sum}\exp\{\V{f}_j(y)^{\text{T}}\B{\mu}+\nu\}\leq 0\\
&\qquad \qquad \qquad \qquad \  \  \   \forall j\in\{1,2,\ldots,r\}\\&\quad
\B{\eta}+\B{\mu}\succeq\V{0},\B{\eta}-\B{\mu}\succeq\V{0}.\end{split}\end{align}
\end{proposition}
\begin{IEEEproof}
The equivalence between optimization problems $\mathscr{P}_{0\text{-}1}^{\V{a},\V{b}}$ in \eqref{opt-prob-0-1} and that in \eqref{opt-prob-0-1-LP} is due to the fact that 
\begin{align*}&\sum_{y\in\set{Y}}(\Phi(x,y)^{\text{T}}\B{\mu}+\nu+1)_+\leq 1,\  \forall x\in\set{X}\nonumber\\
\Leftrightarrow\;& \max_{\set{C}\subseteq\set{Y}} \sum_{y\in\set{C}}(\Phi(x,y)^{\text{T}}\B{\mu}+\nu+1)\leq 1,\  \forall x\in\set{X}\\
\Leftrightarrow\; &\sum_{y\in\set{C}}\big(\V{f}_j(y)^{\text{T}}\B{\mu}+\nu+1\big) \leq 1\nonumber\\
		&\hspace{1.5cm}  \forall j\in\{1,2,\ldots,r\}, \set{C}\subseteq\set{Y},\set{C}\neq\emptyset.\nonumber
\end{align*}
The equivalence between optimization problem $\mathscr{P}_{\log}^{\V{a},\V{b}}$ in \eqref{opt-prob-log} and that in \eqref{opt-prob-log-GP} is straightforward.
\end{IEEEproof}

The values of the above optimization problems (maximum entropies) provide upper bounds for the risk of 0-1-\acp{MRC} and log-\acp{MRC}. Lower bounds for such risks can be obtained by solving $\underline{\mathscr{P}}_{\;\!\ell}^{\V{a},\V{b}}$ as given by Theorem~\ref{th2}. The next result shows an equivalent formulation of such optimization problem as an \ac{LP}.
\begin{proposition}\label{prop2}
Let $\B{\mu}^*$ and $\nu^*$ be the parameters determining a 0-1-\ac{MRC} or a log-\ac{MRC}. Optimization problem $\underline{\mathscr{P}}_{\;\!\ell}^{\V{a},\V{b}}$ in \eqref{opt-bound} is equivalent to the \ac{LP}
\begin{align}\begin{split}\label{opt-prob-bound-LP}\underset{\B{\mu},\B{\eta},\nu}{\max}&\quad \frac{1}{2}(\V{b}_n+\V{a}_n)^{\text{T}}\B{\mu}-\frac{1}{2}(\V{b}_n-\V{a}_n)^{\text{T}}\B{\eta}+\nu\\
\mbox{s.t.}&\quad \V{f}_j(y)^{\text{T}}\B{\mu}+\nu\leq \varepsilon_j(y),\  \forall j\in\{1,2,\ldots,r\},\ y\in\set{Y}\\
&\quad \B{\eta}+\B{\mu}\succeq\V{0},\B{\eta}-\B{\mu}\succeq\V{0}.\end{split}\end{align}
where for 0-1-loss
\begin{align}\label{epsilon-0-1}\varepsilon_j(y)=\left\{\begin{array}{cc}1-(\V{f}_j(y)^{\text{T}}\B{\mu}^*+\nu^*+1)_+/c_j&\mbox{if }c_j\neq 0\\
1-1/|\set{Y}|&\mbox{if }c_j= 0\end{array}\right.\end{align}
with $c_j=\sum_{i\in\set{Y}}(\V{f}_j(i)^{\text{T}}\B{\mu}^*+\nu^*+1)_+$, and for log-loss
\begin{align}\label{epsilon-log}\varepsilon_j(y)=\log\sum_{i\in\set{Y}}\exp\{\V{f}_j(i)^{\text{T}}\B{\mu}^*\}-\V{f}_j(y)^{\text{T}}\B{\mu}^*.\end{align}
\end{proposition}
\begin{IEEEproof}
Straightforward consequence of the definition of vectors $\V{f}_i(y)$ in \eqref{M-matrices} and classification rules $\up{h}^{\V{a}_n,\V{b}_n}$ in \eqref{h-0-1} and \eqref{h-log}.  
\end{IEEEproof}
\begin{algorithm}
\setstretch{1}
		\caption{\label{codes-train}--  \ac{MRC} learning}	
 \small
 \begin{tabular}{ll}\hspace{-0.1cm}\textbf{Input:}&\hspace{-0.3cm}Training samples $(x_1,y_1),(x_2,y_2),\ldots,(x_n,y_n)$\\
&\hspace{-0.3cm}Width of confidence intervals $\B{\lambda}$\\
&\hspace{-0.3cm}Feature mapping $\Phi$\\
\hspace{-0.1cm}\textbf{Output:}&\hspace{-0.3cm}Parameters $\B{\mu}^*,\nu^*$\\
&\hspace{-0.3cm}Upper bound $H_{\ell}(\set{U}^{\V{a_n},\V{b_n}})$ \\
&\hspace{-0.3cm}[Optional] lower bound $\underline{R}_{\,\ell}^{\V{a_n},\V{b_n}}$
\end{tabular} 
\begin{algorithmic}[1] 
\setstretch{1.2}
\vspace{-0.1cm}
		\STATE $\B{\tau}_n\gets \frac{1}{n}\sum_{i=1}^n\Phi(x_i,y_i)$
		\STATE  $\V{a}_n\gets \B{\tau}_n-\B{\lambda}\frac{1}{\sqrt{n}}$
		\STATE  $\V{b}_n\gets \B{\tau}_n+\B{\lambda}\frac{1}{\sqrt{n}}$
		 \STATE$\B{\mu}^*,\B{\eta}^*,\nu^*\gets$ solution of \eqref{opt-prob-0-1-LP} for 0-1-loss, and that of  \eqref{opt-prob-log-GP} for log-loss
		\STATE $H_{\ell}(\set{U}^{\V{a_n},\V{b_n}})\gets\frac{1}{2}(\V{b}_n-\V{a}_n)^{\text{T}}\B{\eta}^*-\frac{1}{2}(\V{b}_n+\V{a}_n)^{\text{T}}\B{\mu}^*-\nu^*$
%
%
%
\STATE \mbox{[Optional]} $\underline{R}^{\,\V{a_n},\V{b_n}}_{\,\ell}\gets$ value of \eqref{opt-prob-bound-LP} with 
$\varepsilon_j(y)$ given by \eqref{epsilon-0-1} for 0-1-loss and given by \eqref{epsilon-log} for log-loss




\end{algorithmic}\vspace{-.025cm}
%
\end{algorithm}
\begin{algorithm}[t]
\setstretch{1}
		\caption{\label{codes-0-1-pred}-- Prediction with a 0-1-\ac{MRC}}
\small
\begin{tabular}{ll} \textbf{Input:}&\hspace{-0.3cm}Parameters $\B{\mu}^*,\nu^*$ \\&\hspace{-0.3cm}Feature mapping $\Phi$ \\&\hspace{-0.3cm}Instance $x$\\
\textbf{Output:}&\hspace{-0.3cm}Predicted label $y$\\&\hspace{-0.3cm}Prediction probabilities $\V{h}$\end{tabular}
\begin{algorithmic}[1] 
\vspace{0.cm}
\setstretch{1.2}
\STATE $c\gets\underset{i\in\set{Y}}{\mathlarger{\sum}}(\Phi(x,i)^\text{T}\B{\mu}^*+\nu^*+1)_+$
\vspace{0.1cm}
\IF{$c\,=\,0$}
\FOR{$i=1,2,\ldots,|\set{Y}|$}
\STATE $\up{h}^{(i)}\gets1/|\set{Y}|$
\ENDFOR
\ELSE
\FOR{$i=1,2,\ldots,|\set{Y}|$}
\STATE  $\up{h}^{(i)}\gets(\Phi(x,i)^\text{T}\B{\mu}^*+\nu^*+1)_+/c$
\ENDFOR
\ENDIF
\STATE  $y\gets i\mbox{ with probability } \up{h}^{(i)}$
 \end{algorithmic}
		\vspace{-.025cm}
\end{algorithm}
\begin{algorithm}[t]
\setstretch{1}
		\caption{\label{codes-log-pred}-- Prediction with a log-\ac{MRC}}
\small
\begin{tabular}{ll} \textbf{Input:}&\hspace{-0.3cm}Parameters $\B{\mu}^*$ \\&\hspace{-0.3cm}Feature mapping $\Phi$ \\&\hspace{-0.3cm}Instance $x$\\
\textbf{Output:}&\hspace{-0.3cm}Predicted label $y$\\&\hspace{-0.3cm}Estimated conditional probabilities  $\V{h}$ \end{tabular}
\begin{algorithmic}[1] 
\vspace{0.cm}
\setstretch{1.2}
\vspace{0.1cm}
\FOR{$i=1,2,\ldots,|\set{Y}|$}
\STATE $v^{(i)}\gets\Phi(x,i)^\text{T}\B{\mu}^*$
\ENDFOR
\STATE  $y\gets \arg\max \V{v}$
\FOR{$i=1,2,\ldots,|\set{Y}|$}
\STATE $\up{h}^{(i)}\gets \left(\sum_{j=1}^{|\set{Y}|}\exp\{v^{(j)}-v^{(i)}\}\right)^{-1}$
\ENDFOR
 \end{algorithmic}
\end{algorithm}

Algorithm~\ref{codes-train} details the learning stage for \acp{MRC} with \mbox{0-1-loss} and log-loss, while Algorithms \ref{codes-0-1-pred} and \ref{codes-log-pred} detail the prediction stage for \acp{MRC} with 0-1-loss and log-loss, respectively. The main computational complexity of such algorithms is due to the possibly large number of constraints in the optimization problems solved at learning. Specifically, the \ac{LP} optimization problem for learning 0-1-\acp{MRC} given by \eqref{opt-prob-0-1-LP} has $2m+r(2^{|\set{Y}|}-1)$ constraints, the \ac{GP} optimization problem for learning log-\acp{MRC} given by \eqref{opt-prob-log-GP} has $2m+r$ constraints, and the \ac{LP} optimization problem for lower bounding the expected loss of $\ell$-\acp{MRC} given by \eqref{opt-prob-bound-LP} has $2m+r|\set{Y}|$ constraints for any loss.
The number $r$ of vectors $\V{f}_1(y),\V{f}_2(y),\ldots,\V{f}_r(y)$ that cover the range of values of the feature mapping determines the number of such constraints.  Efficient optimization can be achieved by using constraint generation techniques or approximations with a subset of constraints. In particular, in the numerical results of Section~\ref{sec-numerical} we use the $r\leq n$ vectors obtained from the $n$ training instances as
\begin{align}\label{f-exp}\{\V{f}_{j}(y),\  j=1,2,\ldots,r\}=\{\Phi(x_i,y),\  i=1,2,\ldots,n\}\end{align}
since they approximately describe the range of values of $\Phi(x,y)$ for varying $x\in\set{X}$.

\section{Generalized maximum entropy for supervised classification with fixed marginals}
In this section we consider uncertainty sets defined by both expectations' and marginals' constraints. 
The usage of these uncertainty sets with additional constraints can result in a simplified learning process at the expense of inferior performance guarantees. In addition, as we show next, multiple existing classification techniques can be considered as special cases of the presented framework using such uncertainty sets. 
\subsection{Uncertainty sets given by expectations' and marginals' constraints}
We consider uncertainty sets $\set{U}_x^{\V{a},\V{b}}$ with fixed instances' marginal that coincides with the instances' empirical marginal $\up{p}_n(x)$ of $n$ training samples $(x_1,y_1),(x_2,y_2),\ldots,(x_n,y_n)$,\footnote{The techniques developed in this section can be easily adjusted to use other instances' marginals, for example those based on unlabelled examples in semi-supervision scenarios.} that is,
\begin{align}\label{uncertainty-emp-x}\set{U}_x^{\V{a},\V{b}}=\{&\up{p}\in\Delta(\set{X}\times\set{Y}):\ \V{a}\preceq \mathbb{E}_\up{p}\{\Phi(x,y)\}\preceq \V{b}\nonumber\\&\mbox{and } \up{p}(x)=\up{p}_n(x)\}\subset\set{U}^{\V{a},\V{b}}.\end{align}
Besides, we consider uncertainty sets $\set{U}_y^{\V{a},\V{b}}$ with fixed labels' marginal that coincides with distribution on labels $\up{p}_0(y)$, that is, 
\begin{align}\label{uncertainty-emp-y}\set{U}_y^{\V{a},\V{b}}=\{&\up{p}\in\Delta(\set{X}\times\set{Y}):\ \V{a}\preceq \mathbb{E}_\up{p}\{\Phi(x,y)\}\preceq \V{b}\nonumber\\&\mbox{and } \up{p}(y)=\up{p}_0(y)\}\subset\set{U}^{\V{a},\V{b}}.\end{align}
Finally, we also consider uncertainty sets $\set{U}_{x,y}^{\V{a},\V{b}}$ with fixed intances' and labels' marginals that coincide with  $\up{p}_n(x)$ and $\up{p}_0(y)$, respectively, that is,
\begin{align}\label{uncertainty-emp-xy}\set{U}_{x,y}^{\V{a},\V{b}}=\set{U}_x^{\V{a},\V{b}}\cap\set{U}_y^{\V{a},\V{b}}\subset\set{U}^{\V{a},\V{b}}.\end{align}
All the uncertainty sets above are compact and convex due to the linearity of probability distributions' marginalization.

The true marginal distributions are often unknown so that uncertainty sets given by expectations' and marginals' constraints do not include the true underlying distribution $\up{p}^*$. Therefore, \acp{MRC} for such uncertainty sets do not possess performance guarantees such as those shown in Section~\ref{sec-gua}. Since the cardinality of $\set{X}$ is much larger than that of $\set{Y}$, it is more challenging to have access to accurate approximations for instances' marginals than for labels' marginals. However, as shown in the following, fixing the instances' marginal can lead to simpler learning processes. 
\subsection{Fixing the instances' marginal}
The following result shows how \acp{MRC} that use uncertainty sets $\set{U}_x^{\V{a},\V{b}}$ in \eqref{uncertainty-emp-x} can be determined by a function over instances and a linear combination of the feature mapping. The parameters characterizing these \acp{MRC} are obtained at learning by solving the maximum entropy dual problem
\begin{align}\begin{split}\label{opt-prob-emp-x}\mathscr{P}_{x,\ell}^{\V{a},\V{b}}:\;\underset{\B{\mu},\B{\eta}}{\min}&\quad\frac{1}{2}(\V{b}-\V{a})^{\text{T}}\B{\eta}-\frac{1}{2}(\V{b}+\V{a})^{\text{T}}\B{\mu}-\frac{1}{n}\overset{n}{\underset{i=1}{\sum}}\varphi_\ell(\B{\mu},x_i)\\
\mbox{s.t.}&\quad \B{\eta}+\B{\mu}\succeq\V{0},\B{\eta}-\B{\mu}\succeq\V{0}\end{split}\end{align}
where $\varphi_\ell(\B{\mu},x)$ is the concave function on $\B{\mu}$
\begin{align}\begin{split}\label{psi}\varphi_\ell(\B{\mu},x)=\underset{\nu}{\max}& \quad \nu\\
\mbox{s.t.}& \quad \B{\Phi}(x,\pun)\B{\mu}+\nu\V{1}\in \set{L}
\end{split}
\end{align}
and $\set{L}\subset\mathbb{R}^{|\set{Y}|}$ is the convex set given by \eqref{gamma}.
\begin{theorem}\label{th1-emp-x}
Let $\V{a},\V{b}\in\mathbb{R}^m$ satisfy $\set{U}_x^{\V{a},\V{b}}\neq\emptyset$ with $\up{p}_n(x)$ the empirical distribution of instances $x_1,x_2,\ldots,x_n$, and $\B{\mu}^*,\B{\eta}^*$ be a solution of $\mathscr{P}_{x,\ell}^{\V{a},\V{b}}$. If for any $x\in\set{X}$, $y\in\set{Y}$, a classification rule
$\up{h}^{\V{a},\V{b}}\in \text{T}(\set{X},\set{Y})$ satisfies
\begin{align}\label{opt-sol-emp-x}\ell(\up{h}^{\V{a},\V{b}},(x,y))+\Phi(x,y)^{\text{T}}\B{\mu}^*+\varphi_\ell(\B{\mu}^*,x)\leq 0\end{align}
then $\up{h}^{\V{a},\V{b}}$ is an $\ell$-\ac{MRC} for $\set{U}_x^{\V{a},\V{b}}$. 
In addition, the maximum $\ell$-entropy over $\set{U}_x^{\V{a},\V{b}}$ is 
\begin{align}H_\ell(\set{U}_x^{\V{a},\V{b}})=\frac{(\V{b}-\V{a})^{\text{T}}\B{\eta}^*}{2}-\frac{(\V{b}+\V{a})^{\text{T}}\B{\mu}^*}{2}- \sum_{i=1}^n\frac{\varphi_\ell(\B{\mu}^*,x_i)}{n}.\end{align}
\end{theorem}
\begin{IEEEproof}
	See Appendix~\ref{apd:proof_th_1-emp-x}.
	\end{IEEEproof}
	
Note that optimization problem $\mathscr{P}_{x,\ell}^{\V{a},\V{b}}$ is convex because $\set{L}$ is a convex set. In addition, a classification rule $\up{h}^{\V{a},\V{b}}$ satisfying \eqref{opt-sol-emp-x} always exists because for each $x\in\set{X}$, there exists $\up{h}^{\V{a},\V{b}}(\pun|x)\in\Delta(\set{Y})$ that verifies \eqref{opt-sol-emp-x} by definition of $\varphi_\ell$ and $\set{L}$. The parametric form of \acp{MRC} that use uncertainty sets with fixed instances' marginal determined by \eqref{opt-sol-emp-x} is similar to that of \acp{MRC} that use uncertainty sets with only expectations' constraints in \eqref{opt-sol}, where the constant term $\nu^*$ in \eqref{opt-sol} is substituted in \eqref{opt-sol-emp-x} by the term $\varphi_\ell(\B{\mu}^*,x)$ that varies over instances. 
 
Similarly to Corollary~\ref{cor-reg}, the usage of $\V{a}\neq\V{b}$ can be seen as imposing regularization on the model parameters as follows.
\begin{corollary}\label{cor-reg-emp}
Let $\V{a}_n,\V{b}_n\in\mathbb{R}^m$ be expectations' interval estimates obtained from training samples $(x_1,y_1),(x_2,y_2),\ldots,(x_n,y_n)$ as given by \eqref{interval}. Then,  the maximum entropy dual problem $\mathscr{P}_{x,\ell}^{\V{a},\V{b}}$ becomes
\begin{align}
\underset{\B{\mu}}{\min}&\quad-\frac{1}{n}\overset{n}{\underset{i=1}{\sum}}(\Phi(x_i,y_i)^{\text{T}}\B{\mu}+\varphi_\ell(\B{\mu},x_i))+\frac{1}{\sqrt{n}}\overset{m}{\underset{i=1}{\sum}}\lambda^{(i)}|\mu^{(i)}|\end{align}
where
$$\frac{1}{\sqrt{n}}\sum_{i=1}^m\lambda^{(i)}|\mu^{(i)}|=\frac{\lambda}{\sqrt{n}}\|\B{\mu}\|_1$$
if $\lambda^{(i)}=\lambda$ for $i=1,2,\ldots,m$.
\end{corollary}
\begin{IEEEproof}
Analogous to that of Corollary~\ref{cor-reg}.
	\end{IEEEproof}

The results above describe \acp{MRC} based on maximum \mbox{$\ell$-entropy} and uncertainty sets $\set{U}_x^{\V{a},\V{b}}$  for general losses, we next describe such classification rules for 0-1-loss and log-loss. 
\begin{corollary}\label{cor-0-1-emp}
For 0-1-loss $\ell_{0\text{-}1}$, the maximum entropy dual problem $\mathscr{P}_{x,\ell}^{\V{a},\V{b}}$ is equivalent to
\begin{align}\begin{split}\label{opt-prob-emp-0-1}\mathscr{P}_{x,0\text{-}1}^{\V{a},\V{b}}:\;\underset{\B{\mu},\B{\eta}}{\min}&\quad \frac{1}{2}(\V{b}-\V{a})^{\text{T}}\B{\eta}-\frac{1}{2}(\V{b}+\V{a})^{\text{T}}\B{\mu}\\&\quad -\frac{1}{n}\overset{n}{\underset{i=1}{\sum}}\varphi_{0\text{-}1}(\B{\mu},x_i)\\
\mbox{s.t.}&\quad\B{\eta}+\B{\mu}\succeq\V{0},\B{\eta}-\B{\mu}\succeq\V{0}\end{split}\end{align}
with 
\begin{align}\varphi_{0\text{-}1}(\B{\mu},x)=\underset{\set{C}\subseteq\set{Y},\set{C}\neq\emptyset}{\min}\frac{1-\sum_{y\in\set{C}}\big(\Phi(x,y)^{\text{T}}\B{\mu}+1\big)}{|\set{C}|}.\end{align}
In addition, for 0-1-loss the condition \eqref{opt-sol-emp-x} characterizing \acp{MRC} becomes
\begin{align}\up{h}^{\V{a},\V{b}}(y|x)\geq \Phi(x,y)^{\text{T}}\B{\mu}^*+1+\varphi_{0\text{-}1}(\B{\mu}^*,x)\end{align}
for $\B{\mu}^*,\B{\eta}^*$ a solution of $\mathscr{P}_{x,0\text{-}1}^{\V{a},\V{b}}$.
\end{corollary}
\begin{IEEEproof}
	See Appendix~\ref{apd:proof_cor_0-1_emp}.
	\end{IEEEproof}
\begin{corollary}\label{cor-log-emp}
For log-loss $\ell_{\log}$, the maximum entropy dual problem $\mathscr{P}_{x,\ell}^{\V{a},\V{b}}$ is equivalent to
\begin{align}\begin{split}\label{opt-prob-emp-log}\mathscr{P}_{x,\log}^{\V{a},\V{b}}:\;\underset{\B{\mu},\B{\eta}}{\min}&\quad\frac{1}{2}(\V{b}-\V{a})^{\text{T}}\B{\eta}-\frac{1}{2}(\V{b}+\V{a})^{\text{T}}\B{\mu}\\&\quad-\frac{1}{n}\overset{n}{\underset{i=1}{\sum}}\varphi_{\text{log}}(\B{\mu},x_i)\\
\mbox{s.t.}&\quad\B{\eta}+\B{\mu}\succeq\V{0},\B{\eta}-\B{\mu}\succeq\V{0}\end{split}\end{align}
with 
\begin{align}\varphi_{\text{log}}(\B{\mu},x)=-\log \sum_{y\in\set{Y}}\exp\{\Phi(x,y)^{\text{T}}\B{\mu}\}.\end{align}
In addition, for log-loss the condition \eqref{opt-sol-emp-x} characterizing \acp{MRC} becomes
\begin{align}\up{h}^{\V{a},\V{b}}(y|x)&= \exp\{\Phi(x,y)^{\text{T}}\B{\mu}^*+\varphi_{\text{log}}(\B{\mu}^*,x)\}\nonumber\\
&=1\big/\sum_{i\in\set{Y}}\exp\{(\Phi(x,i)-\Phi(x,y))^{\text{T}}\B{\mu}^*\}\end{align}
for $\B{\mu}^*,\B{\eta}^*$ a solution of $\mathscr{P}_{x,\log}^{\V{a},\V{b}}$.
\end{corollary}
\begin{IEEEproof}
	See Appendix~\ref{apd:proof_cor_log_emp}.
	\end{IEEEproof}
Analogous results can be obtained for \acp{MRC} with uncertainty sets $\set{U}_x^{\V{a},\V{b}}$ and general $\alpha$-losses, which we do not reproduce here for brevity. As is shown in Section~\ref{sec-relationship} below, the results in Corollary~\ref{cor-0-1-emp} correspond to the methods proposed in \cite{FatAnq:16,AsiXinBeh:15,FarTse:16} for 0-1-loss, while the results in Corollary~\ref{cor-log-emp} correspond to logistic regression method (see e.g., \cite{Bis:06,MehRos:18,CorKuz:15}) for log-loss.
\subsection{Fixing the labels' marginal}
The following result shows how \acp{MRC} that use uncertainty sets $\set{U}_y^{\V{a},\V{b}}$ in \eqref{uncertainty-emp-y} can be determined by a function over labels and a linear combination of the feature mapping. The parameters characterizing these \acp{MRC} are obtained at learning by solving the maximum entropy dual problem
\begin{align}\begin{split}\label{opt-prob-emp-y}\mathscr{P}_{y,\ell}^{\V{a},\V{b}}:\;\underset{\B{\mu},\B{\eta},\B{\nu}}{\min}&\quad \frac{1}{2}(\V{b}-\V{a})^{\text{T}}\B{\eta}-\frac{1}{2}(\V{b}+\V{a})^{\text{T}}\B{\mu}-\underset{y\in\set{Y}}\sum\up{p}_0(y)\nu^{(y)}\\
\mbox{s.t.}&\quad \B{\Phi}(x,\pun)\B{\mu}+\B{\nu}\in\set{L},\  \forall x\in\set{X}\\&\quad
\B{\eta}+\B{\mu}\succeq\V{0},\B{\eta}-\B{\mu}\succeq\V{0}\end{split}\end{align}
where $\set{L}\subset\mathbb{R}^{|\set{Y}|}$ is the convex set given by \eqref{gamma}.

\begin{theorem}\label{th1-emp-y}
Let $\V{a},\V{b}\in\mathbb{R}^m$ satisfy $\set{U}_y^{\V{a},\V{b}}\neq\emptyset$, and $\B{\mu}^*,\B{\eta}^*,\B{\nu}^*$ be a solution of $\mathscr{P}_{y,\ell}^{\V{a},\V{b}}$. If for any $x\in\set{X}$, $y\in\set{Y}$, a classification rule $\up{h}^{\V{a},\V{b}}\in \text{T}(\set{X},\set{Y})$ satisfies
\begin{align}\label{opt-sol-emp-y}\ell(\up{h}^{\V{a},\V{b}},(x,y))+\Phi(x,y)^{\text{T}}\B{\mu}^*+(\nu^{*})^{(y)}\leq 0 \end{align}
then $\up{h}^{\V{a},\V{b}}$ is an $\ell$-\ac{MRC} for $\set{U}_y^{\V{a},\V{b}}$. 
In addition, the maximum $\ell$-entropy over $\set{U}_y^{\V{a},\V{b}}$ is 
\begin{align}H_\ell&(\set{U}_y^{\V{a},\V{b}})=\frac{(\V{b}-\V{a})^\text{T}\B{\eta}}{2}-\frac{(\V{b}+\V{a})^\text{T}\B{\mu}^*}{2}-\sum_{y\in\set{Y}}\up{p}_0(y)(\nu^*)^{(y)}.\end{align}
\end{theorem}
\begin{IEEEproof}
	See Appendix~\ref{apd:proof_th_1-emp-y}.
	\end{IEEEproof}
		\begin{table*}
\caption{Existing techniques for supervised classification that can be considered as special cases of the proposed framework}
\small
\centering
\def\arraystretch{1.3}
\begin{tabular}{lll}\toprule
Technique & Entropy function & Uncertainty set\\\hline\hline
Zero-one adversarial \cite{AsiXinBeh:15,FatAnq:16}& 0-1-entropy& $\set{U}_x^{\V{a}}$, $\B{\lambda}=\V{0}$\\
Logistic regression e.g., \cite{Bis:06,BerPie:96}&log-entropy (Shannon) & $\set{U}_x^{\V{a}}$, $\B{\lambda}=\V{0}$\\
Maximum entropy machine \cite{FarTse:16}& 0-1-entropy&$\set{U}_x^{\V{a},\V{b}}$, $\B{\lambda}\succ\V{0}$\\
Regularized logistic regression \cite{MehRos:18,CorKuz:15}&log-entropy (Shannon) &$\set{U}_x^{\V{a},\V{b}}$, $\B{\lambda}\succ\V{0}$\\
Minimum mutual information \cite{GloTis:04}& log-relative entropy (Mutual inf.) &$\set{U}_y^{\V{a}}$, $\B{\lambda}=\V{0}$\\
\bottomrule\label{tab-existing}\end{tabular}
\end{table*}
	
Note that the expressions above for the case of using uncertainty sets $\set{U}_y^{\V{a},\V{b}}$ are similar to those for the case of uncertainty sets $\set{U}^{\V{a},\V{b}}$ in Theorem~\ref{th1}. The scalar parameter $\nu$ in Theorem~\ref{th1} is substituted in the result above by vector parameter $\B{\nu}\in\mathbb{R}^{|\set{Y}|}$ that accounts for the fixed labels' probabilities given by $p_0(y)$. Analogous results as those in Corollaries \ref{cor-reg}, \ref{cor-alpha}, \ref{cor-0-1}, and \ref{cor-log} can be obtained using uncertainty sets $\set{U}_y^{\V{a},\V{b}}$, which we do not reproduce here for brevity.
\subsection{Fixing both instances' and labels' marginals}
The following result shows how \acp{MRC} that use uncertainty sets $\set{U}_{x,y}^{\V{a},\V{b}}$ in \eqref{uncertainty-emp-xy} can be determined by a function over instances and labels, and a linear combination of the feature mapping. The parameters describing those \acp{MRC} are obtained at learning by solving the maximum entropy dual problem
\begin{align}\begin{split}\label{opt-prob-emp-xy}
\mathscr{P}_{x,y,\ell}^{\V{a},\V{b}}:\;\;\underset{\B{\mu},\B{\eta},\B{\nu}}{\min}&\quad\frac{1}{2}(\V{b}-\V{a})^{\text{T}}\B{\eta}-\frac{1}{2}(\V{b}+\V{a})^{\text{T}}\B{\mu}\\&\quad
-\frac{1}{n}\overset{n}{\underset{i=1}{\sum}}\psi_\ell(\B{\mu},\B{\nu},x_i)-\underset{y\in\set{Y}}{\sum}p_0(y)\nu^{(y)}\\
\mbox{s.t.}&\quad \B{\eta}+\B{\mu}\succeq\V{0},\B{\eta}-\B{\mu}\succeq\V{0}
\end{split}\end{align}
where $\psi_\ell(\B{\mu},\B{\nu},x)$ is the concave function on $\B{\mu},\B{\nu}$
\begin{align}\label{psi}
\begin{split}
\psi_\ell(\B{\mu},\B{\nu},x)= \underset{\rho}{\max}&\quad \rho\\
\mbox{s.t.}& \quad \B{\Phi}(x,\pun)\B{\mu}+\B{\nu}+\rho\V{1}\in \set{L}
\end{split}
\end{align}
and $\set{L}\subset\mathbb{R}^{|\set{Y}|}$ is the convex set given by \eqref{gamma}.
\begin{theorem}\label{th1-emp-xy}
Let $\V{a},\V{b}\in\mathbb{R}^m$ satisfy $\set{U}_{xy}^{\V{a},\V{b}}\neq\emptyset$ with $\up{p}_n(x)$ the empirical distribution of instances $x_1,x_2,\ldots,x_n$, and $\B{\mu}^*,\B{\eta}^*,\B{\nu}^*$ be a solution of $\mathscr{P}_{x,y,\ell}^{\V{a},\V{b}}$. If for any $x\in\set{X}$, $y\in\set{Y}$, a classification rule $\up{h}^{\V{a},\V{b}}\in \text{T}(\set{X},\set{Y})$ satisfies
\begin{align}\label{opt-sol-emp-xy}\ell(\up{h}^{\V{a},\V{b}},(x,y))+\Phi(x,y)^{\text{T}}\B{\mu}^*+(\nu^*)^{(y)}+\psi_\ell(\B{\mu}^*,\B{\nu}^*,x)\leq0 \end{align}
then $\up{h}^{\V{a},\V{b}}$ is an $\ell$-\ac{MRC} for $\set{U}_{x,y}^{\V{a},\V{b}}$. 
In addition, the maximum $\ell$-entropy over $\set{U}_{x,y}^{\V{a},\V{b}}$ is 
\begin{align}H_\ell(\set{U}_{x,y}^{\V{a},\V{b}})&=\frac{(\V{b}-\V{a})^\text{T}\B{\eta}^*}{2}-\frac{(\V{b}+\V{a})^\text{T}\B{\mu}^*}{2}-\sum_{i=1}^n\frac{\psi_\ell(\B{\mu}^*,\B{\nu}^*,x_i)}{n}\nonumber\\
&\  \  \  - \sum_{y\in\set{Y}}p_0(y)(\nu^*)^{(y)}.\end{align}
\end{theorem}
\begin{IEEEproof}
	See Appendix~\ref{apd:proof_th_1-emp-xy}.
	\end{IEEEproof}
	
Note that, similarly to Theorem~\ref{th1-emp-x}, optimization problem $\mathscr{P}_{x,y,\ell}^{\V{a},\V{b}}$ is convex because $\set{L}$ is a convex set. In addition, a classification rule $\up{h}^{\V{a},\V{b}}$ satisfying \eqref{opt-sol-emp-xy} always exists by definition of $\psi_\ell$. The parametric form of \acp{MRC} that use uncertainty sets with fixed instances' and labels' marginals determined by \eqref{opt-sol-emp-xy} is similar to that of \acp{MRC} that use uncertainty sets with only expectations' constraints determined by \eqref{opt-sol}, where the constant term $\nu^*$ in \eqref{opt-sol} is substituted in \eqref{opt-sol-emp-xy} by the term $\nu^{*(y)}+\psi_\ell(\B{\mu}^*,\B{\nu}^*,x)$ that varies over instances and labels. 

Analogous results as those in Corollaries \ref{cor-reg}, \ref{cor-alpha}, \ref{cor-0-1}, and \ref{cor-log} can be obtained in this case, which we do not reproduce here for brevity. For example, for 0-1-loss and log-loss the expressions of function $\psi_\ell$ are
\begin{align*}\psi_{0\text{-}1}(\B{\mu},\B{\nu},x)&=\underset{\set{C}\subseteq\set{Y},\set{C}\neq\emptyset}{\min}\frac{1-\sum_{y\in\set{C}}\big(\Phi(x,y)^{\text{T}}\B{\mu}+\nu^{(y)}+1\big)}{|\set{C}|}\\
\psi_{\text{log}}(\B{\mu},\B{\nu},x)&=-\log \sum_{y\in\set{Y}}\exp\{\Phi(x,y)^{\text{T}}\B{\mu}+\nu^{(y)}\}.\end{align*}
\subsection{Relationship with existing techniques}\label{sec-relationship}
The results in previous sections can lead to a manifold of classification techniques by varying the entropy function and the uncertainty set considered. Most of such techniques are novel and some of them correspond to current methods. The novel techniques include those described in Sections~\ref{sec-3} and \ref{sec-0-1-log} that provide performance guarantees as well as those that use general $\alpha$-losses. The ones corresponding to existing techniques are obtained using uncertainty sets of distributions with fixed marginals. Specifically, Table~\ref{tab-existing} summarizes how multiple  existing techniques can be considered as special cases of the presented framework for different types of entropy functions and uncertainty sets.

The methods proposed in \cite{FatAnq:16,AsiXinBeh:15,FarTse:16} correspond to 
\acp{MRC} for 0-1-entropy and uncertainty sets $\set{U}^{\V{a}}_x$ with $\V{a}$ given by \eqref{interval} taking $\B{\lambda}=\V{0}$. 
Such correspondence is obtained because the maximum entropy dual problem $\mathscr{P}_{x,0\text{-}1}^{\V{a},\V{b}}$ in \eqref{opt-prob-emp-0-1} of Corollary~\ref{cor-0-1-emp} with $\B{\lambda}=\V{0}$ becomes
\begin{align}&\underset{\B{\mu}}{\min}\;\;\frac{1}{n}\overset{n}{\underset{i=1}{\sum}}-\Phi(x_i,y_i)^{\text{T}}\B{\mu}-\varphi_{0\text{-}1}(\B{\mu},x_i)\nonumber\\
=&\underset{\B{\mu}}{\min}\;\;\frac{1}{n}\overset{n}{\underset{i=1}{\sum}}\underset{\set{C}\subseteq\set{Y},\set{C}\neq\emptyset}{\max}\frac{\sum_{y\in\set{C}}(\Phi(x_i,y)-\Phi(x_i,y_i))^{\text{T}}\B{\mu}+|\set{C}|-1}{|\set{C}|}\label{opt-prob-emp-0-1-conventional}
\end{align}
that coincides with the empirical risk minimization of the loss function 
referred to as ``adversarial zero-one loss function'' in \cite{FatAnq:16,AsiXinBeh:15} and ``minimax-hinge loss function'' in \cite{FarTse:16}. In addition, the L1 reguralized approach using ``minimax-hinge loss function'' in \cite{FarTse:16} corresponds to \acp{MRC} for 0-1-entropy and uncertainty sets $\set{U}^{\V{a},\V{b}}_x$ with $\V{a}$ and $\V{b}$ given by \eqref{interval} taking $\B{\lambda}\succ\V{0}$, as a direct consequence of Corollaries \ref{cor-reg-emp} and \ref{cor-0-1-emp}.

Logistic regression method (see e.g., \cite{Bis:06,BerPie:96}) corresponds to \acp{MRC} for log-entropy and uncertainty sets $\set{U}^{\V{a}}_x$ with $\V{a}$ given by \eqref{interval} taking $\B{\lambda}=\V{0}$. Such correspondence is obtained because the maximum entropy dual problem $\mathscr{P}_{x,\log}^{\V{a},\V{b}}$ in \eqref{opt-prob-emp-log} of Corollary~\ref{cor-log-emp} with $\B{\lambda}=\V{0}$ becomes
\begin{align*}
&\underset{\B{\mu}}{\min}\;\;\frac{1}{n}\sum_{i=1}^n-\Phi(x_i,y_i)^{\text{T}}\B{\mu}-\varphi_{\text{log}}(\B{\mu},x_i)\\
=&\underset{\B{\mu}}{\min}\;\; \frac{1}{n}\overset{n}{\underset{i=1}{\sum}}\log\underset{y\in\set{Y}}{\sum}\exp\{\Phi(x_i,y)^{\text{T}}\B{\mu}\}-\Phi(x_i,y_i)^{\text{T}}\B{\mu}\\
=&\underset{\B{\mu}}{\min}\;\; \frac{1}{n}\overset{n}{\underset{i=1}{\sum}}-\log\left\{\frac{\exp\{\Phi(x_i,y_i)^{\text{T}}\B{\mu}\}}{\sum_{y\in\set{Y}}\exp\{\Phi(x_i,y)^{\text{T}}\B{\mu}\}}\right\}\\
=&\underset{\B{\mu}}{\min}\;\; \frac{1}{n}\overset{n}{\underset{i=1}{\sum}}\log\Big\{\underset{y\in\set{Y}}{\sum}\exp\left\{(\Phi(x_i,y)-\Phi(x_i,y_i))^{\text{T}}\B{\mu}\right\}\Big\}
\end{align*}
that coincides with the empirical risk minimization of the logistic loss function. In addition, the L1 reguralized approach for logistic regression in \cite{MehRos:18,CorKuz:15} corresponds to \acp{MRC} for log-entropy and uncertainty sets $\set{U}^{\V{a},\V{b}}_x$ with $\V{a}$ and $\V{b}$ given by \eqref{interval} taking $\B{\lambda}\succ\V{0}$, as a direct consequence of Corollaries \ref{cor-reg-emp} and \ref{cor-log-emp}.
\begin{table*}
\caption{Main results presented}\label{table-results}
\small
\centering
\begin{tabular}{llll}\toprule
&Results&Entropy function&Uncertainty set\vspace{0.1cm}\\\hline\hline\vspace{-0.25cm}\\
Theoretical framework&\hspace{-0.2cm}\begin{tabular}{l}Definitions \ref{def-entropy}, \ref{def-MRC}, Theorem~\ref{th0},\\ Corollary~\ref{cor0}\end{tabular}&General&General\vspace{0.1cm}\\\hline\vspace{-0.25cm}\\
\multirow{7}{*}{Learning techniques}&Theorem~\ref{th1}, Corollary~\ref{cor-reg}&General&Expectations' constraints\\
&Theorems \ref{th1-emp-x}, \ref{th1-emp-y},  \ref{th1-emp-xy}, Corollary~\ref{cor-reg-emp}&General&Expectations' and marginals' constraints\\
&Corollary~\ref{cor-alpha}&$\alpha$-entropy&Expectations' constraints\\
&Corollary~\ref{cor-0-1}, Propositions \ref{prop-equivalence}, \ref{prop2}&0-1-entropy&Expectations' constraints\\
&Corollary~\ref{cor-log}, Propositions \ref{prop-equivalence}, \ref{prop2}&log-entropy&Expectations' constraints\\
&Corollary~\ref{cor-0-1-emp}&0-1-entropy&Expectations' and marginals' constraints\\
&Corollary~\ref{cor-log-emp}&log-entropy&Expectations' and marginals' constraints\vspace{0.1cm}\\\hline\vspace{-0.25cm}\\
Performance guarantees&Theorems \ref{th2}, \ref{th3}&General&Expectations' constraints\vspace{0.1cm}\\\hline\vspace{-0.25cm}\\
\hspace{-0.2cm}\begin{tabular}{l}Relationship with\\ existing techniques\end{tabular}& Section~\ref{sec-relationship}&\hspace{-0.2cm}\begin{tabular}{l}0-1, log, and\\ relative-log \end{tabular}&Expectations' and marginals' constraints\vspace{0.1cm}\\
\bottomrule
\end{tabular}
\end{table*}

The minimum mutual information method proposed in \cite{GloTis:04} corresponds to \acp{MRC} for log-relative entropy with respect to $\up{p}_0(y)$ for uncertainty sets $\set{U}^{\V{a}}_y$ with $\V{a}$ given by \eqref{interval} taking $\B{\lambda}=\V{0}$. Such correspondence is obtained because the maximum entropy dual problem $\mathscr{P}_{y,\ell}^{\V{a},\V{b}}$ in \eqref{opt-prob-emp-y} for log-relative entropy with $\B{\lambda}=\V{0}$ ($\V{a}=\V{b}$) becomes
\begin{align*}
\begin{split}\underset{\B{\mu},\B{\nu}}{\min}&\quad -\V{a}^{\text{T}}\B{\mu}-\underset{y\in\set{Y}}{\sum}p_0(y)\nu^{(y)}\\
\mbox{s.t.}&\quad \log\underset{y\in\set{Y}}{\sum}p_0(y)\exp\{\Phi(x,y)^{\text{T}}\B{\mu}+\nu^{(y)}\}\leq 0,\  \forall x\in\set{X}\end{split}
\end{align*}
since, for log-relative loss with respect to $\up{p}_0(y)$, we have that $L(\up{q},y)=\log \frac{\up{p}_0(y)}{\up{q}(y)}$ so that the convex set $\set{L}$ is
$$\set{L}=\{\V{c}\in\mathbb{R}^{|\set{Y}|}:\  \sum_{y\in\set{Y}}\up{p}_0(y)\exp\{c^{(y)}\}\leq 1\}.$$
In addition, the derivation above does not require assumptions on strict positivity of instances' marginal, thus addressing the problem postulated in \cite{GloTis:04}.

\section{Discussion}
The sections above develop the generalized maximum entropy principle for supervised classification. The results establish a theoretical framework, develop learning techniques, present performance guarantees, and describe the relationship with existing techniques, as summarized in Table~\ref{table-results}. 

The theoretical framework shows that the maximum entropy principle for supervised classification leads to \acp{MRC} that minimize worst-case expected loss with respect to distributions in an uncertainty set. In addition, the maximum entropy provides a performance bound for \acp{MRC} in cases where the uncertainty set considered includes the true underlying distribution. 

The learning techniques allow to determine \acp{MRC} by means convex optimization for general entropy functions and uncertainty sets. \acp{MRC} minimize the worst-case expected loss over general classification rules and yet have a simple parametric expression determined by a linear combination of the feature mapping considered. \acp{MRC}' parameters are obtained at learning by solving a maximum entropy dual problem, and are used at prediction to classify general instances. Such maximum entropy dual problems are specified for general $\alpha$-entropies as well as for 0-1 and log entropies for which they are shown to be equivalent to \ac{LP} and \ac{GP} optimization problems using uncertainty sets given by expectations' constraints. In addition, the results show that the usage of expectations' interval estimates is equivalent to imposing L1 regularization at learning, and the regularization parameters are directly given by the accuracy of expectations' estimates.
  
The performance guarantees presented show that the usage of uncertainty sets given by expectations' constraints can enable to obtain tight upper and lower bounds for \acp{MRC}' risks. In addition, the results provide finite-sample generalization bounds for \acp{MRC} in terms of the smallest maximum entropy and the training size. Finally, we show that multiple existing classification techniques can be considered as special cases of the presented framework using uncertainty sets given by expectations' and marginals' constraints.

The general framework presented yields a manifold of classification techniques that accommodate to assorted goals and situations by varying the entropy function and type of uncertainty set. Entropies determined by classification-calibrated losses are preferred for pure-discriminative approaches, while entropies determined by proper losses are preferred for conditional approaches. The paper describes techniques corresponding with general $\alpha$-entropies including 0-1-entropy and log-entropy,  but other entropy functions can be similarly used. Additional convenient entropy functions are those given by losses that account for cost-sensitive classification errors \cite{AsiXinBeh:15} as well as those given by general losses developed for scoring probability assessments \cite{GneRaf:07}. In addition, entropy functions determined by relative losses can enable to use prior knowledge such as that related to prior label-probabilities. Regarding the choice of uncertainty set, those given by expectations' constraints can enable to obtain performance guarantees at learning, while those given by both expectations' and marginals' constraints can result in simplified learning processes. In addition, uncertainty sets given only by expectations' constraints enable to use training samples through statistical queries \cite{Kea:98} since the corresponding \acp{MRC} use expectations' estimates at learning. 
\begin{figure*}
\psfrag{Y0}[l][t][0.7]{\hspace{-3mm}0\text{-}1-Risk}
\psfrag{Yl}[l][t][0.7]{\hspace{-3mm}log-Risk}
\psfrag{X}[l][b][0.7]{\hspace{-10mm}Training size n}
\psfrag{A1234567890123456789012345678901}[l][][0.6]{\hspace{-21mm}\ac{MRC} 0-1-Risk}
\psfrag{B}[l][][0.6]{\hspace{-0.4mm}Upper bound $H_{0\text{-}1}(\set{U}^{\V{a}_n,\V{b}_n})$}
\psfrag{C}[l][][0.6]{\hspace{-0.4mm}Lower bound $\underline{R}^{\,\V{a_n},\V{b_n}}_{\;\!0\text{-}1}$}
\psfrag{D}[l][][0.6]{\hspace{-0.4mm}\ac{MEM} 0-1-Risk}
\psfrag{E}[l][][0.6]{\hspace{-0.4mm}\ac{SVM} 0-1-Risk}
\psfrag{100}[l][][0.5]{\hspace{-2mm}$100$}
\psfrag{500}[l][][0.5]{\hspace{-2mm}$500$}
\psfrag{1000}[l][][0.5]{\hspace{-4mm}$1000$}
\psfrag{5000}[l][][0.5]{\hspace{-4mm}$5000$}
\psfrag{0}[l][][0.5]{\hspace{-2mm}$0$}
\psfrag{0.05}[l][][0.5]{\hspace{-4mm}$0.05$}
\psfrag{0.02}[l][][0.5]{\hspace{-4mm}$0.02$}
\psfrag{0.04}[l][][0.5]{\hspace{-4mm}$0.04$}
\psfrag{0.06}[l][][0.5]{\hspace{-4mm}$0.06$}
\psfrag{0.1}[l][][0.5]{\hspace{-3mm}$0.1$}
\psfrag{0.15}[l][][0.5]{\hspace{-4mm}$0.15$}
\psfrag{0.2}[l][][0.5]{\hspace{-3mm}$0.2$}
\psfrag{0.25}[l][][0.5]{\hspace{-4mm}$0.25$}
\psfrag{0.3}[l][][0.5]{\hspace{-3mm}$0.3$}
\psfrag{0.4}[l][][0.5]{\hspace{-3mm}$0.4$}
\psfrag{0.5}[l][][0.5]{\hspace{-3mm}$0.5$}
\psfrag{0.6}[l][][0.5]{\hspace{-3mm}$0.6$}
\subfigure[0-1-risk. ]{\includegraphics[width=0.5\textwidth]{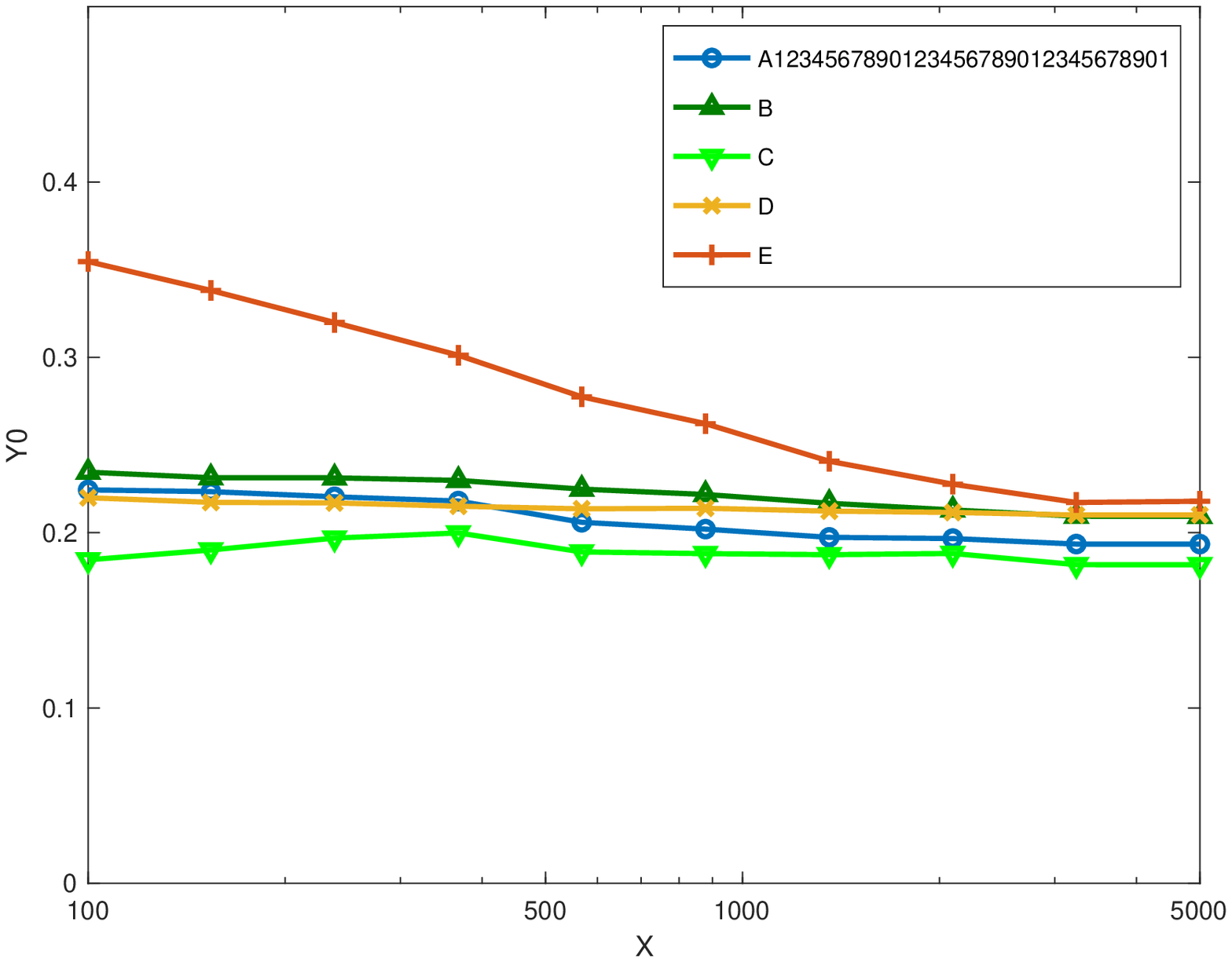}\label{fig_bounds_01_adult}}
\psfrag{A1234567890123456789012345678901}[l][][0.6]{\hspace{-21mm}\ac{MRC} log-Risk}
\psfrag{B}[l][][0.6]{\hspace{-0.4mm}Upper bound $H_{\log}(\set{U}^{\V{a}_n,\V{b}_n})$}
\psfrag{C}[l][][0.6]{\hspace{-0.4mm}Lower bound $\underline{R}^{\,\V{a_n},\V{b_n}}_{\;\!\log}$}
\psfrag{D}[l][][0.6]{\hspace{-0.4mm}\ac{LR} log-Risk}
\psfrag{E}[l][][0.6]{\hspace{-0.4mm}\ac{SVM} log-Risk}
\subfigure[log-risk. ]{\includegraphics[width=0.5\textwidth]{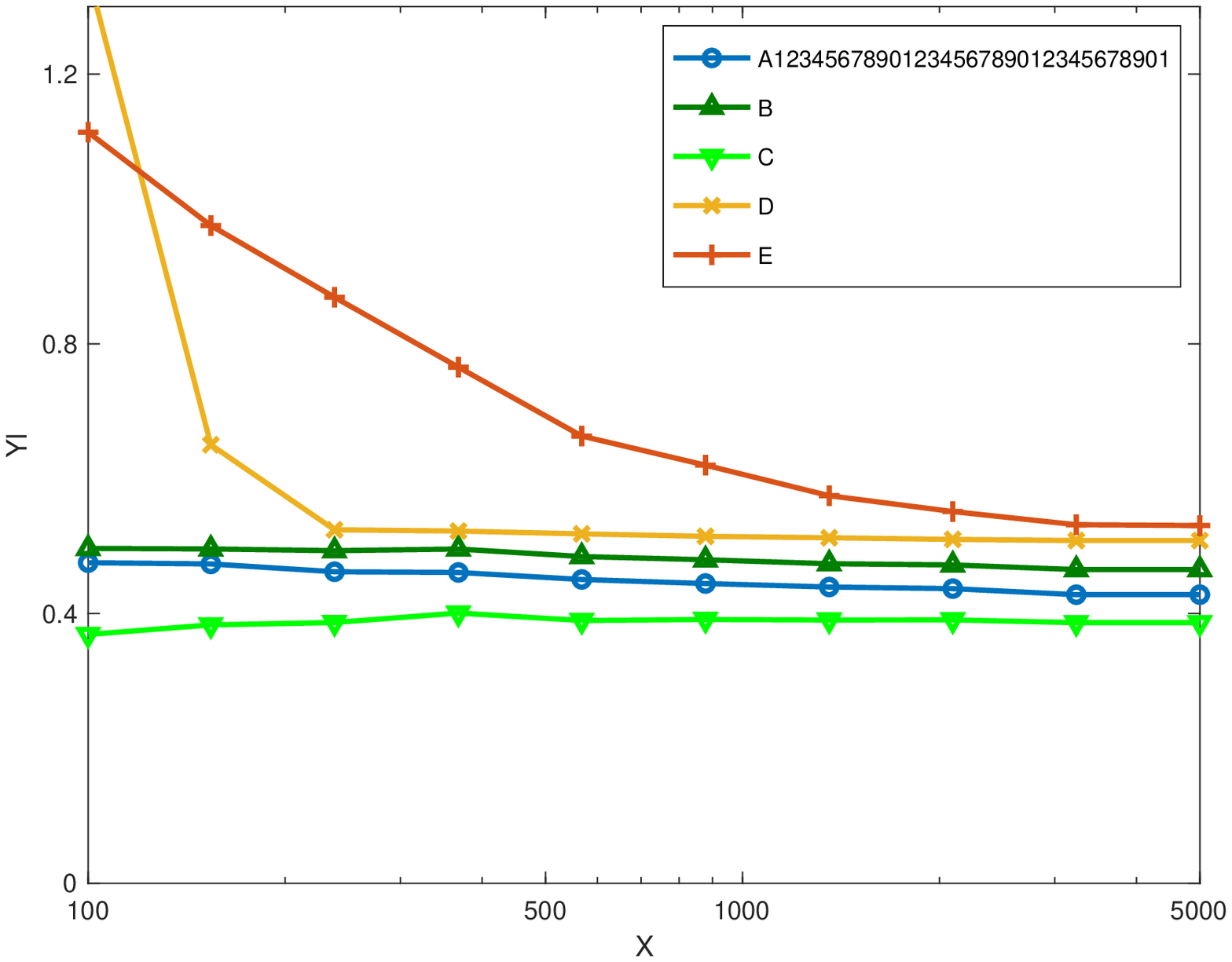}\label{fig_bounds_log_adult}}
	\vspace{-0.15cm}\caption{Risks of \acp{MRC} for ``Adult'' dataset in comparison with their upper and lower bounds and with conventional techniques.}\label{fig_bounds_adult}
	\vspace{-0.15cm}
\end{figure*} 
\vspace{-0.cm}
\section{Numerical results}\label{sec-numerical}
\vspace{-0.0cm}
In this section we show experimental results for \acp{MRC} using benchmark datasets, and compare their performance with the presented performance bounds and with conventional techniques for supervised classification. Specifically, we utilized ``Adult'', ``Pulsar'', and ``Thyroid'' datasets from UCI repository \cite{DuaGra:2019}. ``Adult'' dataset is formed by $48, 842$ samples with $15$-dimensional instances and two different labels, ``Pulsar'' dataset is formed by $17, 898$ samples with $9$-dimensional instances and two different labels, and ``Thyroid'' dataset is formed by $7,200$ samples with $22$-dimensional instances and three different labels.

\acp{MRC} for 0-1-entropy and log-entropy with expectations' constraints are implemented as described in Algorithms \ref{codes-train}, \ref{codes-0-1-pred}, and \ref{codes-log-pred}. \acp{MRC} results are obtained using feature mappings given by instances' thresholding, similarly to those used by maximum entropy and logistic regression methods \cite{MehRos:18,DudPhiSha:07,DudPhi:04,PhiAnd:06}. Such feature mappings are adequate for a streamlined implementation of \acp{MRC} because they take a reduced number of values.
\begin{figure*}
\psfrag{Y0}[l][t][0.7]{\hspace{-3mm}0\text{-}1-Risk}
\psfrag{Yl}[l][t][0.7]{\hspace{-3mm}log-Risk}
\psfrag{X}[l][b][0.7]{\hspace{-10mm}Training size n}
\psfrag{A1234567890123456789012345678901}[l][][0.6]{\hspace{-21mm}\ac{MRC} 0-1-Risk}
\psfrag{B}[l][][0.6]{\hspace{-0.4mm}Upper bound $H_{0\text{-}1}(\set{U}^{\V{a}_n,\V{b}_n})$}
\psfrag{C}[l][][0.6]{\hspace{-0.4mm}Lower bound $\underline{R}^{\,\V{a_n},\V{b_n}}_{\;\!0\text{-}1}$}
\psfrag{D}[l][][0.6]{\hspace{-0.4mm}\ac{MEM} 0-1-Risk}
\psfrag{E}[l][][0.6]{\hspace{-0.4mm}\ac{SVM} 0-1-Risk}
\psfrag{100}[l][][0.5]{\hspace{-2mm}$100$}
\psfrag{500}[l][][0.5]{\hspace{-2mm}$500$}
\psfrag{1000}[l][][0.5]{\hspace{-4mm}$1000$}
\psfrag{5000}[l][][0.5]{\hspace{-4mm}$5000$}
\psfrag{0}[l][][0.5]{\hspace{-2mm}$0$}
\psfrag{0.05}[l][][0.5]{\hspace{-4mm}$0.05$}
\psfrag{0.02}[l][][0.5]{\hspace{-4mm}$0.02$}
\psfrag{0.04}[l][][0.5]{\hspace{-4mm}$0.04$}
\psfrag{0.06}[l][][0.5]{\hspace{-4mm}$0.06$}
\psfrag{0.1}[l][][0.5]{\hspace{-3mm}$0.1$}
\psfrag{0.15}[l][][0.5]{\hspace{-4mm}$0.15$}
\psfrag{0.2}[l][][0.5]{\hspace{-3mm}$0.2$}
\psfrag{0.25}[l][][0.5]{\hspace{-4mm}$0.25$}
\psfrag{0.3}[l][][0.5]{\hspace{-3mm}$0.3$}
\psfrag{0.4}[l][][0.5]{\hspace{-3mm}$0.4$}
\psfrag{0.5}[l][][0.5]{\hspace{-3mm}$0.5$}
\psfrag{0.6}[l][][0.5]{\hspace{-3mm}$0.6$}
\subfigure[0-1-risk. ]{\includegraphics[width=0.5\textwidth]{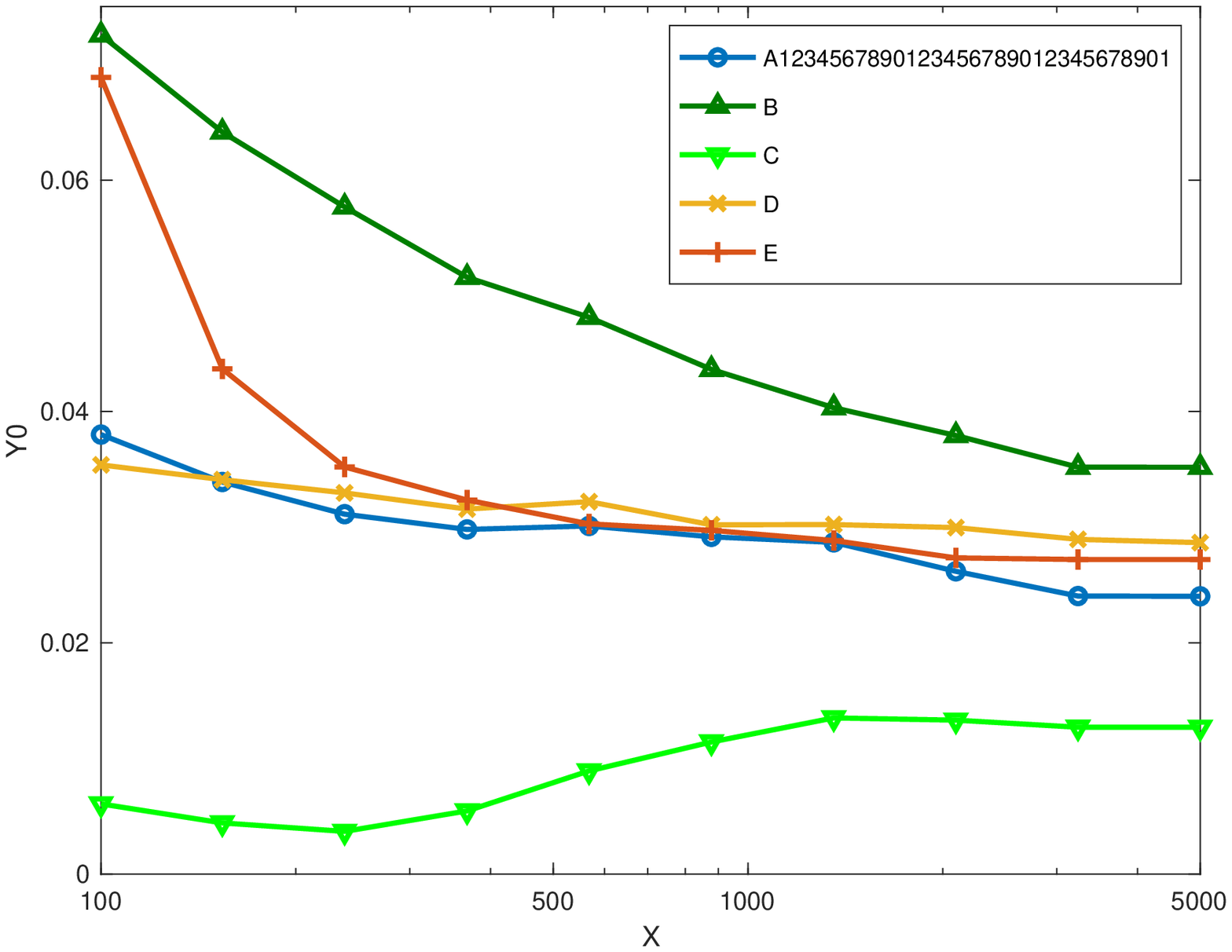}\label{fig_bounds_01_adult}}
\psfrag{A1234567890123456789012345678901}[l][][0.6]{\hspace{-21mm}\ac{MRC} log-Risk}
\psfrag{B}[l][][0.6]{\hspace{-0.4mm}Upper bound $H_{\log}(\set{U}^{\V{a}_n,\V{b}_n})$}
\psfrag{C}[l][][0.6]{\hspace{-0.4mm}Lower bound $\underline{R}^{\,\V{a_n},\V{b_n}}_{\;\!\log}$}
\psfrag{D}[l][][0.6]{\hspace{-0.4mm}\ac{LR} log-Risk}
\psfrag{E}[l][][0.6]{\hspace{-0.4mm}\ac{SVM} log-Risk}
\subfigure[log-risk. ]{\includegraphics[width=0.5\textwidth]{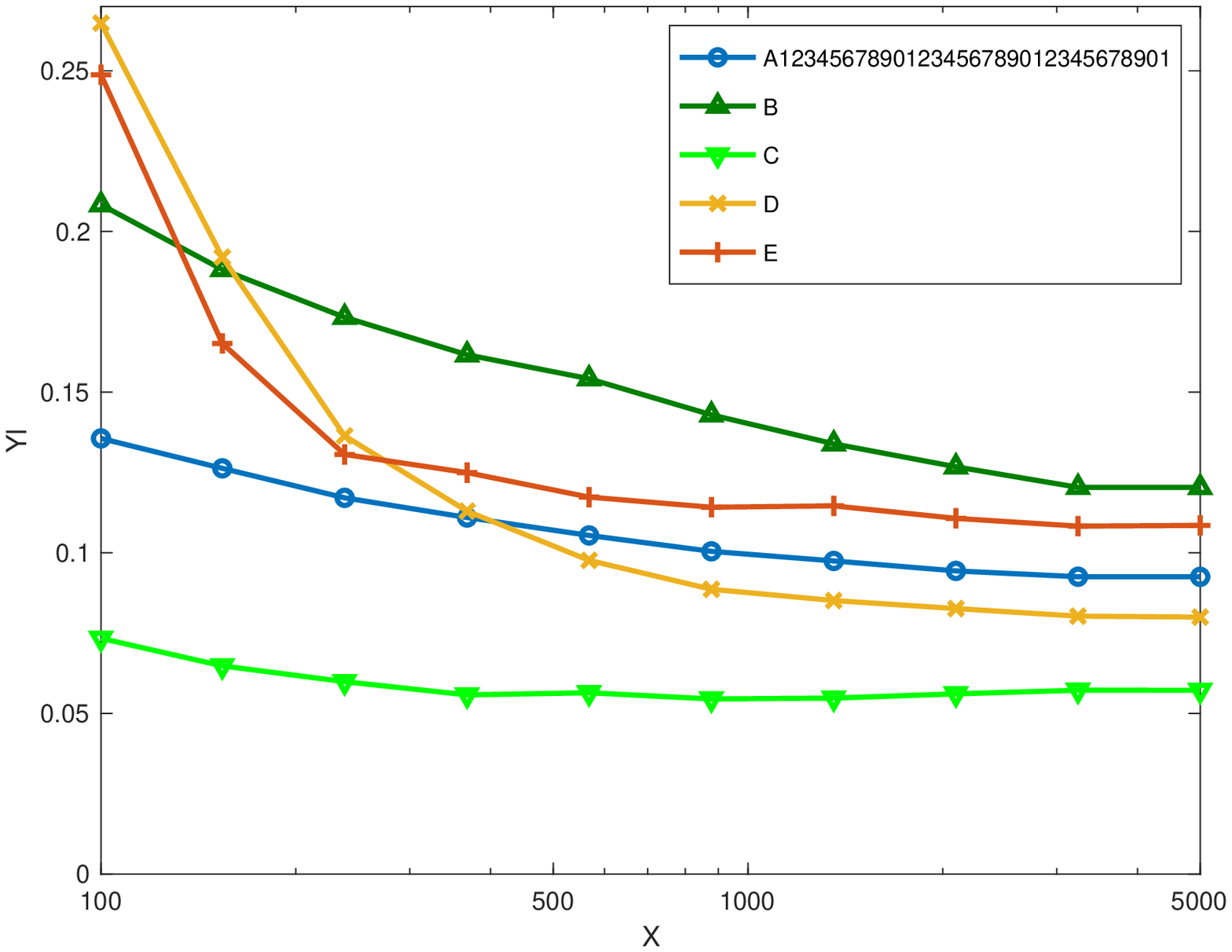}\label{fig_bounds_log_adult}}
	\caption{Risks of \acp{MRC} for ``Pulsar' dataset in comparison with their upper and lower bounds and with conventional techniques.}\label{fig_bounds_pulsar}
\end{figure*} 
\begin{figure*}
\psfrag{Y0}[l][t][0.7]{\hspace{-3mm}0\text{-}1-Risk}
\psfrag{Yl}[l][t][0.7]{\hspace{-3mm}log-Risk}
\psfrag{X}[l][b][0.7]{\hspace{-10mm}Training size n}
\psfrag{A1234567890123456789012345678901}[l][][0.6]{\hspace{-21mm}\ac{MRC} 0-1-Risk}
\psfrag{B}[l][][0.6]{\hspace{-0.4mm}Upper bound $H_{0\text{-}1}(\set{U}^{\V{a}_n,\V{b}_n})$}
\psfrag{C}[l][][0.6]{\hspace{-0.4mm}Lower bound $\underline{R}^{\,\V{a_n},\V{b_n}}_{\;\!0\text{-}1}$}
\psfrag{D}[l][][0.6]{\hspace{-0.4mm}\ac{MEM} 0-1-Risk}
\psfrag{E}[l][][0.6]{\hspace{-0.4mm}\ac{SVM} 0-1-Risk}
\psfrag{100}[l][][0.5]{\hspace{-2mm}$100$}
\psfrag{500}[l][][0.5]{\hspace{-2mm}$500$}
\psfrag{1000}[l][][0.5]{\hspace{-4mm}$1000$}
\psfrag{5000}[l][][0.5]{\hspace{-4mm}$5000$}
\psfrag{0}[l][][0.5]{\hspace{-2mm}$0$}
\psfrag{0.05}[l][][0.5]{\hspace{-4mm}$0.05$}
\psfrag{0.02}[l][][0.5]{\hspace{-4mm}$0.02$}
\psfrag{0.04}[l][][0.5]{\hspace{-4mm}$0.04$}
\psfrag{0.06}[l][][0.5]{\hspace{-4mm}$0.06$}
\psfrag{0.1}[l][][0.5]{\hspace{-3mm}$0.1$}
\psfrag{0.15}[l][][0.5]{\hspace{-4mm}$0.15$}
\psfrag{0.2}[l][][0.5]{\hspace{-3mm}$0.2$}
\psfrag{0.25}[l][][0.5]{\hspace{-4mm}$0.25$}
\psfrag{0.3}[l][][0.5]{\hspace{-3mm}$0.3$}
\psfrag{0.4}[l][][0.5]{\hspace{-3mm}$0.4$}
\psfrag{0.5}[l][][0.5]{\hspace{-3mm}$0.5$}
\psfrag{0.6}[l][][0.5]{\hspace{-3mm}$0.6$}
\subfigure[0-1-risk. ]{\includegraphics[width=0.5\textwidth]{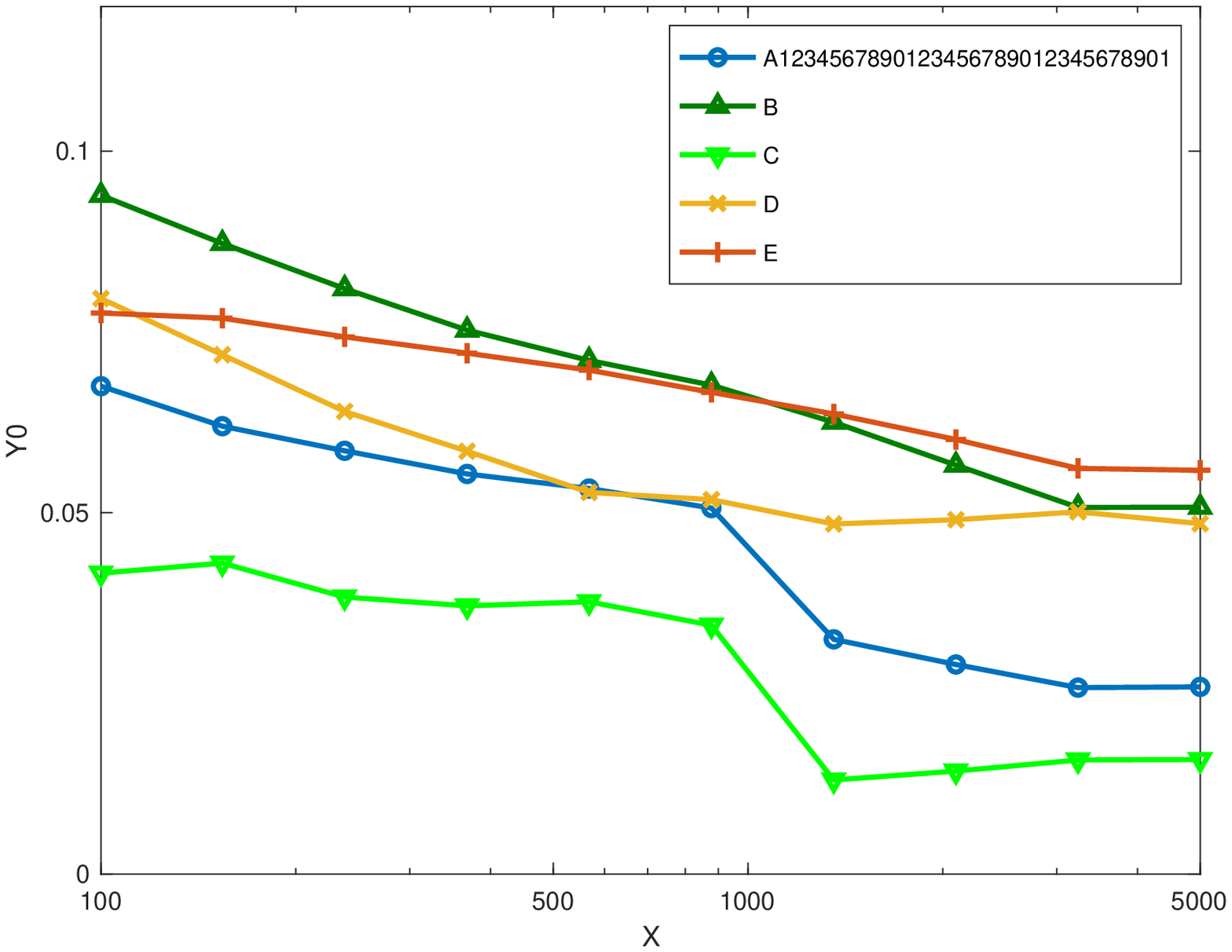}\label{fig_bounds_01_thyro}}
\psfrag{A1234567890123456789012345678901}[l][][0.6]{\hspace{-21mm}\ac{MRC} log-Risk}
\psfrag{B}[l][][0.6]{\hspace{-0.4mm}Upper bound $H_{\log}(\set{U}^{\V{a}_n,\V{b}_n})$}
\psfrag{C}[l][][0.6]{\hspace{-0.4mm}Lower bound $\underline{R}^{\,\V{a_n},\V{b_n}}_{\;\!\log}$}
\psfrag{D}[l][][0.6]{\hspace{-0.4mm}\ac{LR} log-Risk}
\psfrag{E}[l][][0.6]{\hspace{-0.4mm}\ac{SVM} log-Risk}
\subfigure[log-risk. ]{\includegraphics[width=0.5\textwidth]{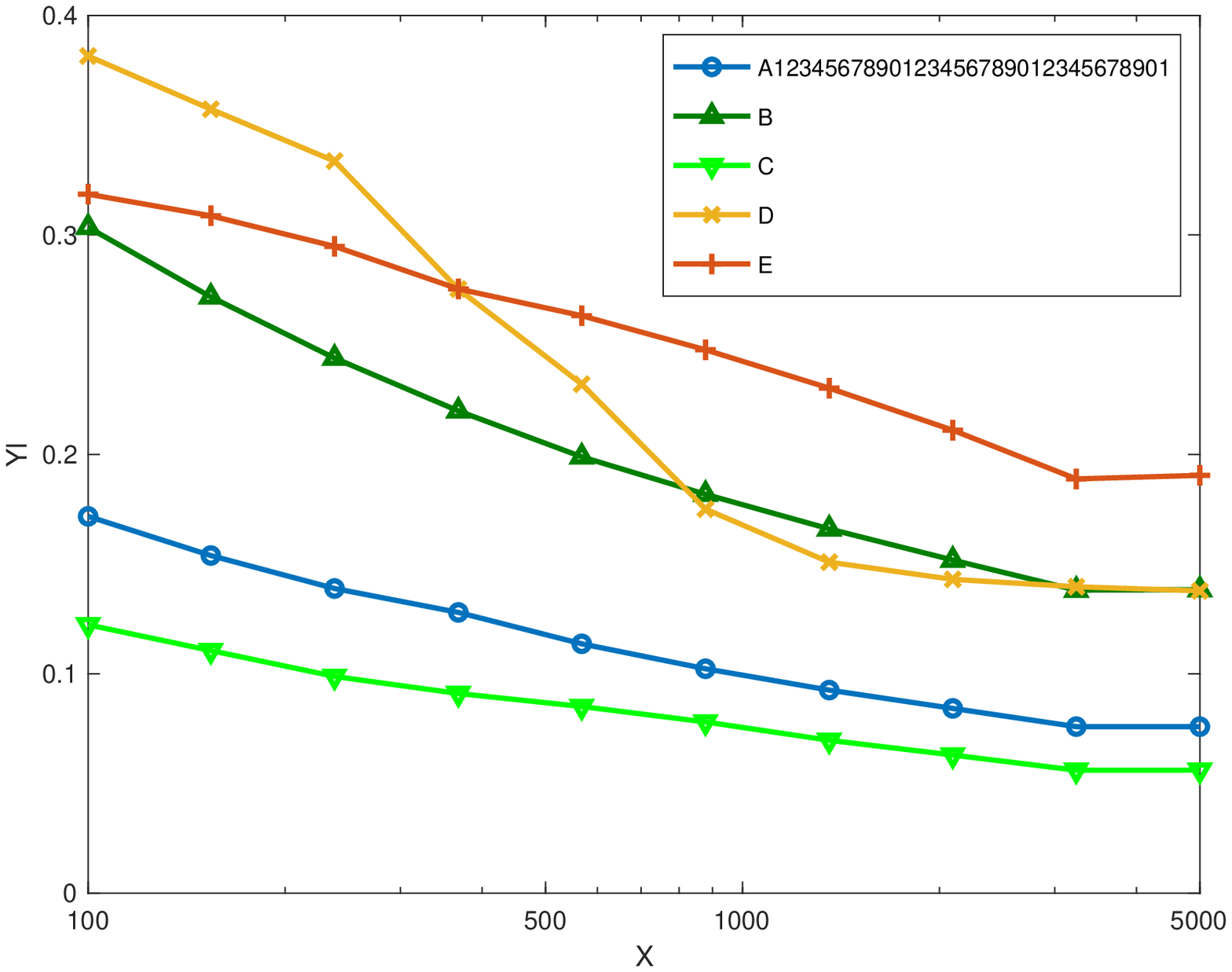}\label{fig_bounds_log_thryro}}
	\vspace{-0.cm}\caption{Risks of \acp{MRC} for ``Thyroid' dataset in comparison with their upper and lower bounds and with conventional techniques.}\label{fig_bounds_thyroid}
	\vspace{-0.cm}
\end{figure*} 

Let each instance $x\in\set{X}$ be given by $\V{x}=[x^{(1)},x^{(2)},\ldots,x^{(D)}]^{\text{T}}\in\mathbb{R}^D$, and for $i=1,2,\ldots,k$ let $\text{Th}_i\in\mathbb{R}$ be a threshold corresponding with dimension $d_i\in~\{1,2,\ldots,D\}$. We consider feature mappings as in \eqref{feature_map} with $m=|\set{Y}|(k+1)$ components corresponding to the different combinations of labels and thresholds. Specifically, 
\begin{align}\label{features}\Phi(x,y)=\V{e}_y\otimes \Psi(x),\  \Psi(x)=\left[\begin{array}{c}1\\\mathbb{I}\{x^{(d_1)}\leq\text{Th}_{1}\}\\\mathbb{I}\{x^{(d_2)}\leq\text{Th}_{2}\}\\\vdots\\\mathbb{I}\{x^{(d_k)}\leq\text{Th}_{k}\}\end{array}\right]\end{align}



Thresholds $\text{Th}_1,\text{Th}_2,\ldots,\text{Th}_k$ defining the feature mapping are obtained by one-dimensional decision trees (decision stumps) with up to $20$ leaves. In addition, the optimization problems for \acp{MRC} learning in Algorithm~\ref{codes-train} are solved using $r\leq n$ vectors $\V{f}_1,\V{f}_2,\ldots,\V{f}_r$ corresponding to the training instances as given by \eqref{f-exp}, and the corresponding \ac{LP} and \ac{GP} problems are solved using CVX package \cite{GraBoyYe:06}. 
\begin{figure*}
\psfrag{g1}[l][t][0.7]{\hspace{-3mm}Adult}
\psfrag{g2}[l][t][0.7]{\hspace{-3mm}Pulsar}
\psfrag{g3}[l][t][0.7]{\hspace{-4.5mm}Thyroid}
\psfrag{a}[l][b][0.7]{\hspace{-5mm}Upper}
\psfrag{b}[l][b][0.7]{\hspace{-5mm}Lower}
\psfrag{Y}[l][t][0.7]{\hspace{-5mm}0-1-Risk}
\psfrag{0}[c][][0.5]{\hspace{-3mm}$0$}
\psfrag{0.02}[c][][0.5]{\hspace{-0mm}$0.02$}
\psfrag{0.04}[c][][0.5]{\hspace{-0mm}$0.04$}
\psfrag{0.06}[c][][0.5]{\hspace{-0mm}$0.06$}
\psfrag{-0.02}[c][][0.5]{\hspace{-2mm}$-0.02$}
\subfigure[Differences between 0-1-\acp{MRC}' risks and performance bounds.]{\includegraphics[width=0.5\textwidth]{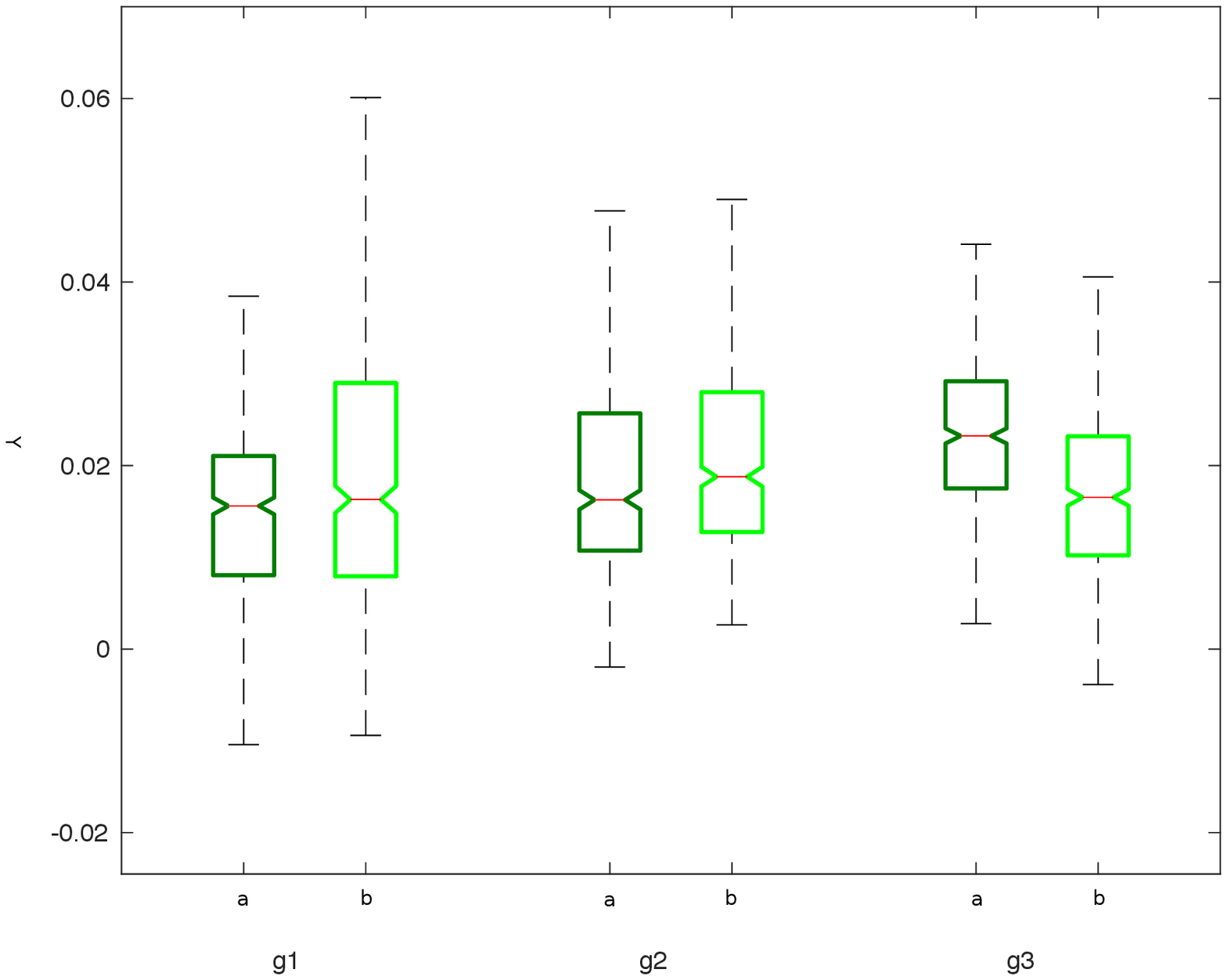}\label{box_0-1}}
\psfrag{Y}[l][t][0.7]{\hspace{-5mm}log-Risk}
\subfigure[Differences between log-\acp{MRC}' risks and performance bounds.]{\includegraphics[width=0.5\textwidth]{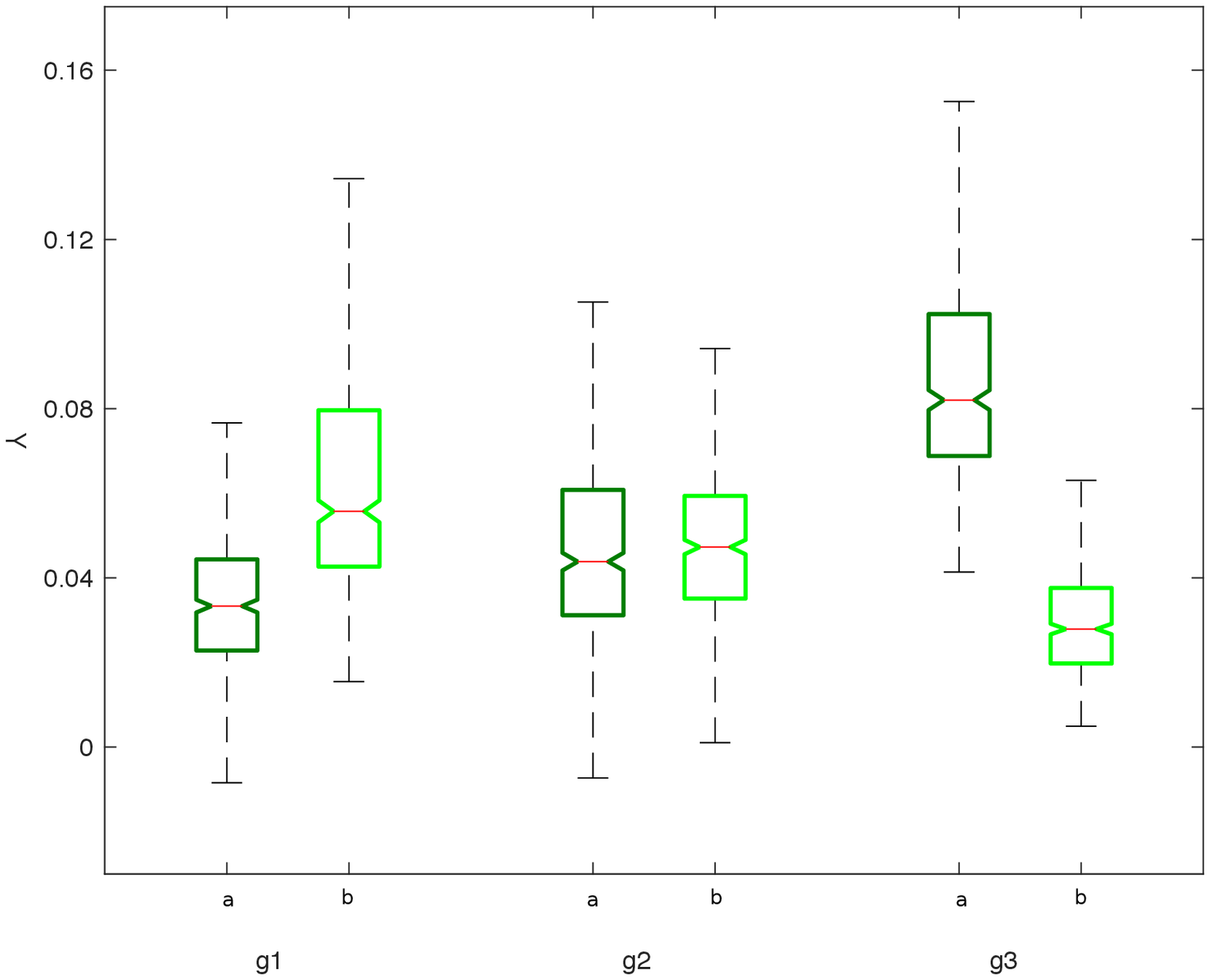}\label{box_log}}
	\caption{Tightness of the performance bounds obtained at training.}\label{fig_box}
\end{figure*}

For each training size $n$, random stratified instantiations of training samples are used to obtain \ac{SVM}, \ac{MEM}, and \ac{LR} classifiers, as well as 0-1-\acp{MRC} and log-\acp{MRC} together with their corresponding upper and lower performance bounds. Then, an additional set of $1000$ test samples is used to estimate the true risks. All classification techniques are implemented using their default parameters; \acp{SVM} are implemented using Gaussian kernels and their log-loss is obtained by estimating posterior class probabilities as presented in \cite{Plat:99}; and \acp{MRC} are implemented taking $\lambda^{(i)}=0.25$, for $i=1,2,\ldots,m$. \ac{MRC} techniques including \ac{MEM} have been implemented using the Python library \texttt{MRCpy} \cite{KarMazPer:21} \url{https://MachineLearningBCAM.github.io/MRCpy/}.


Figures \ref{fig_bounds_adult}, \ref{fig_bounds_pulsar}, and \ref{fig_bounds_thyroid} show the average risks and bounds obtained over $50$ random instantiations. Such figures show that \acp{MRC} can provide reliable performance bounds at learning and achieve classification performances competitive with the state-of-the-art. In addition, Fig.~\ref{fig_box} shows boxplots with the differences between the performance bounds and \acp{MRC}' risks over all instantiations and training sizes. Such figures show that the performance bounds obtained at learning are tight in every instantiation.




The numerical results corroborate over diverse datasets that \acp{MRC} can enable accurate classification with performance guarantees for varied entropy functions.

\vspace{0.4cm}
\section{Conclusion}
The paper presents a framework for supervised classification based on generalized maximum entropy. Such framework corresponds with classification techniques referred to as minimax risk classifiers (MRCs). 
We develop learning algorithms that determine \acp{MRC} for general entropy functions by means of convex optimization and show finite-sample generalization bounds for \acp{MRC} in terms of the training size. We also show that tight performance bounds can be obtained at learning by using uncertainty sets given by expectations' constraints. In addition, multiple existing classification techniques can be considered as special cases of the presented framework by using uncertainty sets given by expectations' and marginals' constraints.  The paper also shows that \acp{MRC} can obtain competitive performances and provide tight risk bounds at learning for assorted entropies. Varying the entropy function and type of uncertainty set used, the presented framework yields a manifold of classification techniques that can offer performance guarantees. Such flexible set of techniques includes multiple existing methods and can enable to design learning algorithms tailored to assorted goals and situations.

\appendices
\section{Proof of Theorem~\ref{th1}}\label{apd:proof_th_1}
\begin{IEEEproof}
Let $\widetilde{\set{U}}$ and $\widetilde{\ell}$ be the set and function
\begin{align*}
\widetilde{\set{U}}&=\{\up{p}:\set{X}\times\set{Y}\to[0,\infty)\mbox{ s.t. }  \sum_{y\in\set{Y}}\up{p}(x,y)\leq 1,\forall x\in\set{X}\}\\
\widetilde{\ell}(\up{h},\up{p})&=\frac{1}{2}(\V{b}-\V{a})^{\text{T}}\B{\eta}^*-\frac{1}{2}(\V{b}+\V{a})^{\text{T}}\B{\mu}^*-\nu^*+\ell(\up{h},\up{p})\\
&\quad +\sum_{x\in\set{X},y\in\set{Y}}\up{p}(x,y)(\Phi(x,y)^{\text{T}}\B{\mu}^*+\nu^*)\end{align*}
in the first step of the proof we show that a classification rule satisfying \eqref{opt-sol} is solution of optimization problem $\min_{\up{h}\in \text{T}(\set{X},\set{Y})}\max_{\up{p}\in\widetilde{\set{U}}}\widetilde{\ell}(\up{h},\up{p})$, and in the second step of the proof we show that a solution of $\min_{\up{h}\in \text{T}(\set{X},\set{Y})}\max_{\up{p}\in\widetilde{\set{U}}}\widetilde{\ell}(\up{h},\up{p})$ is also a solution of $\min_{\up{h}\in \text{T}(\set{X},\set{Y})}\max_{\up{p}\in\set{U}^{\V{a},\V{b}}}\ell(\up{h},\up{p})$.


For the first step, note that
\begin{align*}\widetilde{\ell}(\up{h},\up{p})&=\frac{1}{2}(\V{b}-\V{a})^{\text{T}}\B{\eta}^*-\frac{1}{2}(\V{b}+\V{a})^{\text{T}}\B{\mu}^*-\nu^*\\&\quad+\sum_{x\in\set{X}}g(\up{h}(\pun|x),\up{p}(x,\pun))\end{align*}
with 
\begin{align*}g(\up{h}(\pun|x),\up{p}(x,\pun))=&\sum_{y\in\set{Y}}\up{p}(x,y)L(\up{h}(\pun|x),y)\\
&+\sum_{y\in\set{Y}}\up{p}(x,y)[\Phi(x,y)^{\text{T}}\B{\mu}^*+\nu^*].\end{align*}
Then,
\begin{align*}
\min_{\up{h}\in \text{T}(\set{X},\set{Y})}\max_{\up{p}\in\widetilde{\set{U}}}\widetilde{\ell}(\up{h},\up{p})&=\frac{1}{2}(\V{b}-\V{a})^{\text{T}}\B{\eta}^*-\frac{1}{2}(\V{b}+\V{a})^{\text{T}}\B{\mu}^*-\nu^*\\
&\,\,+\min_{\up{h}\in \text{T}(\set{X},\set{Y})}\max_{\up{p}\in\widetilde{\set{U}}}\sum_{x\in\set{X}}g(\up{h}(\pun|x),\up{p}(x,\pun))\end{align*}
and
\begin{align*}
\min_{\up{h}\in \text{T}(\set{X},\set{Y})}\max_{\up{p}\in\widetilde{\set{U}}}&\sum_{x\in\set{X}}g(\up{h}(\pun|x),\up{p}(x,\pun))\\
=&\sum_{x\in\set{X}}\min_{\up{h}(\pun|x)\in \Delta(\set{Y})}\max_{\up{p}(x,\pun)\in\widetilde{\set{U}}_x}g(\up{h}(\pun|x),\up{p}(x,\pun))\end{align*}
with $\widetilde{\set{U}}_x=\{q:\set{Y}\to[0,\infty),\  \sum_{y\in\set{Y}}q(y)\leq 1\}$. The inner maximization above is given by
\begin{align}\label{max-g}\max_{\up{p}(x,\pun)\in\widetilde{\set{U}}_x}&g(\up{h}(\pun|x),\up{p}(x,\pun))\nonumber\\
&=\max_{y\in\set{Y}}\left(L(\up{h}(\pun|x),y)+\Phi(x,y)^{\text{T}}\B{\mu}^*+\nu^*\right)_+\end{align}
as a direct consequence of $g(\up{h}(\pun|x),\up{p}(x,\pun))$ definition. Then, a classification rule satisfying \eqref{opt-sol} is solution of optimization problem $\min_{\up{h}\in \text{T}(\set{X},\set{Y})}\max_{\up{p}\in\widetilde{\set{U}}}\widetilde{\ell}(\up{h},\up{p})$ since \eqref{max-g} takes its minimum value $0$ for $\up{h}^{\V{a},\V{b}}(\pun|x)$ such that for any $y\in\set{Y}$
 $$L(\up{h}^{\V{a},\V{b}}(\pun|x),y)+\Phi(x,y)^{\text{T}}\B{\mu}^*+\nu^*\leq0 $$
that exists due to the constraints of $\mathscr{P}_{\ell}^{\V{a},\V{b}}$ in \eqref{opt-prob}.

For the second step of the proof, if $\up{h}^{\V{a},\V{b}}$ is a solution of $\min_{\up{h}\in \text{T}(\set{X},\set{Y})}\max_{\up{p}\in\widetilde{\set{U}}}\widetilde{\ell}(\up{h},\up{p})$ we have that
\begin{align}\label{ineqs}\min_{\up{h}\in \text{T}(\set{X},\set{Y})}\max_{\up{p}\in\widetilde{\set{U}}}\widetilde{\ell}(\up{h},\up{p})&=\max_{\up{p}\in\widetilde{\set{U}}}\widetilde{\ell}(\up{h}^{\V{a},\V{b}},\up{p})\nonumber\\
\geq\max_{\up{p}\in\set{U}^{\V{a},\V{b}}}\ell(\up{h}^{\V{a},\V{b}},\up{p})&\geq\min_{\up{h}\in \text{T}(\set{X},\set{Y})}\max_{\up{p}\in\set{U}^{\V{a},\V{b}}}\ell(\up{h},\up{p})\end{align}
where the first inequality is due to the fact that $\set{U}^{\V{a},\V{b}}\subset\widetilde{\set{U}}$ and $\widetilde{\ell}(\up{h},\up{p})\geq\ell(\up{h},\up{p})$ for 
$\up{p}\in\set{U}^{\V{a},\V{b}}$ because
 $$\frac{1}{2}(\V{b}-\V{a})^{\text{T}}\B{\eta}^*-\frac{1}{2}(\V{b}+\V{a})^{\text{T}}\B{\mu}^*+\sum_{x\in\set{X},y\in\set{Y}}\up{p}(x,y)\Phi(x,y)^{\text{T}}\B{\mu}^*$$
is at least $0$ by definition of $\set{U}^{\V{a},\V{b}}$ and since $\B{\eta}^*+\B{\mu}^*\succeq \V{0},\B{\eta}^*-\B{\mu}^*\succeq \V{0}$. 

The min and the max in $\min_{\up{h}\in \text{T}(\set{X},\set{Y})}\max_{\up{p}\in\set{U}^{\V{a},\V{b}}}\ell(\up{h},\up{p})$ can be interchanged using classical minimax theorems (see e.g., \cite{Sim:95}) since
$\ell(\up{h},\up{p})$ is lower semi-continuous and convex in $\up{h}$ because $L$ is lower semi-continuous and convex in its first argument,  $\ell(\up{h},\up{p})$ is concave in $\up{p}$ because it is linear in $\up{p}$, and both $\set{U}^{\V{a},\V{b}}$ and $ \text{T}(\set{X},\set{Y})$ are convex and compact. Hence
$$\min_{\up{h}\in \text{T}(\set{X},\set{Y})}\max_{\up{p}\in\set{U}^{\V{a},\V{b}}}\ell(\up{h},\up{p})=\max_{\up{p}\in\set{U}^{\V{a},\V{b}}}H_\ell(\up{p})=H_\ell(\set{U}^{\V{a},\V{b}}).$$
Such maximum entropy problem is equivalent to
\begin{align}\label{max-ent-proof}\begin{array}{cl}\underset{\up{p}}{\max}&H_\ell(\up{p})-I_+(\up{p})\\
\mbox{s.t.}&-\V{1}^{\text{T}}\V{p}=-1\\
&\V{a}\preceq\B{\Phi}^{\text{T}}\V{p}\preceq\V{b}\end{array}\end{align}
where $\V{p}$ and $\B{\Phi}$ denote the vector and matrix with rows $\up{p}(x,y)$ and $\Phi(x,y)^{\text{T}}$, respectively, for $x\in\set{X}$, $y\in\set{Y}$, and $$I_+(\V{p})=\left\{\begin{array}{cc}0 &\mbox{if}\  \V{p}\succeq \V{0}\\\infty&\mbox{otherwise.}\end{array}\right.$$
The Lagrange dual of \eqref{max-ent-proof} is
\begin{align*}
\begin{split}\underset{\B{\mu}_a,\B{\mu}_b,\nu}{\min}&\quad \V{b}^{\text{T}}\B{\mu}_b-\V{a}^{\text{T}}\B{\mu}_a-\nu\\&\quad +(-H_\ell+I_+)^*(\B{\Phi}(\B{\mu}_a-\B{\mu}_b)+\nu\V{1})\\\mbox{s.t.}&\quad
\B{\mu}_a,\B{\mu}_b\succeq\V{0}\end{split}
\end{align*}
for $(-H_\ell+I_+)^*$ the conjugate function of $(-H_\ell+I_+)$, which is given by 
\begin{align*}(-H_\ell+I_+)^*(\V{v})=&\sum_{x\in\set{X}}\sup_{\up{p}(x,\pun)\succeq\V{0}}\big(\sum_{y\in\set{Y}}v(x,y)\up{p}(x,y)\\&+\inf_{\up{q}\in\Delta(\set{Y})}\sum_{y\in\set{Y}}\up{p}(x,y) L(\up{q},y)\big).\end{align*}
Let for $\V{u}\in\mathbb{R}^{|\set{Y}|}$
$$\chi(\V{u})=\left\{\begin{array}{cc}-\inf_{\up{q}\in\Delta(\set{Y})}\V{u}^{\text{T}} L(\up{q},\pun)&\mbox{if }\V{u}\succeq\V{0}\\
\infty&\mbox{otherwise}\end{array}\right.$$
that equals 
\begin{align}\label{chi}\chi(\V{u})=\sup_{\V{c}\in\set{L}}\V{u}^{\text{T}}\V{c}\end{align}
because $\set{L}$ in \eqref{gamma} is
$$\set{L}=\{-L(\up{q},\pun)-\V{r}\in\mathbb{R}^{|\set{Y}|}:\  \up{q}\in\Delta(\set{Y}), \V{r}\succeq\V{0}\}.$$
Then
$$(-H_\ell+I_+)^*(\V{v})=\sum_{x\in\set{X}}\chi^*(v(x,\pun))$$
and the conjugate function of $\chi$ is
\begin{align}\label{chi-conj}\chi^*(\V{w})=\left\{\begin{array}{cc}0&\mbox{if }\V{w}\in\set{L}\\
\infty&\mbox{otherwise}\end{array}\right.\end{align}
because $\chi$ is the support function of the convex set $\set{L}$. Therefore, the Lagrange dual of \eqref{max-ent-proof} becomes
$\mathscr{P}_{\ell}^{\V{a},\V{b}}$ in \eqref{opt-prob} taking $\B{\mu}=\B{\mu}_a-\B{\mu}_b$, $\B{\eta}=\B{\mu}_a+\B{\mu}_b$.
 
Finally, we have that 
\begin{align*}&\hspace{-0.35cm}\min_{\up{h}\in \text{T}(\set{X},\set{Y})}\max_{\up{p}\in\set{U}^{\V{a},\V{b}}}\ell(\up{h},\up{p})=\max_{\up{p}\in\set{U}^{\V{a},\V{b}}}\min_{\up{h}\in \text{T}(\set{X},\set{Y})}\ell(\up{h},\up{p})\\
&=\max_{\V{p}\succeq\V{0}}\  H_\ell(\up{p})+\frac{1}{2}(\V{b}-\V{a})^{\text{T}}\B{\eta}^*-\frac{1}{2}(\V{b}+\V{a})^{\text{T}}\B{\mu}^*-\nu^*\\&\hspace{0.8cm}+\sum_{x\in\set{X},y\in\set{Y}}\up{p}(x,y)(\Phi(x,y)^{\text{T}}\B{\mu}^*+\nu^*)\\
&=\max_{\up{p}\in\widetilde{\set{U}}}\min_{\up{h}\in \text{T}(\set{X},\set{Y})}\widetilde{\ell}(\up{h},\up{p})=\min_{\up{h}\in \text{T}(\set{X},\set{Y})}\max_{\up{p}\in\widetilde{\set{U}}}\widetilde{\ell}(\up{h},\up{p})
\end{align*}
where the second equality is obtained because strong duality holds between optimization problem \eqref{max-ent-proof} and  $\mathscr{P}_{\ell}^{\V{a},\V{b}}$ in \eqref{opt-prob} since the constraints are affine \cite{BoyVan:04}, the third equality is due to the fact that a solution of the primal \eqref{max-ent-proof} belongs to $\widetilde{\set{U}}$, and the last equality is due to the fact that $\widetilde{\ell}(\up{h},\up{p})$ is lower semi-continuous and convex-concave, and both $\widetilde{\set{U}}$ and $\text{T}(\set{X},\set{Y})$ are convex and compact. Therefore, the inequalities in \eqref{ineqs} are in fact equalities, so that the solutions of $\min_{\up{h}\in \text{T}(\set{X},\set{Y})}\max_{\up{p}\in\widetilde{\set{U}}}\widetilde{\ell}(\up{h},\up{p})$ are solutions of $\min_{\up{h}\in \text{T}(\set{X},\set{Y})}\max_{\up{p}\in\set{U}^{\V{a},\V{b}}}\ell(\up{h},\up{p})$. 
\end{IEEEproof}

\section{Proof of Corollary~\ref{cor-reg}}\label{apd:proof_cor_2}
\begin{IEEEproof}
For $\V{a}_n$ and $\V{b}_n$ given by \eqref{interval}, taking $\B{\mu}_1=(\B{\eta}+\B{\mu})/2$ and $\B{\mu}_2=(\B{\eta}-\B{\mu})/2$, optimization problem $\mathscr{P}_{\ell}^{\V{a},\V{b}}$ becomes
\begin{align}\label{reg-opt}\begin{split}\underset{\B{\mu}_1,\B{\mu}_2,\nu}{\min}&\quad-\B{\tau}_n^{\text{T}}(\B{\mu}_1-\B{\mu}_2)+\frac{1}{\sqrt{n}}\B{\lambda}^\text{T}(\B{\mu}_1+\B{\mu}_2)-\nu\\
\mbox{s.t.}&\quad \B{\Phi}(x,\pun)(\B{\mu}_1-\B{\mu}_2)+\nu\V{1}\in\set{L},\  \forall x\in\set{X}\\
&\quad \B{\mu}_1,\B{\mu}_2\succeq\V{0}
\end{split}\end{align}
for $\B{\lambda}\succeq\V{0}$. 

We first show that if $\B{\mu}_1^*$, $\B{\mu}_2^*$, and $\nu^*$ form a solution of such problem, then $\lambda^{(i)}>0$ implies $(\mu_1^*)^{(i)}(\mu_2^*)^{(i)}=0$. If there is an $i$ with $\lambda^{(i)}>0$ and $(\mu_1^*)^{(i)}(\mu_2^*)^{(i)}\neq 0$, it would exist $\epsilon>0$ with  $(\mu_1^*)^{(i)}>\epsilon$ and $(\mu_2^*)^{(i)}>\epsilon$. Hence, $\widetilde{\B{\mu}}_1$, $\widetilde{\B{\mu}}_2$, $\nu^*$ with
 \begin{align*}(\widetilde{\mu}_1)^{(j)}&=(\mu_1^*)^{(j)},&\   (\widetilde{\mu}_2)^{(j)}&=(\mu_2^*)^{(j)}\mbox{ for }j\neq i\\
(\widetilde{\mu}_1)^{(i)}&=(\mu_1^*)^{(i)}-\epsilon,&\   (\widetilde{\mu}_2)^{(i)}&=(\mu_2^*)^{(i)}-\epsilon\end{align*}
would satisfy the constraints in \eqref{reg-opt} and
\begin{align*}&-\B{\tau}_n^{\text{T}}(\widetilde{\B{\mu}}_1-\widetilde{\B{\mu}}_2)+\frac{1}{\sqrt{n}}\B{\lambda}^\text{T}(\widetilde{\B{\mu}}_1+\widetilde{\B{\mu}}_2)-\nu^*\\
&=-\B{\tau}_n^{\text{T}}(\B{\mu}_1^*-\B{\mu}_2^*)+\frac{1}{\sqrt{n}}\B{\lambda}^\text{T}(\B{\mu}_1^*+\B{\mu}^*_2)-\nu^*-\frac{2\lambda^{(i)}}{\sqrt{n}}\epsilon\end{align*}
which contradicts the fact that $\B{\mu}_1^*$, $\B{\mu}_2^*$, and $\nu^*$ form a solution of optimization problem \eqref{reg-opt}.
Therefore, the result is obtained since for positive vectors such that $\mu_1^{(i)}\mu_2^{(i)}=0$ for $i$ with $\lambda^{(i)}>0$ we have that $$\frac{1}{\sqrt{n}}\B{\lambda}^\text{T}(\B{\mu}_1+\B{\mu}_2)=\frac{1}{\sqrt{n}}\sum_{i=1}^m\lambda^{(i)}|\mu_1^{(i)}-\mu_2^{(i)}|$$
and $\B{\mu}=\B{\mu}_1-\B{\mu}_2$.
\end{IEEEproof}

\section{Proof of Theorem~\ref{th2}}\label{apd:proof_th_2}
\begin{IEEEproof}
The upper bound is given in Corollary~\ref{cor0}. For the lower bound,  we have that 
$$\min_{\up{p}\in\set{U}^{\V{a},\V{b}}}\ell(\up{h}^{\V{a},\V{b}},\up{p})\leq\ell(\up{h}^{\V{a},\V{b}},\up{p}^*)=R_\ell(\up{h}^{\V{a},\V{b}})$$
since $\up{p}^*\in\set{U}^{\V{a},\V{b}}$. The minimization above is equivalent to
\begin{align}\label{opt-aux}\begin{split}\underset{\V{p}}{\min}&\quad \V{p}^\text{T}\B{\ell}+I_+(\V{p})\\
\mbox{s.t.}&\quad -\V{p}^{\text{T}}\V{1}=-1\\
&\quad \V{a}\preceq\B{\Phi}\V{p}\preceq\V{b}\end{split}\end{align}
where $\V{p}$, $\B{\ell}$, and $\B{\Phi}$ denote the vectors and matrix with rows $p(x,y)$, $\ell(\up{h}^{\V{a},\V{b}},(x,y))$ and $\Phi(x,y)^{\text{T}}$, respectively, for $x\in\set{X}$, $y\in\set{Y}$, and $$I_+(\V{p})=\left\{\begin{array}{cc}0 &\mbox{if}\  \V{p}\succeq \V{0}\\\infty&\mbox{otherwise.}\end{array}\right.$$
The Lagrange dual of \ref{opt-aux} is
\begin{align}\begin{split}\label{opt-bound-aux}\underset{\B{\mu}_a,\B{\mu}_b,\nu}{\max}&\quad \V{a}^{\text{T}}\B{\mu}_a-\V{b}^{\text{T}}\B{\mu}_b+\nu\\
\mbox{s.t.}&\quad \Phi(x,y)^{\text{T}}(\B{\mu}_a-\B{\mu}_b)+\nu\leq \ell(\up{h}^{\V{a},\V{b}},(x,y)),\\& \qquad \qquad\qquad\qquad\qquad 
\forall x\in\set{X},y\in\set{Y}\\& \quad 
\B{\mu}_a,\B{\mu}_b\succeq\V{0}\end{split}\end{align}
because the conjugate function of $f(\V{p})=\V{p}^\text{T}\B{\ell}+I_+(\V{p})$ is
$$f^*(\V{v})=\sup_{\V{p}\succeq \V{0}}\V{p}^{\text{T}}(\V{v}-\B{\ell})=\left\{\begin{array}{cc}0&\mbox{if }\V{v}\preceq\B{\ell}\\
\infty&\mbox{otherwise.}\end{array}\right.$$
Hence, the result is obtained since $\underline{\mathscr{P}}_{\;\!\ell}^{\V{a},\V{b}}$ in \eqref{opt-bound} is equivalent to \eqref{opt-bound-aux} taking $\B{\mu}=\B{\mu}_a-\B{\mu}_b$ and $\B{\eta}=\B{\mu}_a+\B{\mu}_b$.
\end{IEEEproof}
\section{Proof of Theorem~\ref{th3}}\label{apd:proof_th_3}
The proof uses the following lemma that can be proven analogously to Theorem~\ref{th2}.
\begin{lemma}\label{lemma}
If $h\in \text{T}(\set{X},\set{Y})$ is a classification rule, and $\overline{R}$ is the optimal value of
\begin{align}\label{opt-bound-upper}
\begin{split}\underset{\B{\mu}_a,\B{\mu}_b,\nu}{\min}&\quad \V{b}^{\text{T}}\B{\mu}_b-\V{a}^{\text{T}}\B{\mu}_a-\nu\\
\mbox{s.t.}&\quad \Phi(x,y)^{\text{T}}(\B{\mu}_a-\B{\mu}_b)+\nu\leq-\ell(\up{h},(x,y)),\\
&\qquad \qquad \qquad \qquad\qquad \forall x\in\set{X},y\in\set{Y}\\
&\quad \B{\mu}_a,\B{\mu}_b\succeq\V{0}.\end{split}\end{align}
Then, we have that $\ell(\up{h},\up{p})\leq \overline{R}$ for any $\up{p}\in\set{U}^{\V{a},\V{b}}$. 
\end{lemma}
\begin{IEEEproof}[Proof of Theorem~\ref{th3}]
Firstly, with probability at least $1-\delta$ we have that $\up{p}^*\in\set{U}^{\V{a}_n,\V{b}_n}$ and 
$$\|\B{\tau}_{\infty}-\B{\tau}_{n}\|_\infty\leq \|\V{d}\|_\infty\sqrt{\frac{\log m+\log\frac{2}{\delta}}{2n}}=\frac{\|\B{\lambda}\|_\infty}{\sqrt{n}}$$
because, using Hoeffding's inequality \cite{BouLugMas:13} we have that for $i=1,2,\ldots,m$
$$\mathbb{P}\left\{|\tau_{\infty}^{(i)}-\tau_{n}^{(i)}|< t^{(i)}\right\}\geq1-2\exp\left\{-\frac{2n^2(t^{(i)})^2}{n(d^{(i)})^2}\right\}$$
so taking $t^{(i)}=d^{(i)}\sqrt{\frac{\log m+\log\frac{2}{\delta}}{2n}}$ we get
\begin{align*}\mathbb{P}&\left\{|\tau_{\infty}^{(i)}-\tau_{n}^{(i)}|< d^{(i)}\sqrt{\frac{\log m+\log\frac{2}{\delta}}{2n}}\right\} \\  &\geq 1-2\exp\left\{-\log m-\log\frac{2}{\delta}\right\}= 1-\frac{\delta}{m}\end{align*}
and using the union bound we have that 
\begin{align*}&\mathbb{P}\Bigg\{|\tau_{\infty}^{(i)}-\tau_{n}^{(i)}|< d^{(i)}\sqrt{\frac{\log m+\log\frac{2}{\delta}}{2n}},\  i=1,2,\ldots,m\Bigg\}
\\&\geq 1-m+\sum_{i=1}^m\mathbb{P}\left\{|\tau_{\infty}^{(i)}-\tau_{n}^{(i)}|< d^{(i)}\sqrt{\frac{\log m+\log\frac{2}{\delta}}{2n}}\right\}\\&\geq 1-\delta.
\end{align*}


For the result in \eqref{bound0}, we have that $\ell(\up{h}^{\V{a}_n,\V{b}_n},\up{p}^*)\leq H_\ell(\set{U}^{\V{a}_n,\V{b}_n})$ with probability at least $1-\delta$ since $p^*\in\set{U}^{\V{a}_n,\V{b}_n}$ with probability at least $1-\delta$. 

For the result in \eqref{bound1}, let $\B{\mu}^*,\nu^*$ be a solution of $\mathscr{P}_{\ell}^{\V{a}}$ in \eqref{opt-eq} for $\V{a}=\B{\tau}_\infty$; $\left[\B{\mu}^*,(\B{\mu}^*)_++(-\B{\mu}^*)_+,\nu^*\right]$ is a feasible point of $\mathscr{P}_{\ell}^{\V{a},\V{b}}$ in \eqref{opt-prob} because $(\B{\mu}^*)_++(-\B{\mu}^*)_++\B{\mu}^*=2(\B{\mu}^*)_+\succeq\V{0}$, $(\B{\mu}^*)_++(-\B{\mu}^*)_+-\B{\mu}^*=2(-\B{\mu}^*)_+\succeq\V{0}$, and $\B{\mu}^*,\nu^*$ is a feasible point of $\mathscr{P}_{\ell}^{\V{a}}$. Hence $H_\ell(\set{U}^{\V{a}_n,\V{b}_n})$ is not larger than
\begin{align*}&(\V{b}_n-\V{a}_n)^{\text{T}}((\B{\mu}^*)_++(-\B{\mu}^*)_+)/2-(\V{b}_n+\V{a}_n)^{\text{T}}\B{\mu}^*/2-\nu^*\nonumber\\
=&\, \V{b}_n^{\text{T}}(-\B{\mu}^*)_+-\V{a}_n^{\text{T}}(\B{\mu}^*)_+-\nu^*\\=&\,H_\ell(\set{U}^{\B{\tau}_{\infty}})+(\V{b}_n-\B{\tau}_{\infty})^{\text{T}}(-\B{\mu}^*)_++(\B{\tau}_{\infty}-\V{a}_n)^{\text{T}}(\B{\mu}^*)_+\\
=&\,H_\ell(\set{U}^{\B{\tau}_{\infty}})-\left(\B{\tau}_{\infty}-\B{\tau}_n-\B{\lambda}/\sqrt{n}\right)^{\text{T}}(-\B{\mu}^*)_+\\&+\left(\B{\tau}_{\infty}-\B{\tau}_n+\B{\lambda}/\sqrt{n}\right)^{\text{T}}(\B{\mu}^*)_+\\=&\,
H_\ell(\set{U}^{\B{\tau}_{\infty}})+(\B{\tau}_n-\B{\tau}_{\infty})^{\text{T}}\B{\mu}^*+\frac{\B{\lambda}^{\text{T}}}{\sqrt{n}}((\B{\mu}^*)_++(-\B{\mu}^*)_+).\end{align*}
Then, the result is obtained using H\"{o}lder inequality and the fact that $\|(\B{\mu}^*)_++(-\B{\mu}^*)_+\|_1=\|\B{\mu}^*\|_1$. 


For the result in \eqref{bound2}, using Lemma~\ref{lemma} we have that with probability at least $1-\delta$
\begin{align*}\ell(\up{h}^{\B{\tau}_n}&,\up{p}^*)\leq\max_{\up{p}\in\set{U}^{\V{a}_n,\V{b}_n}}\ell(\up{h}^{\B{\tau}_n},\up{p})\\
&\begin{array}{cl}=\underset{\B{\mu}_a,\B{\mu}_b,\nu}{\min}& \V{b}_n^{\text{T}}\B{\mu}_b-\V{a}_n^{\text{T}}\B{\mu}_a-\nu\\
\hspace{0.3cm} \mbox{s.t.}&\Phi(x,y)^{\text{T}}(\B{\mu}_a-\B{\mu}_b)+\nu\leq-\ell(\up{h}^{\B{\tau}_n},(x,y))\\
&\forall x\in\set{X},y\in\set{Y}\\
&\B{\mu}_a,\B{\mu}_b\succeq\V{0}\end{array}
\end{align*}
so that, if $\B{\mu}_n^*,\nu_n^*$ is a solution of $\mathscr{P}_{\ell}^{\V{a}}$ in \eqref{opt-eq} for $\V{a}=\B{\tau}_n$, we have that $\ell(\up{h}^{\B{\tau}_n},\up{p}^*)\leq \V{b}_n^{\text{T}}(-\B{\mu}_n^*)_+-\V{a}_n^{\text{T}}(\B{\mu}_n^*)_+-\nu_n^*$ because $\B{\mu}_n^*=(\B{\mu}_n^*)_+-(-\B{\mu}_n^*)_+$ and $$\Phi(x,y)^{\text{T}}((\B{\mu}_n^*)_+-(-\B{\mu}_n^*)_+)+\nu_n^* \leq-\ell(\up{h}^{\B{\tau}_n},(x,y))$$ for all $x\in\set{X}$, $y\in\set{Y}$ by definition of $\up{h}^{\B{\tau}_n}$. Therefore, the result is obtained since 
\begin{align*}\ell(\up{h}^{\B{\tau}_n},\up{p}^*)\leq\,& \left(\B{\tau}_n+\B{\lambda}/\sqrt{n}\right)^{\text{T}}(-\B{\mu}_n^*)_+\\
&-\left(\B{\tau}_n-\B{\lambda}/\sqrt{n}\right)^{\text{T}}(\B{\mu}_n^*)_+- \nu_n^*\\
=\,& H_\ell(\set{U}^{\B{\tau}_n})+\frac{\B{\lambda}^{\text{T}}}{\sqrt{n}}\left((\B{\mu}_n^*)_++(-\B{\mu}_n^*)_+\right).
\end{align*}

For the result in \eqref{bound4}, note that using Lemma~\ref{lemma} and since $\up{p}^*\in\set{U}^{\B{\tau}_{\infty}}$ we have that
\begin{align*}\ell(\up{h}^{\B{\tau}_n},\up{p}^*)&\leq\max_{\up{p}\in\set{U}^{\B{\tau}_{\infty}}}\ell(\up{h}^{\B{\tau}_n},\up{p})\\
&\begin{array}{cl}=\underset{\B{\mu},\nu}{\min}& -\B{\tau}_{\infty}^{\text{T}}\B{\mu}-\nu\\
\hspace{0.3cm} \mbox{s.t.}&\Phi(x,y)^{\text{T}}\B{\mu}+\nu\leq-\ell(\up{h}^{\B{\tau}_n},(x,y))\\
&\forall x\in\set{X},y\in\set{Y}\end{array}
\end{align*}
so that, if $\B{\mu}_n^*,\nu_n^*$ is a solution of $\mathscr{P}_{\ell}^{\V{a}}$ in \eqref{opt-eq} for $\V{a}=\B{\tau}_n$, we have that $\ell(\up{h}^{\B{\tau}_n},p^*)\leq -\B{\tau}_{\infty}^{\text{T}}\B{\mu}_n^*-\nu_n^*$ because 
$$\Phi(x,y)^{\text{T}}\B{\mu}_n^*+\nu_n^*\leq-\ell(\up{h}^{\B{\tau}_n},(x,y))$$
for all $x\in\set{X}$, $y\in\set{Y}$, by definition of $\up{h}^{\B{\tau}_n}$.  Let  $\B{\mu}^*,\nu^*$ be a solution of $\mathscr{P}_{\ell}^{\V{a}}$ in \eqref{opt-eq} for $\V{a}=\B{\tau}_{\infty}$, the result is obtained since $\ell(\up{h}^{\B{\tau}_n},\up{p}^*)$ is not larger than
\begin{align} &-\B{\tau}_{\infty}^{\text{T}}\B{\mu}_n^*
-\nu_n^*+\B{\tau}_n^{\text{T}}\B{\mu}_n^*-\B{\tau}_n^{\text{T}}\B{\mu}_n^*+\B{\tau}_{\infty}^{\text{T}}\B{\mu}^*+\nu^*
+H_\ell(\set{U}^{\B{\tau}_{\infty}})\nonumber\\
&=(\B{\tau}_n-\B{\tau}_{\infty})^{\text{T}}\B{\mu}_n^*+H_\ell(\set{U}^{\B{\tau}_{\infty}})-\B{\tau}_n^{\text{T}}\B{\mu}_n^*-\nu_n^*+\B{\tau}_{\infty}^{\text{T}}\B{\mu}^*+\nu^*\nonumber\\
&\leq (\B{\tau}_n-\B{\tau}_{\infty})^{\text{T}}\B{\mu}_n^*+(\B{\tau}_{\infty}-\B{\tau}_n)^{\text{T}}\B{\mu}^*+H_\ell(\set{U}^{\B{\tau}_{\infty}})\label{ineq1}\\
&\leq\|\B{\tau}_n-\B{\tau}_{\infty}\|_\infty\|\B{\mu}_n^*-\B{\mu}^*\|_1+H_\ell(\set{U}^{\B{\tau}_{\infty}})\nonumber
\end{align}
where \eqref{ineq1} is due to the fact that $-\B{\tau}_n^{\text{T}}\B{\mu}_n^*-\nu_n^*\leq-\B{\tau}_n^{\text{T}}\B{\mu}^*-\nu^*$ since $\B{\mu}^*,\nu^*$ is a feasible point of $\mathscr{P}_{\ell}^{\V{a}}$ in \eqref{opt-eq} for $\V{a}=\B{\tau}_n$. 
\end{IEEEproof}
\section{Proof of Theorem~\ref{th1-emp-x}}\label{apd:proof_th_1-emp-x}
\begin{IEEEproof}
Let $\set{X}_{\text{T}}=\{x_1,x_2,\ldots,x_n\}\subset\set{X}$ and $\bar{\up{p}}:\set{X}_{\text{T}}\times\set{Y}\to\mathbb{R}$ be the set of training instances and a function defined over training instances. If
$$\overline{H}(\bar{\up{p}})=\sum_{i=1}^n\min_{\up{q}\in\Delta(\set{Y})}\sum_{y\in\set{Y}}\bar{\up{p}}(x_i,y)L(\up{q},y)$$ 
then, for $\up{p}\in\set{U}_x^{\V{a},\V{b}}$ if $\bar{\up{p}}(x,y)=\up{p}(x,y)$ for $x\in\set{X}_\text{T}$, $y\in\set{Y}$ we have that $H_\ell(\up{p})=\overline{H}(\bar{\up{p}})$. Therefore, the result can be obtained analogously to that for Theorem~\ref{th1} since
$\max_{\up{p}\in\set{U}_x^{\V{a},\V{b}}}H_\ell(\up{p})$ is equivalent to
\begin{align*}
\begin{split}\underset{\bar{\up{p}}}{\max}&\quad \overline{H}(\bar{\up{p}})-I_+(\bar{\up{p}})\\
\mbox{s.t.}&\quad -\sum_{y\in\set{Y}}\bar{\up{p}}(x_i,y)=-\frac{1}{n}, i=1,2,\ldots,n\\
&\quad \V{a}\preceq\overline{\B{\Phi}}^{\text{T}}\bar{\V{p}}\preceq\V{b}\end{split}
\end{align*}
where $\bar{\V{p}}$ and $\overline{\B{\Phi}}$ denote the vector and matrix with rows $\up{p}(x_i,y)$ and $\Phi(x_i,y)^{\text{T}}$, respectively, for $\in\set{X}_{\text{T}}$, $y\in\set{Y}$, and
\begin{align*}
I_+(\bar{\V{p}})=\left\{\begin{array}{cc}0, &\mbox{if}\  \bar{\V{p}}\succeq \V{0}\\[3pt]
\infty,&\mbox{otherwise.}\end{array}\right.
\end{align*}
The Lagrange dual of such optimization problem is 
\begin{align*}
\begin{split}
\underset{\B{\mu}_a,\B{\mu}_b,\B{\nu}}{\min} &\quad \V{b}^{\text{T}}\B{\mu}_b-\V{a}^{\text{T}}\B{\mu}_a-\frac{1}{n}\sum_{i=1}^n\nu^{(i)}\\ 
& \quad +(-\overline{H}+I_+)^*(\overline{\B{\Phi}}(\B{\mu}_a-\B{\mu}_b)+\widetilde{\B{\nu}})\\
\text{s.t.} & \quad \B{\mu}_a,\B{\mu}_b\succeq\V{0}
\end{split}
\end{align*}
where  $\widetilde{\B{\nu}}$ is the vector in $\mathbb{R}^{|\set{X}_\text{T}||\set{Y}|}$ with component corresponding with $(x_i,y)\in\set{X}_\text{T}\times\set{Y}$ given by $\nu^{(i)}$.
Such optimization problem is equivalent to $\mathscr{P}_{x,\ell}^{\V{a},\V{b}}$ in \eqref{opt-prob-emp-x} because
\begin{align*}(-\overline{H}+I_+)^*(\V{v})=&\sum_{i=1}^n\sup_{\bar{p}(x_i,\pun)\succeq\V{0}}\big(\sum_{y\in\set{Y}}v(x_i,y)\bar{\up{p}}(x_i,y)\\&+\inf_{\up{q}\in\Delta(\set{Y})}\sum_{y\in\set{Y}}\bar{\up{p}}(x_i,y)L(\up{q},y)\big)\\
=&\sum_{i=1}^n\chi^*(v(x_i,\pun))
\end{align*}
for $\chi$ the support function of $\set{L}$ as given in \eqref{chi} and $\chi^*$ its conjugate function given in \eqref{chi-conj}. 
Hence, the dual problem becomes 
\begin{align}\notag
    \begin{split}\underset{\B{\mu}_a,\B{\mu}_b,\B{\nu}}{\min} &\quad \V{b}^{\text{T}}\B{\mu}_b-\V{a}^{\text{T}}\B{\mu}_a-\frac{1}{n}\sum_{i=1}^n\nu^{(i)}\\\mbox{s.t.}&\quad \B{\Phi}(x_i,\pun)(\B{\mu}_a-\B{\mu}_b)+\nu^{(i)}\V{1}\in \set{L}, i=1,2,\ldots,n
\\&\quad \B{\mu}_a,\B{\mu}_b\succeq\V{0}\end{split}
\end{align}
and then the expression of $\mathscr{P}_{x,\ell}^{\V{a},\V{b}}$ in \eqref{opt-prob-emp-x} is obtained carrying out the minimization above with respect to each $\nu^{(i)}$ fixing the other variables, and then taking $\B{\mu}=\B{\mu}_a-\B{\mu}_b$ and $\B{\eta}=\B{\mu}_a+\B{\mu}_b$.\end{IEEEproof}
\section{Proof of Corollary~\ref{cor-0-1-emp}}\label{apd:proof_cor_0-1_emp}
\begin{IEEEproof}
The result is obtained as a direct consequence of Theorem~\ref{th1-emp-x} taking into account that for 0-1-loss
\begin{align*}\B{\Phi}(x,\pun)\B{\mu}+\nu\V{1}\in\set{L}&\;\Leftrightarrow\;\sum_{y\in\set{Y}}(\Phi(x,y)^{\text{T}}\B{\mu}+\nu+1)_+\leq 1\\
&\;\Leftrightarrow\; \underset{\set{C}\subseteq\set{Y}}{\max}\sum_{y\in\set{C}}\big(\Phi(x,y)^{\text{T}}\B{\mu}+\nu+1\big)\leq 1\end{align*}
so that 
\begin{align*}&\begin{array}{ccl}\varphi_{0\text{-}1}(\B{\mu},x)=&\max&\nu\\&
\end{array}
\\[-0.5cm]&\hspace{2.5cm}\mbox{s.t.}\hspace{.5cm}\underset{\set{C}\subseteq\set{Y}}{\max}\sum_{y\in\set{C}}\big(\Phi(x,y)^{\text{T}}\B{\mu}+\nu+1\big)\leq 1\\
&\hspace{1.7cm}=\underset{\set{C}\subseteq\set{Y},\set{C}\neq\emptyset}{\min}\frac{1-\sum_{y\in\set{C}}\big(\Phi(x,y)^{\text{T}}\B{\mu}+1\big)}{|\set{C}|}.\end{align*}
\end{IEEEproof}

\section{Proof of Corollary~\ref{cor-log-emp}}\label{apd:proof_cor_log_emp}
\begin{IEEEproof}
The result is obtained as a direct consequence of Theorem~\ref{th1-emp-x} taking into account that for log-loss 
\begin{align*}\B{\Phi}(x,\pun)\B{\mu}+\nu\V{1}\in\set{L}&\,\Leftrightarrow\,\sum_{y\in\set{Y}}\exp\{\Phi(x,y)^{\text{T}}\B{\mu}+\nu\}\leq 1 \\&\,\Leftrightarrow\,
 \nu\leq -\log\sum_{y\in\set{Y}}\exp\{\Phi(x,y)^{\text{T}}\B{\mu}\}\end{align*}
so that 
$$\varphi_{\log}(\B{\mu},x)=-\log\sum_{y\in\set{Y}}\exp\{\Phi(x,y)^{\text{T}}\B{\mu}\}$$
and in this case
$$\up{h}^{\V{a},\V{b}}(y|x)\geq \exp\{\Phi(x,y)^{\text{T}}\B{\mu}^*+\varphi_{\text{log}}(\B{\mu}^*,x)\}$$
implies that both sides are equal since the sum of right side for varying $y\in\set{Y}$ is one.
\end{IEEEproof}
\section{Proof of Theorem~\ref{th1-emp-y}}\label{apd:proof_th_1-emp-y}
\begin{IEEEproof}
The result is obtained analogously to that for Theorem~\ref{th1} since $\max_{\up{p}\in\set{U}_y^{\V{a},\V{b}}}H_\ell(\up{p})$ is equivalent to
\begin{align*}
    \begin{split}\underset{\up{p}}{\max}&\quad H_\ell(\up{p})-I_+(\up{p})\\
\mbox{s.t.}&\quad -\sum_{x\in\set{X}}\up{p}(x,y)=-\up{p}_0(y),\  y\in\set{Y}\\
&\quad -\B{\Phi}^{\text{T}}\V{p}\preceq-\V{a}\\
&\quad \B{\Phi}^{\text{T}}\V{p}\preceq\V{b}\end{split}
\end{align*}
where $\V{p}$ and $\B{\Phi}$ denote the vector and matrix with rows $\up{p}(x,y)$ and $\Phi(x,y)^{\text{T}}$, respectively, for $x\in\set{X}$, $y\in\set{Y}$, and $$I_+(\V{p})=\left\{\begin{array}{cc}0, &\mbox{if}\  \V{p}\succeq \V{0}\\[3pt]\infty, &\mbox{otherwise.}\end{array}\right.$$
The Lagrange dual of such optimization problem is 
\begin{align*}
\begin{split}
\underset{\B{\mu}_a,\B{\mu}_b,\B{\nu}}{\min}&\quad \V{b}^{\text{T}}\B{\mu}_b-\V{a}^{\text{T}}\B{\mu}_a-\V{p}_0^{\text{T}}\B{\nu}\\&\quad +(-H_\ell+I_+)^*(\B{\Phi}(\B{\mu}_a-\B{\mu}_b)+\widetilde{\B{\nu}})\\\mbox{s.t.} &\quad \B{\mu}_a,\B{\mu}_b\succeq\V{0}\end{split}
\end{align*}
where $\widetilde{\B{\nu}}$ is a vector in $\mathbb{R}^{|\set{X}||\set{Y}|}$ with component corresponding with $(x,y)\in\set{X}\times\set{Y}$ given by $\nu^{(y)}$.
Such optimization problem is equivalent to $\mathscr{P}_{y,\ell}^{\V{a},\V{b}}$ in \eqref{opt-prob-emp-y} taking $\B{\mu}=\B{\mu}_a-\B{\mu}_b$ and $\B{\eta}=\B{\mu}_a+\B{\mu}_b$ because
\begin{align*}(-H_\ell+I_+)^*(\V{v})=\sum_{x\in\set{X}}\chi^*(v(x,\pun))
\end{align*}
for $\chi^*$ the function given in \eqref{chi-conj}. 
\end{IEEEproof}
\section{Proof of Theorem~\ref{th1-emp-xy}}\label{apd:proof_th_1-emp-xy}
\begin{IEEEproof}
Let $\set{X}_{\text{T}}=\{x_1,x_2,\ldots,x_n\}\subset\set{X}$ and  $\bar{\up{p}}:\set{X}_{\text{T}}\times\set{Y}\to\mathbb{R}$ be the set of training instances and a function defined over training instances. If 
$$\overline{H}(\bar{\up{p}})=\sum_{i=1}^n\min_{\up{q}\in\Delta(\set{Y})}\sum_{y\in\set{Y}}\bar{\up{p}}(x_i,y)L(\up{q},y)$$
then, for $\up{p}\in\set{U}_{x,y}^{\V{a},\V{b}}$ if $\bar{\up{p}}(x,y)=\up{p}(x,y)$ for $x\in\set{X}_\text{T}$, $y\in\set{Y}$ we have that $H_\ell(\up{p})=\overline{H}(\bar{\up{p}})$. Therefore, the result is obtained as that for Theorem~\ref{th1} since $\max_{\up{p}\in\set{U}_{x,y}^{\V{a},\V{b}}}\overline{H}(\bar{\up{p}})$ is equivalent to
\begin{align*}
\begin{split}
\underset{\bar{\up{p}}}{\max}&\quad \overline{H}(\bar{\up{p}})-I_+(\bar{\up{p}})\\
\mbox{s.t.}&\quad -\sum_{i=1}^n\bar{\up{p}}(x_i,y)=-\up{p}_0(y),\  y\in\set{Y}\\
&\quad -\sum_{y\in\set{Y}}\bar{\up{p}}(x_i,y)=-\frac{1}{n},\  i=1,2,\ldots,n\\
&\quad \V{a}\preceq\overline{\B{\Phi}}^{\text{T}}\bar{\V{p}}\preceq\V{b}\end{split}    
\end{align*}
where $\bar{\V{p}}$ and $\overline{\B{\Phi}}$ denote the vector and matrix with rows $\up{p}(x_i,y)$ and $\Phi(x_i,y)^{\text{T}}$, respectively, for $x\in\set{X}_{\text{T}}$, $y\in\set{Y}$, and
$$I_+(\bar{\V{p}})=\left\{\begin{array}{cc}0 &\mbox{if}\  \bar{\V{p}}\succeq \V{0}\\\infty&\mbox{otherwise.}\end{array}\right.$$
The Lagrange dual of such optimization problem is 
\begin{align*}
\begin{split}\underset{\B{\mu}_a,\B{\mu}_b,\B{\nu},\B{\rho}}{\min}&\quad \V{b}^{\text{T}}\B{\mu}_b-\V{a}^{\text{T}}\B{\mu}_a-\V{p}_0^{\text{T}}\B{\nu}-\frac{1}{n}\sum\rho^{(i)}\\&\quad +(-\overline{H}+I_+)^*(\overline{\B{\Phi}}(\B{\mu}_a-\B{\mu}_b)+\widetilde{\B{\nu}}+\widetilde{\B{\rho}})\\\mbox{s.t.} & \quad \B{\mu}_a,\B{\mu}_b\succeq\V{0}\end{split}
\end{align*}
where $\widetilde{\B{\nu}}$ and $\widetilde{\B{\rho}}$ are the vectors in $\mathbb{R}^{|\set{X}_\text{T}||\set{Y}|}$ with component corresponding with $(x_i,y)\in\set{X}_\text{T}\times\set{Y}$ given by $\nu^{(y)}$ and $\rho^{(i)}$, respectively.
Such optimization problem is equivalent to $\mathscr{P}_{x,y,\ell}^{\V{a},\V{b}}$ in \eqref{opt-prob-emp-xy} taking $\B{\mu}=\B{\mu}_a-\B{\mu}_b$ and $\B{\eta}=\B{\mu}_a+\B{\mu}_b$ because
\begin{align*}(-\overline{H}+I_+)^*(\V{v})=&\sum_{i=1}^n\sup_{\bar{\up{p}}(x_i,\pun)\succeq\V{0}}\big(\sum_{y\in\set{Y}}v(x_i,y)\bar{\up{p}}(x_i,y)\\&+\inf_{\up{q}\in\Delta(\set{Y})}\sum_{y\in\set{Y}}\bar{\up{p}}(x_i,y)L(q,y)\big)\\
=&\sum_{i=1}^n\chi^*(v(x_i,\pun))
\end{align*}
for $\chi$ the support function of $\set{L}$ as given in \eqref{chi} and $\chi^*$ its conjugate function given in \eqref{chi-conj}. \end{IEEEproof}

\section*{Acknowledgment}
The authors would like to thank Prof. Peter D. Gr\"{u}nwald for his valuable comments and careful reading of the manuscript.

\bibliographystyle{IEEEtran}

\vspace{-1.25cm}
\begin{IEEEbiographynophoto}{Santiago Mazuelas}
(Senior Member, IEEE) received the Ph.D. in Mathematics and Ph.D. in Telecommunications Engineering from the University of Valladolid, Spain, in 2009 and 2011, respectively.

Since 2017 he has been Ramon y Cajal Researcher at the Basque Center for Applied Mathematics (BCAM). Prior to joining BCAM, he was a Staff Engineer at Qualcomm Corporate Research and Development from 2014 to 2017. He previously worked from 2009 to 2014 as Postdoctoral Fellow and Associate at the Massachusetts Institute of Technology (MIT). His general research interest is the application of mathematics to solve practical problems, currently his work is primarily focused on statistical signal processing, machine learning, and data science.

Dr. Mazuelas is Area Editor (signal processing) for the IEEE COMMUNICATIONS LETTERS since Oct. 2019 (Associate Editor from Jan. 2017 to Sep. 2019), and served as Technical Program Vice-chair at the 2021 IEEE Globecom as well as Symposium Co-chair at the 2014 IEEE Globecom, the 2015 IEEE ICC, and the 2020 IEEE ICC. He has received the Young Scientist Prize from the Union Radio-Scientifique Internationale (URSI) Symposium in 2007, and the Early Achievement Award from the IEEE ComSoc in 2018. His papers received the IEEE Communications Society Fred W. Ellersick Prize in 2012, and Best Paper Awards from the IEEE ICC in 2013, the IEEE ICUWB in 2011, and the IEEE Globecom in 2011. 
\end{IEEEbiographynophoto}
\vspace{-1.25cm}
\begin{IEEEbiographynophoto}{Yuan Shen}
(Senior Member, IEEE) received the B.E. degree (Hons.) in electronic engineering from Tsinghua University, Beijing, China, in 2005, and the S.M. and Ph.D. degrees in electrical engineering from the Massachusetts Institute of Technology (MIT), Cambridge, MA, USA, in 2008 and 2014, respectively.

He is currently a Professor and Vice Chair with the Department of Electronic Engineering, Tsinghua University, Beijing, China. His research interests include network localization and navigation, inference techniques, resource allocation, and cooperative networks. 

Dr.\ Shen is a recipient of the IEEE ComSoc Radio Communications Technical Committee Outstanding Service Award, the IEEE Communication Society Asia- Pacific Board Outstanding Young Researcher Award, and the Marconi Society Paul Baran Young Scholar Award. He has received the IEEE ComSoc Fred W. Ellersick Prize and the Best Paper Awards from the IEEE conferences. He was the Elected Chair for the IEEE ComSoc Radio Communications Committee for the period of 2019-2020, and serves as the TPC Symposium Co-Chair for the IEEE ICC and Globecom for several times. He is the Editor of the {\sc IEEE Transactions on Wireless Communications}, {\sc IEEE Wireless Communications Letters}, and {\sc China Communications}.
\end{IEEEbiographynophoto}

\begin{IEEEbiographynophoto}{Aritz P\'{e}rez} received his Ph.D. degree in 2010 from the University of Basque Country, Department of Computer Science and Artificial Intelligence. From 2010 to 2014, he was a Postdoctoral Fellow at the University of the Basque Country. Since 2014 he has been working as a researcher at the Basque Center for Applied Mathematics within the Machine Learning Department. His current scientific interests include supervised, unsupervised and weak classification, probabilistic graphical models, time-series data mining, and combinatorial optimization.
\end{IEEEbiographynophoto}



\end{document}